# Algorithms for Verifying Deep Neural Networks

Changliu Liu
Carnegie Mellon University
cliu6@andrew.cmu.edu

Tomer Arnon
Stanford University
tarnon@stanford.edu

Christopher Lazarus
Stanford University
clazarus@stanford.edu

Christopher Strong
Stanford University
castrong@stanford.edu

Clark Barrett
Stanford University
barrett@cs.stanford.edu

Mykel J. Kochenderfer
Stanford University
mykel@stanford.edu

October 11, 2020

**Abstract**

Deep neural networks are widely used for nonlinear function approximation, with applications ranging from computer vision to control. Although these networks involve the composition of simple arithmetic operations, it can be very challenging to verify whether a particular network satisfies certain input-output properties. This article surveys methods that have emerged recently for soundly verifying such properties. These methods borrow insights from reachability analysis, optimization, and search. We discuss fundamental differences and connections between existing algorithms. In addition, we provide pedagogical implementations of existing methods and compare them on a set of benchmark problems.

## 1 Introduction

Neural networks [26] have been widely used in many applications, such as image classification and understanding [28], language processing [42], and control of autonomous systems [44]. These networks represent functions that map inputs to outputs through a sequence of layers. At each layer, the input to that layer undergoes an affine transformation followed by a simple nonlinear transformation before being passed to the next layer. These nonlinear transformations are often called *activation functions*, and a common example is the *rectified linear unit* (ReLU), which transforms the input by setting any negative values to zero. Although the computation involved in a neural network is quite simple, these networks can represent complex nonlinear functions by appropriately choosing the matrices that define the affine transformations. The matrices are often learned from data using stochastic gradient descent.

Neural networks are being used for increasingly important tasks, and in some cases, incorrect outputs can lead to costly consequences. Traditionally, validation of neural networks has largely focused on evaluating the network on a large collection of points in the input space and determining whether the outputs are as desired. However, since the input space is effectively infinite in cardinality, it is not feasible to check all possible inputs. Even networks that perform well on a large sample of inputs may not correctly generalize to new situations and may be vulnerable to adversarial attacks [47].

This article surveys a class of methods that are capable of formally verifying properties of deep neural networks over the full input space. A property can be formulated as a statement that if the input belongs to some set $\mathcal{X}$, then the output will belong to some set $\mathcal{Y}$. To illustrate, in classification problems, it can be useful to verify that points near a training example belong to the same class as that example. In the control of physical problems, it can be useful to verify that the outputs from a network satisfy hard safety constraints.

The verification algorithms that we survey are *sound*, meaning that they will only report that a property holds if the property actually holds. Some of the algorithms that we discuss are also *complete*, meaning that whenever the property holds, the algorithm will correctly state that it holds. However, some of the algorithms compromise completeness in their use of approximations to improve computational efficiency.

The algorithms may be classified based on whether they draw insights from these three categories of analysis:

1. *Reachability.* These methods use layer-by-layer reachability analysis of the network. Representative methods are ExactReach [72], MaxSens [69], NNV [63], SymBox [38], Ai2 [25], and its advanced version ERAN [53, 54, 55, 56]. Some other approaches also use reachability methods (such as interval arithmetic) to compute bounds on the values of the nodes.

2. *Optimization.* These methods use optimization to falsify the assertion. The function represented by the neural network is a constraint to be considered in the optimization. As a result, the optimization problem is not convex. In *primal optimization*, different methods are developed to encode the nonlinear activation functions as linear constraints. Examples include NSVerify [40], MIPVerify [59], and ILP [8]. The constraints can also be simplified through *dual optimization*. Representative methods for dual optimization include Lagrangian dual methods such as Duality [20], ConvDual [68], and LagrangianDecomposition [13], and semidefinite programming methods such as Certify [50] and SDP [24].

3. *Search.* These methods search for a case to falsify the assertion. Search is usually combined with either reachability or optimization, as the latter two methods provide possible search directions. Representative methods for *search and reachability* include ReluVal [65], Neurify [64], DLV [30], Fast-Lin [66], Fast-Lip [66], CROWN [75], nnenum [5], and VeriNet [29]. Representative methods for *search and optimization* include Reluplex [32], Marabou [33], Planet [22], Sherlock [18], Venus [21], PeregriNN [34], and BaB [12] and its extensions [14, 13, 41]. Some of these methods call Boolean satisfiability (SAT) or satisfiability modulo theories (SMT) solvers [7] to verify networks with only ReLU activations.

**Scope of this article**. This article introduces a unified mathematical framework for verifying neural networks, classifies existing methods under this framework, provides pedagogical implementations of existing methods,[1] and compares those methods on a set of benchmark problems.[2]

The following topics are not included in the discussion:

- neural network testing methods that generate test cases [57, 27, 48, 58];

- white box approaches that build mappings from network parameters to some functional description [46];

- verification of binarized neural networks [16, 45, 15];

- closed-loop safety, stability and robustness by executing control policies defined by neural networks [70, 19], or verification of recurrent neural networks [1];

- training or retraining methods to make a network satisfy a property [43, 50, 68];

- robustness of the verification algorithm under floating point arithmetic [54];

- simplification or compression of the network to improve verification efficiency [23, 49].

Chapter 2 discusses the mathematical problem for verification. Chapter 3 gives an overview of the categories of methods that we will consider. Chapter 4 introduces preliminary and background mathematics. Chapter 5 discusses reachability methods. Chapter 6 discusses methods for primal optimization. Chapter 7 discusses methods for dual optimization. Chapter 8 discusses methods for search and reachability. Chapter 9 discusses methods for search and optimization. Chapter 10 compares those methods.

---

[1]Our implementation is provided in the Julia programming language. We have found the language to be ideal for specifying algorithms in human readable form [9]. The full implementation may be found at https://github.com/sisl/NeuralVerification.jl.

[2]There have been other reviews of methods for verifying neural networks. Leofante *et al.* review primal optimization methods that encode ReLU networks as mixed integer programming problems together with search and optimization under the framework of Boolean satisfiability and SMT [37]. Xiang *et al.* review a broader range of verification techniques in addition to safe control and learning [71]. Salman *et al.* review and compare methods that use convex relaxations to compute robustness bounds of ReLU networks [52].

## 2 Problem Formulation

We first review feedforward neural networks and introduce the mathematical formulation of the verification problem. We will then discuss the results provided by various algorithms along with the properties of soundness and completeness. In our discussion, we will use lowercase letters in italics for scalars and scalar functions ($x$), lowercase letters in upright bold for vectors and vector functions ($\mathbf{x}$), uppercase letters in upright bold for matrices and matrix functions ($\mathbf{X}$), and calligraphic uppercase letters for sets and set functions ($\mathcal{X}$).

### 2.1 Feedforward Neural Network

Consider an $n$-layer *feedforward neural network* that represents a function $\mathbf{f}$ with input $\mathbf{x} \in \mathcal{D}_{\mathbf{x}} \subseteq \mathbb{R}^{k_0}$ and output $\mathbf{y} \in \mathcal{D}_{\mathbf{y}} \subseteq \mathbb{R}^{k_n}$, *i.e.*, $\mathbf{y} = \mathbf{f}(\mathbf{x})$, where $k_0$ is the input dimension and $k_n$ is the output dimension. All non-vector inputs or outputs are reshaped to vectors. Each layer in $\mathbf{f}$ corresponds to a function $\mathbf{f}_i : \mathbb{R}^{k_{i-1}} \to \mathbb{R}^{k_i}$, where $k_i$ is the dimension of the hidden variable $\mathbf{z}_i$ in layer $i$. Moreover, we set $\mathbf{z}_0 = \mathbf{x}$ and $\mathbf{z}_n = \mathbf{y}$. Hence, the network can be represented by

$$\mathbf{f} = \mathbf{f}_n \circ \mathbf{f}_{n-1} \circ \cdots \circ \mathbf{f}_1, \tag{2.1}$$

where $\circ$ means function composition. The function at layer $i$ is

$$\mathbf{z}_i = \mathbf{f}_i(\mathbf{z}_{i-1}) = \boldsymbol{\sigma}_i(\mathbf{W}_i \mathbf{z}_{i-1} + \mathbf{b}_i), \tag{2.2}$$

which consists of a linear transformation defined by a weight matrix $\mathbf{W}_i \in \mathbb{R}^{k_i \times k_{i-1}}$ and a bias vector $\mathbf{b}_i \in \mathbb{R}^{k_i}$, and an activation function $\boldsymbol{\sigma}_i : \mathbb{R}^{k_i} \to \mathbb{R}^{k_i}$. All activation functions are assumed to be monotone and non-decreasing. For simplicity, let $\hat{\mathbf{z}}_i := \mathbf{W}_i \mathbf{z}_{i-1} + \mathbf{b}_i$ denote the node value before activation. Let $z_{i,j}$ be the value of the $j$th node in the $i$th layer, $\mathbf{w}_{i,j} \in \mathbb{R}^{1 \times k_{i-1}}$ be the $j$th row in $\mathbf{W}_i$, $w_{i,j,k}$ be the $k$th entry in $\mathbf{w}_{i,j}$, $b_{i,j}$ be the $j$th entry in $\mathbf{b}_i$. In the case that the activation function is node-wise, we denote the activation for the $j$th node as $\sigma_{i,j}$. We then have

$$z_{i,j} = \sigma_{i,j}\left(\mathbf{w}_{i,j} \mathbf{z}_{i-1} + b_{i,j}\right) = \sigma_{i,j}\left(\sum_k w_{i,j,k}\, z_{i-1,k} + b_{i,j}\right) = \sigma_{i,j}(\hat{z}_{i,j}). \tag{2.3}$$

Figure 2.1 shows the structure of a feedforward neural network.

A network is defined in algorithm 1. The definitions of different activation functions are listed in algorithm 2.

### 2.2 Verification Problem

Verification involves checking whether input-output relationships of a function hold. The input constraint is imposed by a set $\mathcal{X} \subseteq \mathcal{D}_{\mathbf{x}}$. The corresponding output constraint is

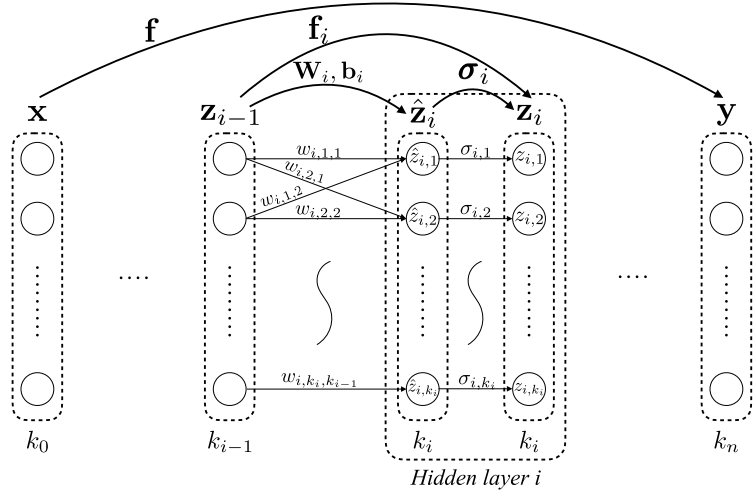

**Figure 2.1:** Illustration of a feedforward neural network with the notation used in this article.

```
abstract type ActivationFunction end

struct Layer{F<:ActivationFunction, N<:Number}
    weights::Matrix{N}
    bias::Vector{N}
    activation::F
end

struct Network
    layers::Vector{Layer}
end
```

**Algorithm 1:** Network structure. A network is a list of layers. Each layer consists of weights $\mathbf{W}$, bias $\mathbf{b}$, and activation $\boldsymbol{\sigma}$.

```
struct GeneralAct <: ActivationFunction end
struct ReLU <: ActivationFunction end
struct Id <: ActivationFunction end
(f::GeneralAct)(x) = f(x)
(f::ReLU)(x) = max.(x,0)
(f::Id)(x) = x
```

**Algorithm 2:** Activation functions. This paper mostly focuses on networks with ReLU activations. Other piece-wise linear activations can be encoded by multiple ReLU activations. For example, the Max activation function can be encoded using the following relationship: $Max(x,y) = ReLU(x-y)+y$. Composing multiple Max functions can enable the encoding of the Max pooling operator.

imposed by a set $\mathcal{Y} \subseteq \mathcal{D}_\mathbf{y}$. In the following discussion, we call the sets $\mathcal{X}$ and $\mathcal{Y}$ *constraints*. Solving the *verification problem* requires showing that the following assertion holds:

$$\mathbf{x} \in \mathcal{X} \Rightarrow \mathbf{y} = \mathbf{f}(\mathbf{x}) \in \mathcal{Y}. \tag{2.4}$$

For example, to verify robustness in a classification network,[1] we need to ensure that all samples in the neighborhood of a given input $\mathbf{x}_0$ are classified with the same label. Suppose the desired label is $i^* \in \{1, \ldots, k_n\}$. We need to ensure that $y_{i^*} > y_j$ for all $j \neq i^*$. The input and output constraints are

$$\mathcal{X} = \{\mathbf{x} : \|\mathbf{x} - \mathbf{x}_0\|_p \leq \epsilon\}, \tag{2.5a}$$
$$\mathcal{Y} = \{\mathbf{y} : y_{i^*} > y_j, \forall j \neq i^*\}, \tag{2.5b}$$

where $\epsilon$ is the maximum allowable disturbance in the input space. The metric to measure disturbance can be any $\ell_p$ norm, though the $\ell_\infty$ or the $\ell_1$ norms are common because they lead to linear constraints.

Our formulation is broader than classification problems. In general, the input set $\mathcal{X}$ and the output set $\mathcal{Y}$ can have any geometry. For simplicity, we assume that $\mathcal{X}$ is a polytope, and $\mathcal{Y}$ is either a polytope or the complement of a polytope. A *polytope* is a generalization of the three-dimensional polyhedron and is defined as the intersection of a set of halfspaces. This is the definition of a *convex polytope*; alternative definitions exist, but this is the one we will use here. Since any compact domain can be approximated by a finite set of polytopes for any required accuracy, this formulation can be easily extended to arbitrary geometries. Moreover, the complement of a polytope allows the encoding of unbounded sets. By default, a polytope is a closed set, while its complement is an open set.

Our discussion will focus on the following subclasses of polytopes:[2]

- *Halfspace-polytope* (or *H-polytope*), which represents polytopes using a set of linear inequality constraints

$$\mathbf{C}\mathbf{x} \leq \mathbf{d}, \tag{2.6}$$

  where $\mathbf{C} \in \mathbb{R}^{k \times k_0}$, $\mathbf{d} \in \mathbb{R}^k$, and $k$ is the number of inequality constraints used to define the polytope. A point $\mathbf{x}$ is in the polytope if and only if $\mathbf{C}\mathbf{x} \leq \mathbf{d}$ is satisfied.

- *Vertex-polytope* (or *V-polytope*), which represents polytopes using a set of vertices. Mathematically, it is described by a concatenation of all vertices $\mathbf{v}_i$ for $i \in \{1, \ldots, k\}$,

$$\begin{bmatrix} \mathbf{v}_1 & \mathbf{v}_2 & \cdots & \mathbf{v}_k \end{bmatrix}, \tag{2.7}$$

  where $k$ is the number of vertices. A point $\mathbf{x}$ is in the polytope if and only if $x$ is in the convex hull of the vertices.

---

[1] Given an input, a classification network outputs weights over several labels. The input is assigned the label with the highest weight.

[2] In our implementations, we use the definitions in LazySets.jl, which is a Julia package for calculus with convex sets [10]. The implementation can be found at https://github.com/JuliaReach/LazySets.jl.

- *Hyperrectangle*, which corresponds to a high-dimensional rectangle, defined by

$$|\mathbf{x} - \mathbf{c}| \leq \mathbf{r}, \tag{2.8}$$

  where $\mathbf{c} \in \mathbb{R}^{k_0}$ is the center of the hyperrectangle and $\mathbf{r} \in \mathbb{R}^{k_0}$ is the radius of the hyperrectangle. In the following discussion, we may refer to hyperrectangles as *intervals*. A hyperrectangle that has uniform side lengths is called a *hypercube*.

- *Zonotope*, which represents symmetric polytopes that can be written as affine transformations of a unit hypercube, defined by

$$\mathbf{x} = \mathbf{c} + [\ \mathbf{r}_1 \quad \mathbf{r}_2 \quad \cdots \quad \mathbf{r}_l\ ]\boldsymbol{\alpha}, |\boldsymbol{\alpha}| \leq \mathbf{1}, \tag{2.9}$$

  where $\mathbf{c} \in \mathbb{R}^{k_0}$ is the center of the zonotope, $\mathbf{r}_i \in \mathbb{R}^{k_0}$ for $i \in \{1, \ldots, l\}$ are generators of the zonotope, and $\boldsymbol{\alpha} \in \mathbb{R}^l$ is the free parameter that belongs to a unit hypercube. The parameter $l$ is called the degree of freedom of the zonotope. When $l = 1$, the zonotope reduces to a line. When $l = k_0$ and $\mathbf{r}_i \in \mathbb{R}^{k_0}$ form an orthogonal basis in $\mathbb{R}^{k_0}$, then the zonotope reduces to a hyperrectangle.

- *Star set* (convex), which generalizes zonotopes. While a general star set can be nonconvex, this survey only considers convex star sets. A convex star set is an affine transformation of an arbitrary convex polytope,

$$\mathbf{x} = \mathbf{c} + [\ \mathbf{r}_1 \quad \mathbf{r}_2 \quad \cdots \quad \mathbf{r}_l\ ]\boldsymbol{\alpha}, \mathbf{C}\boldsymbol{\alpha} \leq \mathbf{d}, \tag{2.10}$$

  where $\mathbf{C} \in \mathbb{R}^{k \times l}$, $\mathbf{d} \in \mathbb{R}^k$, and $k$ is the number of inequality constraints on $\boldsymbol{\alpha}$. The free parameter $\boldsymbol{\alpha}$ now belongs to a general polytope instead of being confined to a hypercube. A star set can encode any zonotope, but it is not constrained to be symmetric. Star sets can be used in the symbolic reachability analysis discussed in chapter 5. When used in symbolic analysis, the free parameters in $\boldsymbol{\alpha}$ are called *symbols*. In the following discussion, we simply use "star sets" to refer to "convex star sets" defined in (2.10).

- *Halfspace*, which is represented by a single linear inequality constraint

$$\mathbf{c}^\mathsf{T}\mathbf{x} \leq d, \tag{2.11}$$

  where $\mathbf{c} \in \mathbb{R}^{k_0}$ and $d \in \mathbb{R}$.

Note that the set in (2.5b) corresponds to a halfspace-polytope. When $p = \infty$, the set in (2.5a) corresponds to a hyperrectangle centered at $\mathbf{x}_0$ with uniform radius $\epsilon$.

The verification problem is defined in algorithm 3.

```julia
struct Problem{P, Q}
    network::Network
    input::P
    output::Q
end
```

**Algorithm 3:** Problem definition. It consists of a network to be verified, an input set constraint, and an output set constraint. The types `P` and `Q` can be any sets that match the requirements of the algorithm.

## 2.3   Results

Verification algorithms attempt to identify whether (2.4) holds. In some cases, algorithms may return *unknown* if no conclusion can be drawn. Different algorithms output different types of results as listed below and illustrated in figure 2.2.

- *Counter example result*, which is a counter example $x^* \in \mathcal{X}$ with

$$\mathbf{f}(\mathbf{x}^*) \notin \mathcal{Y}. \tag{2.12}$$

  The property (2.4) is violated if such a counter example is found.

- *Adversarial result*, which is the maximum allowable disturbance with respect to an $\ell_p$ norm while maintaining $\mathbf{f}(\mathbf{x}) \in \mathcal{Y}$:

$$\epsilon(\mathbf{x}_0, \mathbf{f}, \mathcal{Y}, p) := \min_{\mathbf{x},\ \text{s.t.}\ \mathbf{f}(\mathbf{x}) \notin \mathcal{Y}} \|\mathbf{x} - \mathbf{x}_0\|_p. \tag{2.13}$$

  The property (2.4) is violated if the input set $\mathcal{X}$ exceeds the maximum allowable disturbance,

$$\epsilon(\mathbf{x}_0, \mathbf{f}, \mathcal{Y}, p) < \max_{\mathbf{x} \in \mathcal{X}} \|\mathbf{x} - \mathbf{x}_0\|_p. \tag{2.14}$$

- *Reachability result*, which is the output reachable set:

$$\mathcal{R}(\mathcal{X}, \mathbf{f}) := \{\mathbf{y} : \mathbf{y} = \mathbf{f}(\mathbf{x}), \forall \mathbf{x} \in \mathcal{X}\}. \tag{2.15}$$

  The property (2.4) is violated if the reachable set does not belong to the output set $\mathcal{Y}$,

$$\mathcal{R}(\mathcal{X}, \mathbf{f}) \nsubseteq \mathcal{Y}. \tag{2.16}$$

Algorithm 4 provides definitions of these result types used in our implementation. The status may be `:holds`, `:violated`, or `:unknown`.

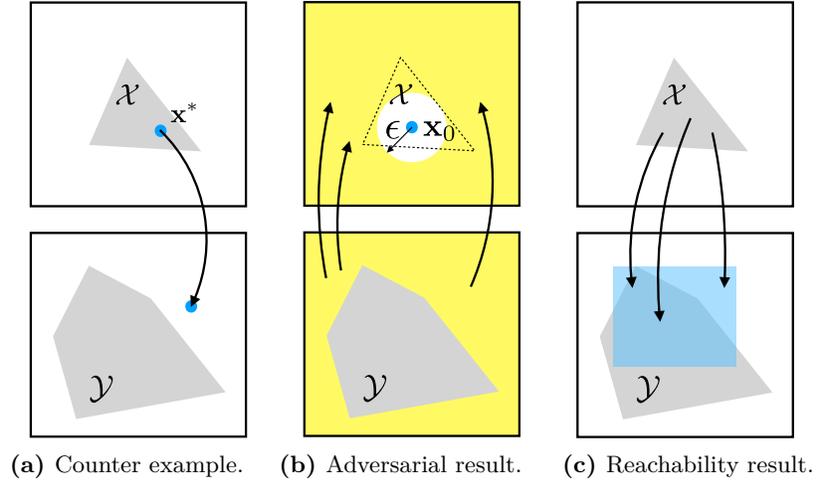

**(a)** Counter example. **(b)** Adversarial result. **(c)** Reachability result.

**Figure 2.2:** The input set $\mathcal{X}$ and the output set $\mathcal{Y}$ are shown as gray polygons. In (a), a counter example is found. In (b), the input set exceeds the maximum allowable disturbance (white circle in the input domain). In (c), the output reachable set (blue set in the output domain) does not belong to the output set.

```
abstract type Result end

struct BasicResult <: Result
    status::Symbol
end

struct CounterExampleResult <: Result
    status::Symbol
    counter_example::Vector{Float64}
end

struct AdversarialResult <: Result
    status::Symbol
    max_disturbance::Float64
end

struct ReachabilityResult <: Result
    status::Symbol
    reachable::Vector{<:AbstractPolytope}
end
```

**Algorithm 4:** Result types. `BasicResult` is for satisfiability only. `CounterExampleResult` also outputs a counter example if the problem is not satisfied. `AdversarialResult` outputs the maximum allowable disturbance. `ReachabilityResult` outputs the reachable set.

## 2.4 Soundness and Completeness

The result returned by a particular solver may not always be correct. A specific instance of (2.4) can either `:hold` or be `:violated`. The status from a solver can be holds, violated, or unknown. Ideally, a solver only outputs holds or violated to match the actual status of a given problem. However, some algorithms make approximations that can result in a mismatch. For example, the computed reachable set (denoted $\tilde{\mathcal{R}}$) may be an over-approximation of $\mathcal{R}$ in (2.15). Then, even if $\tilde{\mathcal{R}} \not\subseteq \mathcal{Y}$, *i.e.*, the solver returns violated, it is possible that $\mathcal{R} \subset \mathcal{Y}$, *i.e.*, the property actually holds.

We use the following definitions to categorize solvers:

- *Soundness*, which requires that when the solver returns holds, the property actually holds.

- *Completeness*, which requires that (i) the solver never returns unknown; and (ii) if the solver returns violated, the property is actually violated.

- *Termination*, which requires that the solver always finishes after a finite number of steps.

A method that is sound, complete, and terminating always outputs the correct result with no unknowns. All methods discussed in this survey are sound and terminating, but not all of them are complete. Some methods use over-approximations to speed up computation and result in incompleteness. Our implementation may be slightly different from the original implementation of some methods. For example, the original implementation of DLV is always complete and is sound under the minimality assumption of the search tree. Our implementation is sound but not complete. We will highlight any difference in our implementation.

Table 2.1 summarizes the characteristics of all methods considered in this survey. The input and output sets may include hyperrectangles (HR), halfspaces (HS), halfspace-polytopes (HP), vertex-polytopes (VP), and polytope complements (PC). Additionally, the superscripts in the table indicate the following constraints on the output sets:

1. The polytope must be bounded. This restriction is not due to a theoretical limitation, but rather to our implementation and will eventually be relaxed.

2. Polytope complements encode unbounded sets. They are used in optimization-based methods, to be explained in chapter 6. These methods usually encode the complement of the output set as a constraint and require the constraint to be convex. The complement of a polytope complement is a convex polytope, and hence satisfying the requirement.

| Method Name | Activation | Approach | Input/Output | Complete |
| --- | --- | --- | --- | --- |
| ExactReach [72] | ReLU | Exact Reachability | HP/HP(bound)[1] | ✓ |
| AI2 [25] | PWL | Split and Join | HP/HP(bound)[1] | × |
| MaxSens [69] | Any | Interval Arithmatic | HP/HP(bound)[1] | × |
| NSVerify [40] | ReLU | Naive MILP | HR/PC[2] | ✓ |
| MIPVerify [59] | ReLU / Max | MILP with bounds | HR/PC[2] | ✓ |
| ILP [8] | ReLU | Iterative LP | HR/PC[2] | × |
| Duality [20] | Any | Lagrangian Relaxation | HR(uni)/HS | × |
| ConvDual [68] | ReLU | Convex Relaxation | HR(uni)/HS | × |
| Certify [50] | Differentiable | Semidefinite Relaxation | HR(uni)/HS | × |
| Fast-Lin [66] | ReLU | Network Relaxation | HR/HS | × |
| Fast-Lip [66] | ReLU | Lipschitz Estimation | HR/HS | × |
| ReluVal [65] | ReLU | Symbolic Interval | HR/HR | ✓ |
| Neurify [64] | ReLU | Symbolic Interval | HP(bound)/HP | ✓ |
| DLV [30] | Any | Search in Hidden Layers | HR/HR(1-D)[3] | ✓* |
| Sherlock [18] | ReLU | Local and Global Search | HR/HR(1-D)[3] | × |
| BaB [12] | PWL | Branch and Bound | HR/HR(1-D)[3] | × |
| Planet [22] | PWL | Satisfiability (SAT) | HR/PC[2] | ✓ |
| Reluplex [32] | ReLU | Simplex | HR/PC[2] | ✓ |

**Table 2.1:** List of existing methods. We name the method if it does not have a name. The entries under "activation" show the type of activations supported in the methods. PWL stands for piecewise linear activations. The entries under "approach" summarize the key ideas of the methods. All the methods presented in this paper are sound and the methods that are complete as defined in section 2.4 are marked in the "complete" column. For DLV, the original implementation is complete but may not be sound, while our implementation is sound but not complete.

3. The output set must be 1-dimensional.[3]

## 3 Overview of Methods

This chapter overviews existing methods that are studied in this survey. Their common components will be outlined. As mentioned earlier, there are three basic verification methods, *i.e.*, reachability, optimization, and search. Regarding the three basic methods, we categorize those methods into the following five categories as shown in figure 3.1. The methods are also summarized in table 2.1.

---

[3]Note that we can always turn a verification problem on a network with multiple dimensional outputs into a verification problem on a network with 1-dimensional output. The idea is to add a few layers to the original network to directly encode the original output constraint. The new output would indicate whether the original output constraint is satisfied or not. For example, suppose the original network $\mathbf{y} = \mathbf{f}(\mathbf{x})$ needs to satisfy $\mathbf{Cy} \leq \mathbf{d}$. Then we can add a linear layer with weights $\mathbf{C}$ and bias $-\mathbf{d}$ and a max layer to the original network. The new network becomes $y' = \max(\mathbf{Cf}(\mathbf{x}) - \mathbf{d})$. The original constraint is equivalent to the constraint $y' \leq 0$.

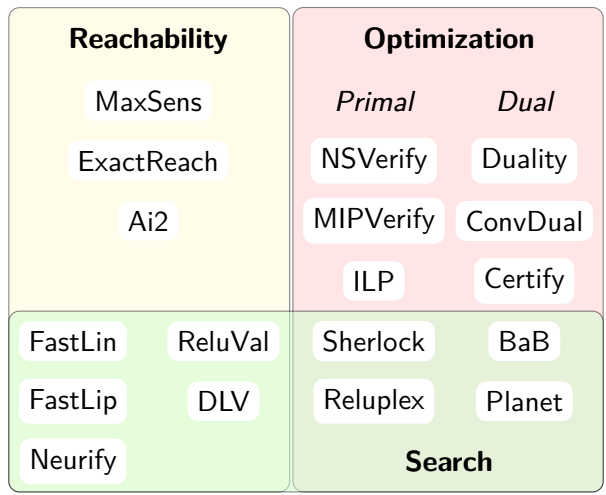

**Figure 3.1:** Overview of all methods in the survey. Given the three basic methods: reachability, optimization, and search, existing methods are divided into five categories: reachability, primal optimization, dual optimization, search and reachability, and search and optimization. Relationships among different methods are shown in the figure.

## 3.1 Reachability

These methods perform layer-by-layer reachability analysis to compute the reachable set $\mathcal{R}(\mathcal{X}, \mathbf{f})$. They usually generate `ReachabilityResult`.

ExactReach [72] computes the exact reachable set for networks with only ReLU activations. The key insight is that if the input set to a ReLU function is a union of polytopes, then the output reachable set is also a union of polytopes. As there is no over-approximation, this method is sound and complete. However, because the number of polytopes grows exponentially with each layer, this method does not scale.

Ai2 [25] uses representations that over-approximate the reachable set. It trades precision of the reachable set for scalability of the algorithm. It works for any piecewise linear activation functions, such as ReLU and max pooling. Due to its approximation, the number of geometric objects to be traced during layer-by-layer analysis is greatly reduced. Though Ai2 is not complete, it scales well.

MaxSens [69] partitions the input domain into small grid cells, and loosely approximates the reachable set for each grid cell considering the maximum sensitivity of the network at each grid cell. Sensitivity of a function is equivalent to the Lipschitz constant of the function. The union of those reachable sets is the output reachable set. The finer the partition, the tighter the output reachable set. MaxSens is not complete. It works for any activation function and its runtime scales well with the number of layers but poorly with the input dimension.

## 3.2 Primal Optimization

Primal optimization methods try to falsify assertion (2.4). The network structure is a constraint to be considered in the optimization. Existing methods only work for ReLU activations. Different methods are developed to encode the network as a set of linear constraints or mixed integer linear constraints by exploiting the piecewise linearity in the ReLU. The encoding methods will be discussed in section 6.1. The return type can be either `CounterExampleResult` or `AdversarialResult`, depending on the objective of the optimization.

NSVerify [2, 40] encodes the network as a set of mixed integer linear constraints. It solves a feasibility problem without an objective function. It tries to find a counter example for the verification problem.[1] This method is sound and complete.

MIPVerify [59] also encodes the network as a set of mixed integer linear constraints. There are two differences between MIPVerify and NSVerify. First, MIPVerify determines the bounds on the nodes to tighten the constraints. Second, MIPVerify solves an adversarial problem that tries to estimate the maximum allowable disturbance on the input side. This method is also complete.

ILP (iterative linear programming) [8] encodes the network as a set of linear constraints by linearizing the network at a reference point. The optimization problem in ILP is an adversarial problem that tries to estimate the maximum allowable disturbance on the input side. It iteratively solves the optimization. This method is not complete as it only considers one linear segment of the network.

## 3.3 Dual Optimization

In primal optimization methods, different methods are developed to encode the constraints imposed by the network. We can also use dual optimization to simplify the constraints. In dual optimization, the constraints are much simpler than those in primal optimization. On the other hand, objectives in dual optimization, which correspond to the constraints in primal optimization, are much more complicated than those in the primal problem. Relaxations are usually involved during the construction of the dual problem. Due to relaxation, these approaches are incomplete. The return type is `BasicResult`.

---

[1] NSVerify [2] is developed for verification of a closed-loop system that has neural network components. It has been extended to verify recurrent neural networks [1]. We only review the method used to verify non-recurrent neural networks, which was first discussed in [40].

Duality [20] solves a Lagrangian relaxation of the optimization problem to obtain bounds on the output. The dual problem is formulated as an unconstrained convex optimization problem, which can be computed efficiently. Duality works for any activation function.

ConvDual [68] also uses a dual approach to estimate the bounds on the output. It obtains a simplified dual problem by first making a convex relaxation of the network in the primal optimization. The bounds are heuristically computed by choosing a fixed, dual feasible solution, without any explicit optimization. In this way, ConvDual is more computationally efficient than Duality. In the original ConvDual approach, the bounds are then used to robustly train the network. This survey focuses on the method to compute the bounds.

Certify [50] uses a semidefinite relaxation to compute over approximated certificates (*i.e.*, bounds). It only works for neural networks with only one hidden layer. It works for any activation function that is differentiable almost everywhere.[2] In the original Certify approach, the certificates are then optimized jointly with network parameters to provide an adaptive regularizer that improves robustness of the network. This survey focuses on the method to obtain the certificates.

## 3.4 Search and Reachability

Reachability methods need to balance computational efficiency and precision of the approximation. When reachability is combined with search, it is possible to improve both efficiency and accuracy. These methods usually search in the input or the hidden spaces for a counter example. However, due to over-approximation in reachability analysis, these methods are sound but may be incomplete.

ReluVal [65] uses symbolic interval analysis along with search to minimize over-approximations of the output sets. During the search process, ReluVal iteratively bisects its input range. This process is called iterative interval refinement, which is also used in BaB [12].

Neurify [64] improves ReluVal by two major modifications. First, it introduces symbolic linear relaxation which is tighter than symbolic interval analysis in computing the reachable sets. Second, it introduces direct constraint refinement to perform the search more efficiently.

Fast-Lin [66] computes a certified lower bound on the allowable input disturbance for ReLU networks using a layer-by-layer approach and binary search in the input domain.

Fast-Lip [66] depends on Fast-Lin to compute the bounds on the activation functions, and further estimates the local Lipchitz constant of the network. In general, Fast-Lin is more scalable than Fast-Lip, while Fast-Lip provides better solutions for $\ell_1$ bounds.

DLV [30] searches for adversarial inputs layer by layer in the hidden layers. This is the only approach that searches in the hidden spaces we have seen so far. DLV supports any activation function.

---

[2]Differentiable a.e. means that the function is differentiable everywhere except for countably many points. Piecewise linear activation functions are all differentiable almost everywhere.

## 3.5 Search and Optimization

Search can also be combined with optimization. We can either search in the input space, or search in the function space. Searching in the function space is done by exploring possible activation patterns. An activation pattern is an assignment (*e.g.*, on or off for ReLU) to each activation function in the network. These methods may use SAT or SMT solvers.

Sherlock [18] estimates the output range using a combination of local search and global search. Local search solves a linear program to find local optima. Global search solves a mixed integer linear program to escape local optima, which is similar to the method in NSVerify and MIPVerify. Sherlock is incomplete.

BaB [12] uses branch and bound to compute the output bounds of a network. It has a modularized design that can serve as a unified framework that can support other methods such as Reluplex and Planet.

Planet [22] integrates with a SAT solver for tree search in the function space. The objective of the search is to find an activation pattern of ReLU networks that maps an input in $\mathcal{X}$ to an output not in $\mathcal{Y}$. It combines optimization-based filtering and pruning in the search process. Planet is complete.

Reluplex [32] performs tree search in the function space. It extends the simplex algorithm, a standard algorithm for solving linear programming (LP) instances, to support ReLU networks. The algorithm is called Reluplex, for ReLU with the simplex method. Reluplex is complete.

## 4 Preliminaries

This chapter introduces additional notation and operations that will be used in different verification algorithms to be discussed in the following sections. Section 4.1 discusses interval arithmetic to compute node-wise bounds given an input set. Such node-wise bounds are needed in many methods, such as MIPVerify, Duality, ConvDual, Planet, and Reluplex. Section 4.2 discusses set split, which includes input interval refinement that is used in ReluVal and BaB as well as constraint refinement on hidden nodes that is used in ExactReach, Ai2, and Neurify. Section 4.3 discusses methods to compute network gradient and bounds on the gradient given a non-trivial input set. The bounds on the gradient are used in ReluVal, Neurify, and FastLin. Section 4.4 introduces specific notation for ReLU activations.

We use $[a]_+ \coloneqq \max\{a, 0\}$ and $[a]_- = \min\{a, 0\}$ to represent the positive and negative parts of a scalar variable $a$. For a vector $\mathbf{a}$ or a matrix $\mathbf{A}$, $[\cdot]_+$ and $[\cdot]_-$ take element-wise max and min, respectively.

## 4.1 Bounds

The lower and upper bounds on $z_{i,j}$, *i.e.*, the value of node $j$ at layer $i$ after activation, are denoted $\ell_{i,j}$ and $u_{i,j}$. The lower and upper bounds on $\hat{z}_{i,j}$, *i.e.*, the value of node $j$ at layer $i$ before activation, are denoted $\hat{\ell}_{i,j}$ and $\hat{u}_{i,j}$. The after-activation bounds for the whole layer $i$ are denoted $\boldsymbol{\ell}_i$ and $\mathbf{u}_i$. The before-activation bounds for the whole layer $i$ are denoted $\hat{\boldsymbol{\ell}}_i$ and $\hat{\mathbf{u}}_i$. The bounds can be computed using different methods. For example, MaxSens uses interval arithmetic, which will be discussed in section 5.4. Planet solves a primal optimization problem that adopts triangle relaxation to tighten the bounds from interval arithmetic, which will be discussed in section 9.3. ConvDual and FastLin are able to use dynamic programming to analytically solve a dual optimization problem that uses parallel relaxation which is looser than triangle relaxation, which will be discussed in section 7.3 and section 8.3, respectively. Both triangle and parallel relaxation will be discussed in section 6.1. Here we introduce interval arithmetic.

**Interval arithmetic** By interval arithmetic, given the bounds at layer $i-1$, the bounds at layer $i$ satisfy

$$\hat{\ell}_{i,j} = \min_{\mathbf{z}_{i-1} \in [\boldsymbol{\ell}_{i-1}, \mathbf{u}_{i-1}]} \mathbf{w}_{i,j} \mathbf{z}_{i-1} + b_{i,j} = [\mathbf{w}_{i,j}]_+ \boldsymbol{\ell}_{i-1} + [\mathbf{w}_{i,j}]_- \mathbf{u}_{i-1} + b_{i,j}, \tag{4.1a}$$

$$\hat{u}_{i,j} = \max_{\mathbf{z}_{i-1} \in [\boldsymbol{\ell}_{i-1}, \mathbf{u}_{i-1}]} \mathbf{w}_{i,j} \mathbf{z}_{i-1} + b_{i,j} = [\mathbf{w}_{i,j}]_+ \mathbf{u}_{i-1} + [\mathbf{w}_{i,j}]_- \boldsymbol{\ell}_{i-1} + b_{i,j}, \tag{4.1b}$$

$$\ell_{i,j} = \min_{\hat{z}_{i,j} \in [\hat{\ell}_{i,j}, \hat{u}_{i,j}]} \sigma_{i,j}(\hat{z}_{i,j}) = \sigma_{i,j}(\hat{\ell}_{i,j}), \tag{4.1c}$$

$$u_{i,j} = \max_{\hat{z}_{i,j} \in [\hat{\ell}_{i,j}, \hat{u}_{i,j}]} \sigma_{i,j}(\hat{z}_{i,j}) = \sigma_{i,j}(\hat{u}_{i,j}), \tag{4.1d}$$

where the implicit assumption for the last two equalities is that the activation $\sigma_{i,j}$ is non-decreasing. In the implementation, the bounds are usually stored as a list of hyperrectangles. The bounds with respect to the input constraint $\mathcal{X}$ can be computed layer-by-layer using interval arithmetic (4.1) in algorithm 5.

In the following discussion, we write interval arithmetic with respect to linear mappings compactly as $\otimes$ where

$$\mathbf{W} \otimes [\boldsymbol{\ell}, \mathbf{u}] := \left[[\mathbf{W}]_+ \boldsymbol{\ell} + [\mathbf{W}]_- \mathbf{u}, [\mathbf{W}]_+ \mathbf{u} + [\mathbf{W}]_- \boldsymbol{\ell}\right], \tag{4.2}$$

where $\boldsymbol{\ell}$ and $\mathbf{u}$ can also be replaced with matrices. The function (4.2) is implemented in algorithm 6. Hence, $[\hat{\boldsymbol{\ell}}_i, \hat{\mathbf{u}}_i] = \mathbf{W}_i \otimes [\boldsymbol{\ell}_{i-1}, \mathbf{u}_{i-1}] + [\mathbf{b}_i, \mathbf{b}_i]$.

```
function get_bounds(nnet::Network, input, act::Bool = true)
    input = overapproximate(input)
    bounds = Vector{Hyperrectangle}(undef, length(nnet.layers) + 1)
    bounds[1] = input
    b = input
    for (i, layer) in enumerate(nnet.layers)
        if act
            b = approximate_affine_map(layer, bounds[i])
            bounds[i+1] = approximate_act_map(layer, b)
        else
            bounds[i+1] = approximate_affine_map(layer, b)
            b = approximate_act_map(layer, bounds[i+1])
        end
    end
    return bounds
end
```

**Algorithm 5:** Function to compute node-wise bounds. When the boolean `act` is set to `true`, the function outputs the after activation bounds $\ell_i$ and $\mathbf{u}_i$ for all $i$. Otherwise, the function outputs the before activation bounds $\hat{\ell}_i$ and $\hat{\mathbf{u}}_i$ for all $i$. The function `approximate_act_map` solves (4.1a) and (4.1b), while the function `approximate_act_map` solves (4.1d) and (4.1d).

```
function interval_map(W, l, u)
    l_new = max.(W, 0) * l + min.(W, 0) * u
    u_new = max.(W, 0) * u + min.(W, 0) * l
    return (l_new, u_new)
end
```

**Algorithm 6:** A function to compute linear mapping on intervals, corresponding to equation (4.2).

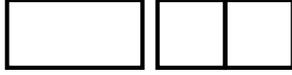

**Figure 4.1:** Illustration of interval refinement. The high dimensional interval on the left is split into two high dimensional intervals on the right, with respect to the horizontal axis.

**Bounds for polytopes and star sets** In the interval arithmetic, we essentially compute the bounds assuming that all reachable sets are encoded as intervals. In certain cases, we may need to compute the bounds for a node $z_{i,j}$ whose reachable set is encoded in a polytope or a star set. Suppose the reachable set for the $i$th layer satisfies $\mathbf{Cz}_i \leq \mathbf{d}$, then the bounds $\ell_{i,j}$ and $u_{i,j}$ for the $j$th node can be computed by solving the following linear program:

$$\ell_{i,j} = \min_{\mathbf{z}_i} z_{i,j} \quad \text{s.t. } \mathbf{Cz}_i \leq \mathbf{d}, \tag{4.3a}$$

$$u_{i,j} = \max_{\mathbf{z}_i} z_{i,j} \quad \text{s.t. } \mathbf{Cz}_i \leq \mathbf{d}. \tag{4.3b}$$

Suppose the reachable set for the node satisfies $z_{i,j} = c + \mathbf{g}^\mathsf{T} \boldsymbol{\alpha}$ with $\mathbf{C}\boldsymbol{\alpha} \leq \mathbf{d}$, then the bounds $\ell_{i,j}$ and $u_{i,j}$ can be computed by solving the following linear program:

$$\ell_{i,j} = \min_{\boldsymbol{\alpha}} c + \mathbf{g}^\mathsf{T} \boldsymbol{\alpha} \quad \text{s.t. } \mathbf{C}\boldsymbol{\alpha} \leq \mathbf{d}, \tag{4.4a}$$

$$u_{i,j} = \max_{\boldsymbol{\alpha}} c + \mathbf{g}^\mathsf{T} \boldsymbol{\alpha} \quad \text{s.t. } \mathbf{C}\boldsymbol{\alpha} \leq \mathbf{d}. \tag{4.4b}$$

We define two functions $\underline{h}$ and $\bar{h}$ that map either a polytope or a star set to its lower and upper bounds in (4.3) and (4.4). These two functions are used in ReluVal and Neurify.

## 4.2 Set Split

Set split is used to minimize over approximation during the verification process. In general, over approximation is introduced by different verification algorithms when the neurons operate in their nonlinear regions, i.e., when there are ambiguities in the activation status of neurons. A smaller input set will lead to less ambiguity, hence reduce over approximation. The amount of reduction has been formally quantified [3]. In practice, either the input set or the reachable set in hidden nodes can be split. Interval refinement is usually used to split input sets into equal halves, which is used in ReluVal and BaB. Constraint refinement is usually used to split reachable sets for hidden nodes based on their activation status, which is used in ExactReach, Ai2, and Neurify. In either case, set split will generate additional geometric objects that belong to the same type as the original set.

**Interval refinement** Interval refinement only operates on hyperrectangles or intevals. A high dimensional interval $[\boldsymbol{\ell}, \mathbf{u}]$ is equally split in two intervals at index $i^*$:

$$[\boldsymbol{\ell}, \mathbf{u}^*] \text{ and } [\boldsymbol{\ell}^*, \mathbf{u}], \tag{4.5}$$

where $\mathbf{u}^* = \mathbf{u} - r_{i^*}\mathbf{e}_{i^*}$, $\boldsymbol{\ell}^* = \boldsymbol{\ell} + r_{i^*}\mathbf{e}_{i^*}$, $r_{i^*} = \frac{1}{2}(u_{i^*} - \ell_{i^*})$, and $\mathbf{e}_i$ is a unit vector whose entries are all zero except at the $i$th entry. The index $i^*$ is determined using different methods. For example, in BaB, the index $i^*$ is chosen to be the longest dimension, *i.e.*, $i^* = \arg\max_i(u_i - l_i)$. The function to perform the split is implemented in algorithm 7. Figure 4.1 illustrates the split of a two dimensional interval.

**Constraint refinement** Constraint refinement can operate on any geometric object. It adds a set of complementary halfspace constraints to the original set. Suppose the constraints to be added are $\mathbf{a}^\mathsf{T}\mathbf{z} \leq b$ and $\mathbf{a}^\mathsf{T}\mathbf{z} \geq b$.[1] A halfspace-polytope $\mathbf{Cz} \leq \mathbf{d}$ can be split into two halfspace-polytopes

$$\begin{bmatrix} \mathbf{C} \\ \mathbf{a}^\mathsf{T} \end{bmatrix} \mathbf{z} \leq \begin{bmatrix} \mathbf{d} \\ b \end{bmatrix} \text{ and } \begin{bmatrix} \mathbf{C} \\ -\mathbf{a}^\mathsf{T} \end{bmatrix} \mathbf{z} \leq \begin{bmatrix} \mathbf{d} \\ -b \end{bmatrix}. \tag{4.6}$$

A star set $\mathbf{x} = \mathbf{x}_0 + \mathbf{R}\boldsymbol{\alpha}$ where $\mathbf{C}\boldsymbol{\alpha} \leq \mathbf{d}$ can be split into two star sets

$$\mathbf{x} = \mathbf{x}_0 + \mathbf{R}\boldsymbol{\alpha}, \begin{bmatrix} \mathbf{C} \\ \mathbf{a}^\mathsf{T}\mathbf{R} \end{bmatrix} \boldsymbol{\alpha} \leq \begin{bmatrix} \mathbf{d} \\ b - \mathbf{a}^\mathsf{T}\mathbf{x}_0 \end{bmatrix}, \tag{4.7a}$$

$$\mathbf{x} = \mathbf{x}_0 + \mathbf{R}\boldsymbol{\alpha}, \begin{bmatrix} \mathbf{C} \\ -\mathbf{a}^\mathsf{T}\mathbf{R} \end{bmatrix} \boldsymbol{\alpha} \leq \begin{bmatrix} \mathbf{d} \\ -b + \mathbf{a}^\mathsf{T}\mathbf{x}_0 \end{bmatrix}. \tag{4.7b}$$

```
function split_interval(dom::Hyperrectangle, i::Int64)
    input_lower, input_upper = low(dom), high(dom)
    input_upper[i] = dom.center[i]
    input_split_left = Hyperrectangle(low = input_lower, high = input_upper)
    input_lower[i] = dom.center[i]
    input_upper[i] = dom.center[i] + dom.radius[i]
    input_split_right = Hyperrectangle(low = input_lower, high = input_upper)
    return (input_split_left, input_split_right)
end
```

**Algorithm 7:** Function to split interval at a specified index. The argument `dom` is the interval to be split. The argument `index` is the index $i^*$ where the interval should be split at. The function returns the two hyperrectangles after the split, which correspond to equation (4.5).

---

[1]The constraints may represent the activation status of a particular node if the activation function is ReLU. For example, $\mathbf{c}^\mathsf{T}\mathbf{z} \geq d$ corresponds to the case that $\boldsymbol{\sigma}_i(\mathbf{Wz} + \mathbf{b})$ is activated and $\mathbf{c}^\mathsf{T}\mathbf{z} \leq d$ corresponds to the other case.

## 4.3 Gradient

The gradient of a neural network (of the output with respect to the input) satisfies the chain rule,

$$\nabla \mathbf{f} := \frac{\partial \mathbf{y}}{\partial \mathbf{x}} = \frac{\partial \mathbf{y}}{\partial \hat{\mathbf{z}}_n} \frac{\partial \hat{\mathbf{z}}_n}{\partial \mathbf{z}_{n-1}} \cdots \frac{\partial \mathbf{z}_1}{\partial \hat{\mathbf{z}}_1} \frac{\partial \hat{\mathbf{z}}_1}{\partial \mathbf{x}} = \nabla \boldsymbol{\sigma}_n \mathbf{W}_n \cdots \nabla \boldsymbol{\sigma}_1 \mathbf{W}_1, \qquad (4.8)$$

where $\nabla \boldsymbol{\sigma}_i := \frac{\partial \mathbf{z}_i}{\partial \hat{\mathbf{z}}_i} \in \mathbb{R}^{k_i \times k_i}$.

In some cases, we evaluate the point-wise gradient $\nabla \mathbf{f}(\mathbf{x}_0)$ for some $\mathbf{x}_0$. The point-wise gradient is easy to compute by following the chain rule as shown in the first function in algorithm 8. In other cases, *e.g.*, in ReluVal and FastLip, we need to compute the maximum gradient given an input set $\mathcal{X}$, *i.e.*, $\max_{\mathbf{x} \in \mathcal{X}} \nabla \mathbf{f}(\mathbf{x})$. The maximum over a vector is taken point-wise. The maximum gradient can be computed using interval arithmetic.

Denote the lower and upper bounds of $\nabla \boldsymbol{\sigma}_i$ with respect to the input set $\mathcal{X}$ as $\underline{\boldsymbol{\Lambda}}_i \in \mathbb{R}^{k_i \times k_i}$ and $\overline{\boldsymbol{\Lambda}}_i \in \mathbb{R}^{k_i \times k_i}$. The matrices $\underline{\boldsymbol{\Lambda}}_i$ and $\overline{\boldsymbol{\Lambda}}_i$ are diagonal. Due to the monotonicity assumption on $\boldsymbol{\sigma}_i$,

$$\overline{\boldsymbol{\Lambda}}_i \geq \underline{\boldsymbol{\Lambda}}_i \geq \mathbf{0}, \qquad (4.9)$$

where the inequalities are interpreted point-wise.

Define $\mathbf{G}_i := \frac{\partial \mathbf{z}_i}{\partial \mathbf{x}}$ and $\hat{\mathbf{G}}_i := \mathbf{W}_i \mathbf{G}_{i-1}$. Denote the lower and upper bounds of $\mathbf{G}_i$ as $\underline{\mathbf{G}}_i, \overline{\mathbf{G}}_i \in \mathbb{R}^{k_i \times k_0}$, and the lower and upper bounds of $\hat{\mathbf{G}}_i$ as $\underline{\hat{\mathbf{G}}}_i, \overline{\hat{\mathbf{G}}}_i \in \mathbb{R}^{k_i \times k_0}$. The bounds on the gradients are initialized as $\underline{\mathbf{G}}_0 = \overline{\mathbf{G}}_0 = \mathbf{I}$, and can be updated inductively using interval arithmetic by forward propagation,[2]

$$\mathbf{G}_i = \nabla \boldsymbol{\sigma}_i \hat{\mathbf{G}}_i = \nabla \boldsymbol{\sigma}_i \mathbf{W}_i \mathbf{G}_{i-1}. \qquad (4.10)$$

Given $\underline{\mathbf{G}}_{i-1}$ and $\overline{\mathbf{G}}_{i-1}$,

$$[\underline{\hat{\mathbf{G}}}_i, \overline{\hat{\mathbf{G}}}_i] = \mathbf{W}_i \otimes [\underline{\mathbf{G}}_{i-1}, \overline{\mathbf{G}}_{i-1}]. \qquad (4.11)$$

The lower bound $\underline{\mathbf{G}}_i$ on the gradient $\mathbf{G}_i$ is the minimum of $\nabla \boldsymbol{\sigma}_i \hat{\mathbf{G}}_i$, where $\nabla \boldsymbol{\sigma}_i \in [\underline{\boldsymbol{\Lambda}}_i, \overline{\boldsymbol{\Lambda}}_i]$ and $\hat{\mathbf{G}}_i \in [\underline{\hat{\mathbf{G}}}_i, \overline{\hat{\mathbf{G}}}_i]$. Since $\nabla \boldsymbol{\sigma}_i \geq \mathbf{0}$ according to (4.9), the minimum of $\nabla \boldsymbol{\sigma}_i \hat{\mathbf{G}}_i$ is achieved on the lower bound of $\hat{\mathbf{G}}_i$. Hence,

$$\underline{\mathbf{G}}_i = \min\{\underline{\boldsymbol{\Lambda}}_i \underline{\hat{\mathbf{G}}}_i, \overline{\boldsymbol{\Lambda}}_i \underline{\hat{\mathbf{G}}}_i\} = \underline{\boldsymbol{\Lambda}}_i [\underline{\hat{\mathbf{G}}}_i]_+ + \overline{\boldsymbol{\Lambda}}_i [\underline{\hat{\mathbf{G}}}_i]_-. \qquad (4.12)$$

Similarly, the upper bound on the gradient $\frac{\partial \mathbf{z}_i}{\partial \mathbf{x}}$ satisfies

$$\overline{\mathbf{G}}_i = \underline{\boldsymbol{\Lambda}}_i [\overline{\hat{\mathbf{G}}}_i]_- + \overline{\boldsymbol{\Lambda}}_i [\overline{\hat{\mathbf{G}}}_i]_+. \qquad (4.13)$$

---

[2] We can also use backward propagation in the chain rule to compute $\nabla \mathbf{f}$. The update equation becomes $\frac{\partial \mathbf{y}}{\partial \mathbf{z}_i} = \frac{\partial \mathbf{y}}{\partial \mathbf{z}_{i+1}} \nabla \boldsymbol{\sigma}_{i+1} \mathbf{W}_{i+1}$.

The function to compute the bounds on the gradient given a non-trivial input set is implemented in algorithm 8. There is a solver, called RecurJac [76], that can efficiently compute the maximum gradient in a recursive manner.[3]

```
function get_gradient(nnet::Network, x::Vector)
    z = x
    gradient = Matrix(1.0I, length(x), length(x))
    for (i, layer) in enumerate(nnet.layers)
        z_hat = affine_map(layer, z)
        σ_gradient = act_gradient(layer.activation, z_hat)
        gradient = Diagonal(σ_gradient) * layer.weights * gradient
        z = layer.activation(z_hat)
    end
    return gradient
end

function get_gradient(nnet::Network, input::AbstractPolytope)
    LΛ, UΛ = act_gradient_bounds(nnet, input)
    return get_gradient(nnet, LΛ, UΛ)
end

function get_gradient(nnet::Network, LΛ::Vector{Matrix}, UΛ::Vector{Matrix})
    n_input = size(nnet.layers[1].weights, 2)
    LG = Matrix(1.0I, n_input, n_input)
    UG = Matrix(1.0I, n_input, n_input)
    for (i, layer) in enumerate(nnet.layers)
        LG_hat, UG_hat = interval_map(layer.weights, LG, UG)
        LG = LΛ[i] * max.(LG_hat, 0) + UΛ[i] * min.(LG_hat, 0)
        UG = LΛ[i] * min.(UG_hat, 0) + UΛ[i] * max.(UG_hat, 0)
    end
    return (LG, UG)
end
```

**Algorithm 8:** Functions to compute gradient. The first function computes point-wise gradient, where `act_gradient` (not shown) computes the gradient of the activation function. The second and third functions compute the bounds on the gradient given a non-trivial input set. The input to the second function is the input set. It calls `act_gradient_bounds` (not shown) to compute $\underline{\mathbf{\Lambda}}_i$ and $\overline{\mathbf{\Lambda}}_i$. The third function directly takes the bounds on the gradient of activation functions.

## 4.4 ReLU Activation

If $\boldsymbol{\sigma}_i$ is a ReLU activation function, *i.e.*, $\boldsymbol{\sigma}_i(\hat{\mathbf{z}}_i) = [\mathbf{z}_i]_+$, we associate a binary vector $\boldsymbol{\delta}_i \in \{0,1\}^{k_i}$ for $i \in \{1,\ldots,n\}$ to specify activation status (off or on) of the nodes. At layer $i$, given the bounds on the values of the nodes, we define the set of nodes that are activated

---
[3]https://github.com/huanzhang12/RecurJac-Jacobian-Bounds

$\Gamma_i^+$, not activated $\Gamma_i^-$, and undetermined $\Gamma_i$:

$$\Gamma_i^+ = \{j : \hat{\ell}_{i,j} \geq 0\}, \tag{4.14a}$$
$$\Gamma_i^- = \{j : \hat{u}_{i,j} \leq 0\}, \tag{4.14b}$$
$$\Gamma_i = \{j : \hat{\ell}_{i,j} < 0 < \hat{u}_{i,j}\}. \tag{4.14c}$$

The diagonal entries $\underline{\lambda}_{i,j}$ and $\overline{\lambda}_{i,j}$ of the bounds $\underline{\mathbf{\Lambda}}_i, \overline{\mathbf{\Lambda}}_i \in \mathbb{R}^{k_i \times k_i}$ on the gradient $\nabla \boldsymbol{\sigma}_i$ satisfy

$$\overline{\lambda}_{i,j} = \begin{cases} 1 & j \in \Gamma_i^+ \\ 0 & j \in \Gamma_i^- \\ 1 & j \in \Gamma_i \end{cases}, \underline{\lambda}_{i,j} = \begin{cases} 1 & j \in \Gamma_i^+ \\ 0 & j \in \Gamma_i^- \\ 0 & j \in \Gamma_i \end{cases}. \tag{4.15}$$

We also define a relaxed gradient $\tilde{\mathbf{\Lambda}}_i \in \mathbb{R}^{k_i \times k_i}$ for $\nabla \boldsymbol{\sigma}_i$. The diagonal entries $\tilde{\lambda}_{i,j}$ of $\tilde{\mathbf{\Lambda}}_i$ are

$$\tilde{\lambda}_{i,j} = \begin{cases} 1 & j \in \Gamma_i^+ \\ 0 & j \in \Gamma_i^- \\ \frac{\hat{u}_{i,j}}{\hat{u}_{i,j} - \hat{\ell}_{i,j}} & j \in \Gamma_i \end{cases}. \tag{4.16}$$

The gradient in the case $j \in \Gamma_i$ uses triangle relaxation as shown in figure 6.4. The triangle relaxation will be discussed in section 6.1. The relaxed gradient is compliant with the bounds, *i.e.*, $\underline{\mathbf{\Lambda}}_i \leq \tilde{\mathbf{\Lambda}}_i \leq \overline{\mathbf{\Lambda}}_i$.

# 5 Reachability

Reachability methods compute the output reachable set $\mathcal{R}(\mathcal{X}, \mathbf{f})$ to verify the problem through layer-by-layer analysis. This chapter first reviews the general methodology, and then discusses the specific methods.

## 5.1 Overview

The layer-by-layer propagation in reachability methods is illustrated in figure 5.1 and implemented in algorithm 9. The function `forward_network` computes $\mathcal{R}$. It calls `forward_layer` to perform layer-by-layer forward propagation. Once it computes the reachable set, `check_inclusion` verifies whether $\mathcal{R}(\mathcal{X}, \mathbf{f}) \subset \mathcal{Y}$.

The function `forward_layer` involves the mapping from $\mathbf{z}_{i-1}$ to $\mathbf{z}_i$, *i.e.*, $\mathbf{z}_{i-1} \mapsto \boldsymbol{\sigma}_i(\mathbf{W}_i \mathbf{z}_{i-1} + \mathbf{b}_i)$. The linear mapping defined by $\mathbf{z}_{i-1} \mapsto \mathbf{W}_i \mathbf{z}_{i-1} + \mathbf{b}_i$ is relatively easy to handle. The nonlinear mapping $\hat{\mathbf{z}}_i \mapsto \boldsymbol{\sigma}_i(\hat{\mathbf{z}}_i)$ is non-trivial. Different methods introduce different ways to handle nonlinear mappings. Hence, the implementation of `forward_layer` varies across different methods. The methods can be divided into two categories: exact reachability and over approximation.

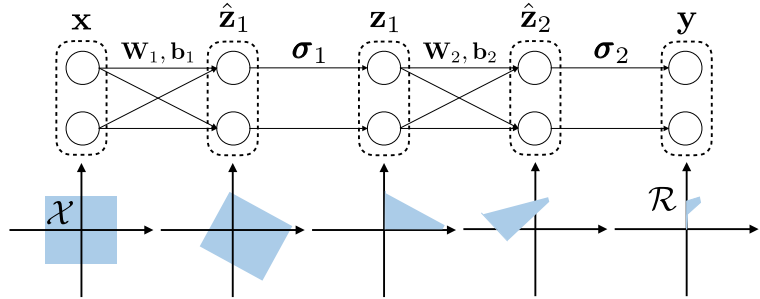

**Figure 5.1:** Illustration of reachability methods. The network in the illustration only contains one hidden layer. The input set $\mathcal{X}$ is first passed through the linear mapping defined by $\mathbf{W}_1$ and $\mathbf{b}_1$. Then it goes through the nonlinear mapping defined by $\boldsymbol{\sigma}_1$ (ReLU is considered). The corresponding reachable sets are illustrated in the shaded area. The process is repeated for the next layer and the output reachable set is then obtained.

```
function solve(solver, problem::Problem)
    reach = forward_network(solver, problem.network, problem.input)
    return check_inclusion(reach, problem.output)
end

function forward_network(solver, nnet::Network, input::AbstractPolytope)
    reach = input
    for layer in nnet.layers
        reach = forward_layer(solver, layer, reach)
    end
    return reach
end

function check_inclusion(reach::Vector{<:AbstractPolytope}, output)
    for poly in reach
        issubset(poly, output) || return ReachabilityResult(:violated, reach)
    end
    return ReachabilityResult(:holds, similar(reach, 0))
end
```

**Algorithm 9:** General structure of reachability methods. The problem is solved by first performing layer-by-layer reachability analysis as specified in `forward_network`, then checking if the output reachable set belongs to the output constraint using `check_inclusion`. Different methods have different implementations of `forward_layer`. The outer loop `solve` can also vary between methods.

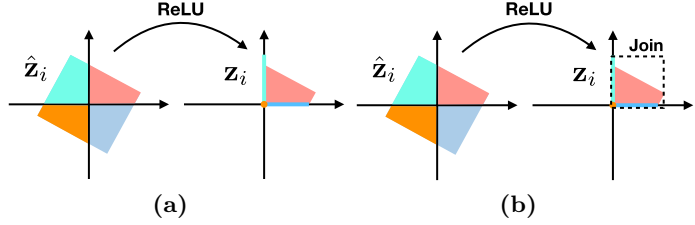

**Figure 5.2:** Illustration of different approaches to handle the nonlinear mapping $\hat{\mathbf{z}}_i \mapsto \boldsymbol{\sigma}_i(\hat{\mathbf{z}}_i)$. (a) Exact reachability. (b) Split-and-join.

**Exact reachability** *Exact reachability* works for piecewise linear networks and computes the reachable set for every linear segment of the network and keeps track of all sets. This is done by ExactReach [72]. During the reachability analysis, constraint refinement[1] is needed when a set operates on the nonlinear region of the network, *i.e.*, the set needs to be split into different pieces that operate only on linear segments of the network. However, the number of linear segments of a network grows exponentially as the number of nodes in one layer and the number of layers increases, making exact reachability computationally expensive.

**Over approximation** *Over approximation* works for any nonlinear activation function and over approximates the reachable set. Since over approximation methods do not keep track of the reachable set for every linear segment of the network, they are more computationally tractable. Nonetheless, there is a trade-off between the amount of over approximation introduced by a verification algorithm and its computational efficiency. There are different approaches to perform the over approximation. Considering the nonlinear mapping $\hat{\mathbf{z}}_i \mapsto \boldsymbol{\sigma}_i(\hat{\mathbf{z}}_i)$ over a set $\hat{\mathcal{Z}}_i$, we can either approximate the reachable set after applying the nonlinear mapping (called *set over approximation*) or approximate the nonlinear mapping directly then propagate the set through the approximated nonlinear mapping (called *function over approximation*).

**Set over approximation** For set over approximation, we approximate the reachable area $\boldsymbol{\sigma}_i(\hat{\mathcal{Z}}_i)$ using a predefined geometric template, *e.g.*, polytope or interval. Representative methods are:

- *Split-and-join* (for piecewise linear networks), which computes the reachable set for every linear segment of the network and joins those sets by over-approximation. This is done by Ai2 [25].

---

[1]Constraint refinement is discussed in section 4.2.

- *Interval arithmetic* (for networks with monotone activation functions), which computes the bounds for each node separately by interval arithmetic. This is done by MaxSens [69].

Figure 5.2 illustrates the difference between exact reachability and split-and-join in the case that the activation function is ReLU. Two nodes are considered, *i.e.*, $\mathbf{z}_i \in \mathbb{R}^2$. Hence, there are four piecewise linear components in the nonlinear mapping $\hat{\mathbf{z}}_i \mapsto \boldsymbol{\sigma}_i(\hat{\mathbf{z}}_i)$, which correspond to the four quadrants shown in the left plots of figure 5.2. Under ReLU, the set in the first quadrant (*i.e.*, both values are greater than zero) is kept the same after the mapping. The sets in the second and the fourth quadrants (*i.e.*, only one value is greater than zero) are mapped to line segments. The set in the third quadrant (*i.e.*, both values are smaller than zero) is mapped to the origin. Exact reachability keeps track of all geometric objects after the mapping. In figure 5.2a, one input set generates four geometric objects to represent the reachable set. The number of geometric objects grows exponentially with the number of nodes at each layer. On the other hand, to improve scalability of the method, split-and-join methods merge all geometric objects together as shown in figure 5.2b. Mathematical derivations will be introduced in section 5.2 for exact reachability and in section 5.3 for split-and-join.

**Function over approximation** For function over approximation, we first relax the function $\boldsymbol{\sigma}_i$ to its lower bound $\underline{\boldsymbol{\sigma}}_i$ and upper bound $\bar{\boldsymbol{\sigma}}_i$. The lower and upper bounds should be linear functions that satisfy $\underline{\boldsymbol{\sigma}}_i(\hat{\mathbf{z}}_i) \leq \boldsymbol{\sigma}_i(\hat{\mathbf{z}}_i) \leq \bar{\boldsymbol{\sigma}}_i(\hat{\mathbf{z}}_i)$.[2] Then the set $\hat{\mathcal{Z}}_i$ is propagated through the lower and upper bounds. The reachable set is then approximated by a convex hull of $\underline{\boldsymbol{\sigma}}_i(\hat{\mathcal{Z}}_i) \bigcup \bar{\boldsymbol{\sigma}}_i(\hat{\mathcal{Z}}_i)$. These operations need to be performed using symbolic representations or star sets to keep track of the dependencies between $\underline{\boldsymbol{\sigma}}_i(\hat{\mathcal{Z}}_i)$ and $\bar{\boldsymbol{\sigma}}_i(\hat{\mathcal{Z}}_i)$. A linear transformation of a star set is still a star set. Recall that a star set (2.10) has two major components: a linear relationship (called the *generator*) between the variable $\mathbf{x}$ and the symbol $\boldsymbol{\alpha}$ and a polytope (called the *domain*) that the symbol $\boldsymbol{\alpha}$ belongs to. Hence, there are two ways to compute the reachable set using function over approximation, by either symbolic propagation with fixed domain or symbolic propagation with fixed generator.

- *Symbolic propagation with fixed domain* needs to use a symbolic lower bound and a symbolic upper bound to keep track of the reachable set as a function of the input $\mathbf{x}$. The domains in the symbolic representations are always the input set $\mathcal{X}$. The nonlinear activation is relaxed to two linear functions. The symbolic lower bound propagates through $\underline{\boldsymbol{\sigma}}_i$, while the symbolic upper bound propagates through $\bar{\boldsymbol{\sigma}}_i$. During the propagation, only the generators of the symbolic bounds change. ReluVal [65] and

---

[2]It is possible to have multiple lower and upper bounds. For example, in a triangle relaxation of a ReLU activation function, we can have two lower bounds $\sigma(\hat{z}) \geq 0$ and $\sigma(\hat{z}) \geq \hat{z}$. Section 6.1 discusses several relaxation methods.

Neurify [64] take this approach, which will be discussed in chapter 8. Meanwhile, the symbolic propagation with fixed domain can also be performed by dual optimization as done in FastLin [66] for ReLU activations and CROWN [75] for general activations. In the dual optimization approach, we regard the generators as the dual variables, which can be computed using dynamic programming. The details will be discussed in chapter 7 and section 8.3.

- *Symbolic propagation with fixed generator* updates the domain, hence it only needs one geometric object instead of the symbolic lower and upper bounds. The idea is to directly add the constraints $\underline{\boldsymbol{\sigma}}_i(\hat{\mathbf{z}}_i) \leq \boldsymbol{\sigma}_i(\hat{\mathbf{z}}_i) \leq \bar{\boldsymbol{\sigma}}_i(\hat{\mathbf{z}}_i)$ to the domain. For example, let us consider one node with ReLU activation. Suppose the reachable set before activation is represented as $\hat{z} = c + \mathbf{g}^\mathsf{T}\boldsymbol{\alpha}$ with $\mathbf{C}\boldsymbol{\alpha} \leq \mathbf{d}$. Let the lower and upper bounds of the node be $\hat{\ell}$ and $\hat{u}$, which can be computed by linear programming as discussed in section 4.1. Suppose the set operates on the nonlinear region of the ReLU activation function, *i.e.*, $\hat{\ell} < 0 < \hat{u}$. The activation function $\sigma$ is relaxed to be $\sigma(\hat{z}) \geq 0$, $\sigma(\hat{z}) \geq \hat{z}$, and $\sigma(\hat{z}) \leq \tilde{\lambda}(\hat{z} - \hat{\ell})$ where $\tilde{\lambda} = \frac{\hat{u}}{\hat{u}-\hat{\ell}}$. Since the relaxed function has additional degree of freedom, *i.e.*, $\hat{z} = 0$ can be mapped to the range between 0 and $-\tilde{\lambda}\hat{\ell}$, we need to introduce a new variable $\alpha^*$ to capture this over approximation. The symbolic reachable set after the mapping $z = \sigma(\hat{z})$ is then constructed as $z = c + \mathbf{g}^\mathsf{T}\boldsymbol{\alpha} + \alpha^*$ subject to the following constraints: $\alpha^* \geq 0$, $\mathbf{g}^\mathsf{T}\boldsymbol{\alpha} + \alpha^* \geq -c$, $(1-\tilde{\lambda})\mathbf{g}^\mathsf{T}\boldsymbol{\alpha} + \alpha^* \leq (\tilde{\lambda}-1)c - \tilde{\lambda}\hat{\ell}$, and $\mathbf{C}\boldsymbol{\alpha} \leq \mathbf{d}$. The first three constraints are the constraints introduced by function over approximation, while the last constraint is inherited from the set before activation. There are different ways to relax the ReLU activation. The star set algorithm in NNV [61] uses the triangle relaxation as we did in the example; DeepZ in ERAN [54] uses zonotope relaxation; DeepPoly in ERAN [55] uses a custom polyhedral abstract domain relaxation. These methods only consider function over approximation for one scalar function at a moment, *i.e.*, node-wise approximation. To provide tight convex over approximation over multiple ReLU activation functions simultaneously, Singh *et al.* introduces k-ReLU in ERAN [53], where 1-ReLU corresponds to the triangle relaxation in the star set algorithm [61].

Exact reachability and split-and-join methods do not distinguish individual nodes, but consider all nodes at a layer as a whole. Though these methods make it easy to track dependencies between nodes, the resulting high dimensional geometric objects (usually polytopes) can be inefficient to manipulate. Interval arithmetic considers node-wise reachability by computing reachable intervals for all nodes separately as shown in figure 5.3a. Consequently, the reachable set is a hyperrectangle, which is easy to manipulate. However, as the dependencies among nodes are removed, interval arithmetic may result in significant over-approximation. Symbolic propagation then introduces symbolic intervals to track those dependencies as shown in figure 5.3b. As illustrated in figure 5.3, symbolic propagation can generate much tighter bounds.[3] Rigorous mathematical derivations will be introduced in section 5.4 for interval arithmetic and section 8.1 for symbolic propagation.

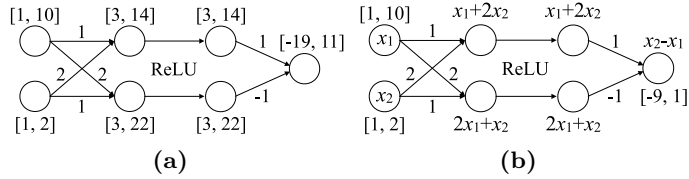

**Figure 5.3:** Illustration of the difference between interval arithmetic and symbolic propagation. (a) Interval arithmetic. (b) Symbolic propagation.

For different methods, their outer loops, *e.g.*, the `solve` function, can also be different. The simplest outer loop is shown in algorithm 9, which works for ExactReach and Ai2. In some methods, the inputs are split into smaller segments to minimize over-approximation as illustrated in figure 5.4.[4] The reachable set for each segment is computed and then joined together. MaxSens uses this approach. However, blindly splitting the input set may not be efficient. By combining with search, we can look for the most influential inputs to split as done in ReluVal. We may also use binary search to adjust the radius of the segment, which is adopted in FastLin. Instead of input interval refinement, we can perform constraint refinement by directly splitting the input set according to different activation patterns as done in Neurify.

This chapter discusses ExactReach, Ai2, and MaxSens. ReluVal, Neurify, and FastLin will be discussed in chapter 8.

## 5.2 ExactReach

ExactReach [72] performs exact reachability analysis for networks with linear or ReLU activations. For any ReLU function, if the input set is a union of polytopes, then the output reachable set is also a union of polytopes as shown in figure 5.2a. The method can be implemented with either H-polytope or star set [61]. For simplicity, we only implemented the H-polytope version where the input set and output set are both set to `HPolytope`. The function to compute the reachable set for a single layer is shown in algorithm 10, which is called in the main loop of algorithm 9.

---

[3]To tighten the bounds in interval arithmetic, in addition to doing symbolic propagation, we can also divide the input space into smaller intervals, and compute the reachable output intervals for those small intervals. This approach is used in MaxSens and ReluVal.

[4]Recall `split_interval`, first introduced in chapter 4.

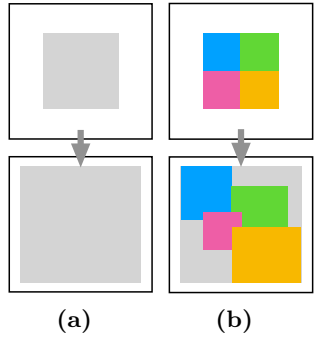

**Figure 5.4:** Splitting the input set to minimize over-approximation. The upper square is the input domain and the lower square is the output domain. (a) Over-approximated reachable set (gray square in the output domain) without partition. (b) Over-approximated reachable sets for different segments of the input set. Colors show correspondence.

**Linear Mapping**  The input set to layer $i$ consists of a list of H-polytopes. One input H-polytope parameterized by $\mathbf{C} \in \mathbb{R}^{k \times k_{i-1}}$ and $\mathbf{d} \in \mathbb{R}^k$ defines a set

$$\mathcal{I} = \{\mathbf{z}_{i-1} : \mathbf{C}\mathbf{z}_{i-1} \leq \mathbf{d}\}, \tag{5.1}$$

where $k$ is the number of constraints. After the linear mapping $\mathbf{z}_{i-1} \mapsto \mathbf{W}_i \mathbf{z}_{i-1} + \mathbf{b}_i$, the set before activation is denoted

$$\hat{\mathcal{I}} = \{\hat{\mathbf{z}}_i : \hat{\mathbf{C}}\hat{\mathbf{z}}_i \leq \hat{\mathbf{d}}\}, \tag{5.2}$$

where $\hat{\mathbf{C}} \in \mathbb{R}^{k \times k_i}$ and $\hat{\mathbf{d}} \in \mathbb{R}^k$. The number of inequality constraints $k$ in (5.2) may be different from that in (5.1). The set $\hat{\mathcal{I}}$ can be computed by calling `affine_map`, which converts $\mathcal{I}$ into a V-polytope, applies the linear map to all vertices, then converts it back to an H-polytope.

In the case that the input set $\mathcal{I}$ is parameterized as a star set

$$\mathcal{I} = \{\mathbf{z}_{i-1} : \mathbf{z}_{i-1} = \mathbf{c} + \mathbf{G}\boldsymbol{\alpha}, \mathbf{C}\boldsymbol{\alpha} \leq \mathbf{d}\}, \tag{5.3}$$

where $\mathbf{c} \in \mathbb{R}^{k_{i-1}}$, $\mathbf{G} \in \mathbb{R}^{k_{i-1} \times l}$, $\mathbf{C} \in \mathbb{R}^{k \times l}$, and $\mathbf{d} \in \mathbb{R}^k$. The scalar $l$ is the number of symbols and $k$ is the number of constraints. After the linear mapping $\mathbf{z}_{i-1} \mapsto \mathbf{W}_i \mathbf{z}_{i-1} + \mathbf{b}_i$, the set before activation is still a star set

$$\hat{\mathcal{I}} = \{\hat{\mathbf{z}}_i : \hat{\mathbf{z}}_i = \underbrace{\mathbf{W}_i \mathbf{c} + \mathbf{b}_i}_{\hat{\mathbf{c}}} + \underbrace{\mathbf{W}_i \mathbf{G}}_{\hat{\mathbf{G}}} \boldsymbol{\alpha}, \mathbf{C}\boldsymbol{\alpha} \leq \mathbf{d}\}. \tag{5.4}$$

The domain of the new star set remains the same, while the generator is updated by the linear mapping. The linear mapping on a star set can be computed more efficiently than that on a H-polytope.

**Nonlinear Mapping** The set $\hat{\mathcal{I}}$ can be partitioned into several non-intersecting subsets according to different activation patterns. The activation status for $\mathbf{z}_i$ is denoted $\boldsymbol{\delta}_i \in \{0,1\}^{k_i}$. Since the entries in $\boldsymbol{\delta}_i$ are binary, there is a bijection between $\boldsymbol{\delta}_i$ and an integer $h \in \{0, 1, \ldots, 2^{k_i} - 1\}$.[5] Define a diagonal matrix $\mathbf{P}_h \in \mathbb{R}^{k_i \times k_i}$, whose diagonal entries are the entires in the binary vector $\boldsymbol{\delta}_i$. Hence, there is a correspondence between $\mathbf{P}_h$ and the integer $h$. For a given activation $\boldsymbol{\delta}_i$, the before activation node $\hat{\mathbf{z}}_i$ needs to satisfy that if $\delta_{i,j} = 1$, then $\hat{z}_{i,j} \geq 0$; and if $\delta_{i,j} = 0$, then $\hat{z}_{i,j} \leq 0$. This partition corresponds to the constraint refinement discussed in section 4.2. According to the activation $\boldsymbol{\delta}_i$, the constraints on $\hat{\mathbf{z}}_i$ can be compactly written as

$$(\mathbf{I} - 2\mathbf{P}_h)\,\hat{\mathbf{z}}_i \leq 0. \tag{5.5}$$

The expression $\mathbf{I} - 2\mathbf{P}_h$ is a diagonal matrix whose entries are either 1 (for inactive nodes) or $-1$ (for active nodes). The constraint $(\mathbf{I} - 2\mathbf{P}_h)\hat{\mathbf{z}}_i \leq 0$ can be interpreted as a combination of the two constraints, $\mathbf{P}_h \hat{\mathbf{z}}_i \geq \mathbf{0}$ for active nodes and $(\mathbf{I} - \mathbf{P}_h)\hat{\mathbf{z}}_i \leq \mathbf{0}$ for inactive nodes. In the H-polytope case, the subset of $\hat{\mathcal{I}}$ that corresponds to the $h$th activation pattern is

$$\hat{\mathcal{I}}_h = \{\hat{\mathbf{z}}_i : \begin{bmatrix} \hat{\mathbf{C}} \\ \mathbf{I} - 2\mathbf{P}_h \end{bmatrix} \hat{\mathbf{z}}_i \leq \begin{bmatrix} \hat{\mathbf{d}} \\ \mathbf{0} \end{bmatrix} \}. \tag{5.6}$$

In the star set case, the subset of $\hat{\mathcal{I}}$ that corresponds to the $h$th activation pattern is

$$\hat{\mathcal{I}}_h = \{\hat{\mathbf{z}}_i : \hat{\mathbf{z}}_i = \hat{\mathbf{c}} + \hat{\mathbf{G}}\boldsymbol{\alpha}, \begin{bmatrix} \mathbf{C} \\ (\mathbf{I} - 2\mathbf{P}_h)\,\hat{\mathbf{G}} \end{bmatrix} \boldsymbol{\alpha} \leq \begin{bmatrix} \mathbf{d} \\ (2\mathbf{P}_h - \mathbf{I})\,\hat{\mathbf{c}} \end{bmatrix} \}. \tag{5.7}$$

The generator of the new star set remains the same, while the domain is updated by the constraint refinement.

For $\hat{\mathbf{z}}_i \in \hat{\mathcal{I}}_h$, the after activation nodes satisfy that $\mathbf{z}_i = \mathbf{P}_h \hat{\mathbf{z}}_i$. Hence, the reachable set $\mathcal{O}_h$ for $\hat{\mathcal{I}}_h$ is a linear transform $\hat{\mathbf{z}}_i \to \mathbf{P}_h \hat{\mathbf{z}}_i$ of $\hat{\mathcal{I}}_h$ defined by (5.6). We write the linear transformation as

$$\mathcal{O}_h = \mathbf{P}_h \circ \hat{\mathcal{I}}_h. \tag{5.8}$$

This process is implemented in `forward_partition`.

Finally, the output reachable set for $\mathcal{I}$ is the union of all $\mathcal{O}_h$,

$$\mathcal{O} = \bigcup_{h=0}^{2^{k_i}-1} \mathcal{O}_h. \tag{5.9}$$

It has been shown that the output set $\mathcal{O}$ is *tight* [72], meaning that it is not an over-approximation in the sense that for any point $\mathbf{z}_i$ in $\mathcal{O}$, there is a point $\mathbf{z}_{i-1}$ in $\mathcal{I}$ satisfying $\mathbf{z}_i = \mathbf{f}_i(\mathbf{z}_{i-1})$.

---

[5] When $h = 0$, all nodes are inactive. When $h = 2^{k_i} - 1$, all nodes are active.

For one input polytope, the output for one layer generates $2^{k_i}$ polytopes. Hence, the number of polytopes grows exponentially with the depth. Though the empty sets can be pruned out in the process, it is still inefficient to keep track of the exact reachable set for large neural networks. One way to address the problem is to parallelize the computation for individual geometric objects [62], which has been applied in the toolbox NNV [63].[6] Another way is to develop more efficient data structure to handle linear and nonlinear mappings of polytopes, *e.g.*, face lattices for feed-forward neural networks [74] and ImageStar for convolutional neural networks [60].

```
struct ExactReach end

function forward_layer(solver::ExactReach, layer::Layer, input::HPolytope)
    input = affine_map(layer, input)
    return forward_partition(layer.activation, input)
end

function forward_layer(solver::ExactReach, layer, input::Vector{<:HPolytope})
    output = Vector{HPolytope}(undef, 0)
    for i in 1:length(input)
        input[i] = affine_map(layer, input[i])
        append!(output, forward_partition(layer.activation, input[i]))
    end
    return output
end

function forward_partition(act::ReLU, input)
    N = dim(input)
    output = HPolytope{Float64}[]
    for h in 0:(2^N)-1
        P = Diagonal(1.0.*digits(h, base = 2, pad = N))
        orthant = HPolytope(Matrix(I - 2.0P), zeros(N))
        S = intersection(input, orthant)
        if !isempty(S)
            push!(output, linear_map(P, S))
        end
    end
    return output
end
```

**Algorithm 10:** ExactReach. The main process follows from algorithm 9. In the `forward_layer` function, each input set $\mathcal{I}$ first goes through an affine map to $\hat{\mathcal{I}}$. Then the function `forward_partition` partitions $\hat{\mathcal{I}}$ into $\hat{\mathcal{I}}_h$ for the $h$th activation pattern. The output set $\mathcal{O}_h$ for each $\hat{\mathcal{I}}_h$ is computed using linear transformation and the reachable set is a union of all these $\mathcal{O}_h$.

---

[6]https://github.com/verivital/nnv.

## 5.3 Ai2

In many cases, the exact reachable set is intractable. In Ai2 [25], an estimate $\tilde{\mathcal{R}}(\mathcal{X}, \mathbf{f})$ of the reachable set is obtained, which satisfies $\mathcal{R}(\mathcal{X}, \mathbf{f}) \subseteq \tilde{\mathcal{R}}(\mathcal{X}, \mathbf{f})$.

Ai2 uses an *abstract domain* to approximate the reachable set at each layer, which is represented by a set of logical formulas that capture certain geometric shapes, such as the geometries and their formulas introduced in section 2.2.[7] The choice of abstract domain needs to balance between precision and scalability. For example, the polytopes used in ExactReach are precise but not scalable, while the hyperrectangles used in MaxSens are scalable but too loose. The original implementation of Ai2 uses zonotopes, center-symmetric convex closed polytopes, which are more scalable than polytopes and tighter than hyperrectangles. Our implementation of Ai2 supports polytopes, zonotopes, and hyperrectangles.

Ai2 works for piecewise linear activation functions, *e.g.*, ReLU and max pooling. Any piecewise linear activation function can be described as one conditional affine transformation (CAT), which consists of a set of linear conditions and a set of affine mappings corresponding to the linear conditions. For example, for a ReLU activation,

$$\boldsymbol{\sigma}_i(\hat{\mathbf{z}}_i) = \begin{cases} \mathbf{P}_0 \hat{\mathbf{z}}_i & \text{if } (\mathbf{I} - 2\mathbf{P}_0)\hat{\mathbf{z}}_i \leq \mathbf{0} \\ \mathbf{P}_1 \hat{\mathbf{z}}_i & \text{if } (\mathbf{I} - 2\mathbf{P}_1)\hat{\mathbf{z}}_i \leq \mathbf{0} \\ \vdots \\ \mathbf{P}_{2^{k_i}-1} \hat{\mathbf{z}}_i & \text{if } (\mathbf{I} - 2\mathbf{P}_{2^{k_i}-1})\hat{\mathbf{z}}_i \leq \mathbf{0} \end{cases}, \tag{5.10}$$

where $\mathbf{P}_h$ for different $h$ is defined in section 5.2. Each condition corresponds to one specific activation pattern.

To propagate an abstract domain through a CAT, Ai2 introduces two basic operations: *meet* ($\sqcap$) and *join* ($\sqcup$). The meet operation splits an abstract domain into different subdomains that correspond to different conditions of the CAT. Due to the restriction of the abstract domain, those subdomains may be over-approximated and overlap with each other. For example, if the abstract domain is chosen to be hyperrectangles, then a subdomain corresponding to the condition $(\mathbf{I} - 2\mathbf{P}_h)\hat{\mathbf{z}}_i \leq \mathbf{0}$ is the smallest hyperrectangle that includes all points that satisfy the condition. After the meet operation, the reachable sets of those subdomains are computed with respect to the linear maps under the corresponding conditions. The join operation then uses one instance of the abstract domain to cover and approximate all the reachable sets. The implementation of the meet and join operations is deeply related to the chosen abstract domain. If the abstract domain is the polytope domain, ExactReach also splits the input set according to different cases and computes $\mathcal{O}_h$ in different cases. However, ExactReach does not have a join operation. Instead, it keeps track of all sets in (5.9).

---

[7] A detailed discussion of an abstract domain for certifying neural networks can be found in [55].

Our implementation is shown in algorithm 11. For simplicity, we only consider ReLU activations. The input and output sets can be any `AbstractPolytope`. An input set $\mathcal{I}$ at layer $i$ goes through the linear map defined by $\mathbf{W}_i$ and $\mathbf{b}_i$. The matrix $\mathbf{W}_i$ rotates the set $\mathcal{I}$ and $\mathbf{b}_i$ shifts the center of the set. The set after those linear maps is denoted $\hat{\mathcal{I}}$. Then the output set is

$$\hat{\mathcal{I}}_h = \hat{\mathcal{I}} \sqcap \{\hat{\mathbf{z}}_i : (\mathbf{I} - 2\mathbf{P}_h)\hat{\mathbf{z}}_i \leq \mathbf{0}\}, \tag{5.11a}$$

$$\mathcal{O}_h = \mathbf{P}_h \circ \hat{\mathcal{I}}_h, \tag{5.11b}$$

$$\mathcal{O} = \sqcup_{h=0}^{2^{k_i}-1} \mathcal{O}_h. \tag{5.11c}$$

where (5.11a) is the meet step, (5.11b) the linear mapping, and (5.11c) is the join step. Equation (5.11a) is the same as (5.8).

The number of geometric objects that describe the output reachable set equals to the number of geometric objects that describe the input set. Hence, Ai2 is much more scalable than ExactReach, though the over approximation results in incompleteness. Li *et al.* have shown that the efficiency of the algorithm can be further improved by combining abstract domain with symbolic propagation.

Following Ai2, the authors introduced DeepZ [54] which can scale to other activation functions; DeepPoly [55] that combines floating point polyhedra with intervals; RefineZono [56] that combines DeepZ analysis with optimization-based methods for more precision; and RefinePoly [53] that combines DeepPoly analysis with optimization-based methods for precision and scalability. All these methods are implemented in the toolbox ERAN.[8] Nonetheless, these methods no longer follow the split-and-join paradigm, but directly over approximate the activation function, which turns out to be more efficient.

## 5.4 MaxSens

MaxSens [69] is also a reachability method that uses over-approximation. It works for networks with monotone activation functions and low-dimensional input and output spaces. The key idea of MaxSens is to grid the input space and compute the reachable set for each grid cell. The finer the grid cells, the smaller the over-approximation. Computing the reachable sets for different cells can be done in parallel.

Algorithm 12 provides an implementation of MaxSens. The input set is a hyperrectangle. The output set can be any abstract polytope. There is one more step in the main loop than in the general `solve` function in algorithm 9, which is to partition the input set into several grid cells. The `forward_layer` function takes a hyperrectangle input set and outputs the reachable hyperrectangle.

---

[8]https://github.com/eth-sri/eran.

```julia
struct Ai2{T<:Union{Hyperrectangle, Zonotope, HPolytope}} <: Solver end
const Ai2h = Ai2{HPolytope}
const Ai2z = Ai2{Zonotope}
const Box = Ai2{Hyperrectangle}

function forward_layer(solver::Ai2h, L::Layer{ReLU}, input::AbstractPolytope)
    Ẑ = affine_map(L, input)
    relued_subsets = forward_partition(L.activation, Ẑ)
    return convex_hull(UnionSetArray(relued_subsets))
end

function forward_layer(solver::Ai2z, L::Layer{ReLU}, input::AbstractZonotope)
    Ẑ = affine_map(L, input)
    return overapproximate(Rectification(Ẑ), Zonotope)
end

function forward_layer(solver::Box, L::Layer{ReLU}, input::AbstractZonotope)
    Ẑ = approximate_affine_map(L, input)
    return rectify(Ẑ)
end
```

**Algorithm 11:** Ai2. The implementation has three versions that use three different abstract domains: H-polytopes, zonotopes, and hyperrectangles. Every geometric object needs to go through meet and join as defined in equation (5.11). In the case of H-polytopes, meet and join are equivalently replaced with `forward_partition` (defined in algorithm 10) and `convex_hull`, respectively. The meet operation computes the part of the input that satisfies a given activation pattern by adding more linear constraints to the input `HPolytope` according to equation (5.11a). For zonotopes, the meet and join are done by `Rectification` and `overapproximate` to zonotope. For hyperrectangles, the meet and join are done directly through `rectify`. The geometric operations are all supported by `LazySets`.

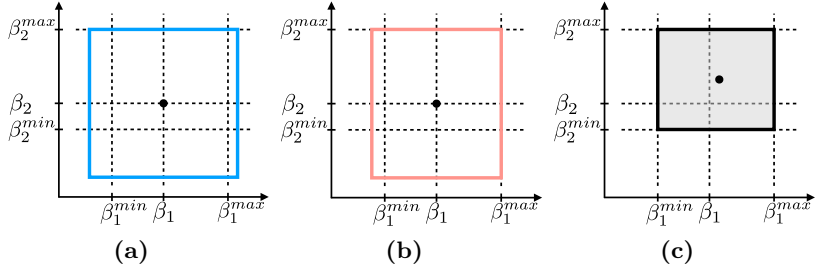

**Figure 5.5:** Illustration of different approximations of output reachable sets. (a) Center-aligned set with uniform radius. (b) Center-aligned set with non uniform radius. (c) Tight set.

Suppose the input set at layer $i$ is

$$\mathcal{I} = \{\mathbf{z}_{i-1} : |\mathbf{z}_{i-1} - \mathbf{c}_{i-1}| \leq \mathbf{r}_{i-1}\}, \tag{5.12}$$

where $\mathbf{c}_{i-1} \in \mathbb{R}^{k_{i-1}}$ is the center of the hyperrectangle and $\mathbf{r}_{i-1} \in \mathbb{R}^{k_{i-1}}$ is the radius of the hyperrectangle. The output reachable set is over-approximated by a hyperrectangle

$$\mathcal{O} = \{\mathbf{z}_i : |\mathbf{z}_i - \mathbf{c}_i| \leq \mathbf{r}_i\}, \tag{5.13}$$

where $\mathbf{c}_i, \mathbf{r}_i \in \mathbb{R}^{k_i}$.

For simplicity, define the following node-wise values for layer $i$,

$$\beta_j = \sigma_{i,j}(\mathbf{w}_{i,j}\mathbf{c}_{i-1} + b_{i,j}), \tag{5.14a}$$
$$\beta_j^{max} = \sigma_{i,j}(\mathbf{w}_{i,j}\mathbf{c}_{i-1} + |\mathbf{w}_{i,j}|\mathbf{r}_{i-1} + b_{i,j}), \tag{5.14b}$$
$$\beta_j^{min} = \sigma_{i,j}(\mathbf{w}_{i,j}\mathbf{c}_{i-1} - |\mathbf{w}_{i,j}|\mathbf{r}_{i-1} + b_{i,j}), \tag{5.14c}$$

where $\sigma_{i,j}$ is the activation function for the $j$th node, and $\mathbf{w}_{i,j}$ is the $j$th row of $\mathbf{W}_i$. Due to monotonicity of $\sigma_{i,j}$, we have

$$z_{i,j} = \sigma_{i,j}(\mathbf{w}_{i,j}\mathbf{z}_{i-1} + b_{i,j}) \in [\beta_j^{min}, \beta_j^{max}], \forall \mathbf{z}_{i-1} \in \mathcal{I}. \tag{5.15}$$

There are three different ways to define the over-approximated set $\mathcal{O}$ as illustrated in figure 5.5.

- Center-aligned set with uniform radius.

    As illustrated in figure 5.5a, the output set $\mathcal{O}$ is constructed to be a hypercube by setting all entries of $\mathbf{r}_i$ to be equal to its maximum element. Though the reachable set becomes looser, this method requires less memory to store intermediate results because we only need to store a scalar bound instead of a vector of bounds. The uniform bound indeed represents the "maximum sensitivity" of the network given

the input set. However, since we use `Hyperrectangle` objects to store intermediate results anyway, a uniform bound does not significantly enhance efficiency. Hence, our implementation uses a non-uniform bound for both $\mathcal{I}$ and $\mathcal{O}$.

- Center-aligned set with non-uniform radius.

  As illustrated in figure 5.5b, the output set $\mathcal{O}$ is no longer required to have uniform radius. By requiring that the center of the input hyperrectangle $\mathcal{I}$ maps to the center of the output hyperrectangle $\mathcal{O}$, the center and radius of the $\mathcal{O}$ can be obtained, for all $j \in \{1, \ldots, k_i\}$,

$$c_{i,j} = \beta_j, \tag{5.16a}$$

$$r_{i,j} = \max_{\mathbf{z}_{i-1} \in \mathcal{I}} |\sigma_{i,j}(\mathbf{w}_{i,j}\mathbf{z}_{i-1} + b_{i,j}) - c_{i,j}| \tag{5.16b}$$

$$= \max\{\beta_j^{max} - \beta_j, \beta_j - \beta_j^{min}\}. \tag{5.16c}$$

  where $r_{i,j}$ and $c_{i,j}$ are the $j$th entry in $\mathbf{r}_i$ and $\mathbf{c}_i$. The last equality is due to the monotonicity of $\sigma_{i,j}$.

- Tight set.

  A even tighter result can be obtained by directly using $\beta_j^{min}$ and $\beta_j^{max}$ as bounds and not aligning the centers of $\mathcal{I}$ and $\mathcal{O}$. As illustrated in figure 5.5c, the center and radius of the tight set can be defined as

$$c_{i,j} = \frac{\beta_j^{min} + \beta_j^{max}}{2}, \tag{5.17a}$$

$$r_{i,j} = \frac{\beta_j^{max} - \beta_j^{min}}{2}. \tag{5.17b}$$

  Equivalently, the output set can be represented as

$$\mathcal{O}^* = \{\mathbf{z}_i : \beta_j^{min} \leq z_{i,j} \leq \beta_j^{max}, \forall j\}. \tag{5.18}$$

  It is easy to show that the resulting node-wise bounds are the same as the bounds computed in (4.1) by interval arithmetic.

The last two cases are implemented in `forward_node`. In algorithm 12, the solver has a boolean field `tight`. When `tight` is set to be true, it computes the set in (5.17). Otherwise, it computes the center-aligned set in (5.16). The center-aligned set is desired if we want to estimate the maximum sensitivity (or maximum gradient) of the network $\mathbf{f}$ at the center of the input set. Note that in the tight case, the `forward_layer` function is identical to the `get_bounds` function in algorithm 5 introduced in chapter 4.

The advantage of MaxSens over other reachability methods is that the number of geometric objects does not grow during the layer-by-layer propagation. The total number of hyperrectangles only depends on the initial partition. Though the number of hyperrectangles
will not grow during the layer-by-layer propagation, the error of over-approximation will
accumulate quickly with respect to the number of layers. For networks with many input
nodes, the number of hyperrectangles on the initial partition can be prohibitively large
for a tight estimation. Otherwise, the computation with a sparse partition will be overly
conservative [73]. To improve the initial partition, the authors developed a specification-
guided method [73] that can adaptively choose the partition resolution according to the
problem specification.

```julia
struct MaxSens
    resolution::Float64
    tight::Bool
end

function solve(solver::MaxSens, problem::Problem)
    inputs = partition(problem.input, solver.resolution)
    f_n(x) = forward_network(solver, problem.network, x)
    outputs = map(f_n, inputs)
    return check_inclusion(outputs, problem.output)
end

function forward_layer(solver::MaxSens, L::Layer, input::Hyperrectangle)
    output = approximate_affine_map(L, input)
    β    = L.activation.(output.center)
    βmax = L.activation.(high(output))
    βmin = L.activation.(low(output))
    if solver.tight
        center = (βmax + βmin)/2
        rad =   (βmax - βmin)/2
    else
        center = β
        rad = @. max(abs(βmax - β), abs(βmin - β))
    end
    return Hyperrectangle(center, rad)
end
```

**Algorithm 12:** MaxSens. The main `solve` function is slightly different from the general reachability method
in algorithm 9 by adding a partition step. In `forward_layer`, either the tight bounds (5.17) and center-aligned
bounds (5.16) can be computed.

# 6 Primal Optimization

There can be many different designs of the optimization problem to verify (2.4). A common structure is

$$\min_{\mathbf{x},\mathbf{y}} o(\mathbf{x},\mathbf{y},\mathcal{X},\mathcal{Y}), \tag{6.1a}$$

$$\text{s.t. } \mathbf{x} \in \mathcal{X},\ \mathbf{y} \notin \mathcal{Y},\ \mathbf{y} = \mathbf{f}(\mathbf{x}), \tag{6.1b}$$

where $o(\mathbf{x},\mathbf{y},\mathcal{X},\mathcal{Y})$ is an objective function, which may depend on the input $\mathbf{x}$, the output $\mathbf{y}$ and their domains $\mathcal{X}$ and $\mathcal{Y}$. We can either minimize or maximize the objective function. The major difficulty in solving (6.1) is the nonlinear and non-convex constraint imposed by the network $\mathbf{f}$.

NSVerify [40] and MIPVerify [59] reformulate the problem (6.1) into a mixed integer linear program (MILP). Iterative LP (ILP) [8] approximates the problem (6.1) as a linear program and solves it by iteratively adding the constraints. These are all first-order linear programming methods. Higher order methods such as ones that use semidefinite programming [50, 24] will be introduced in the next chapter.

This chapter first provides an overview of common methods to simplify the primal optimization problem, then discusses the three methods (*i.e.*, NSVerify, MIPVerify, ILP) in detail. For simplicity, we only consider ReLU activations. There are different ways to encode ReLU networks as linear constraints. The following two types of variables are usually used as decision variables in the optimization.[1]

- Variables for all nodes, *i.e.*, , $\mathbf{z}_i \in \mathbb{R}^{k_i}$ for $i \in \{0,1,\ldots,n\}$.

- Variables for all activations, *i.e.*, $\boldsymbol{\delta}_i \in \{0,1\}^{k_i}$ for $i \in \{1,\ldots,n\}$.

Our implementation uses JuMP.jl[2] [17] to model and solve the optimizations. In the following discussion, `Model` refers to a jump model and `solve(::Model)` calls an LP solver through JuMP to solve the optimization problem encoded in the model.

## 6.1 Encoding a Network as Constraints

Primal optimization needs to deal with the constraint imposed by the network. The constraint $\mathbf{y} = \mathbf{f}(\mathbf{x})$ is equivalent to

$$z_{i,j} = [\mathbf{w}_{i,j}\mathbf{z}_{i-1} + b_{i,j}]_+,\ \forall i \in \{1,\ldots,n\},\ \forall j \in \{1,\ldots,k_i\}. \tag{6.2}$$

---

[1]NSVerify, MIPVerify, and ILP use these two kinds of variables. Other methods may use different sets of variables. For example, Reluplex uses $\mathbf{z}_i$ and $\hat{\mathbf{z}}_i$ as decision variables.

[2]https://jump.dev

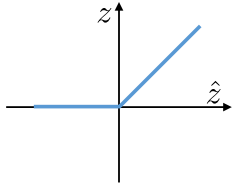

**Figure 6.1:** Illustration of ReLU activation function.

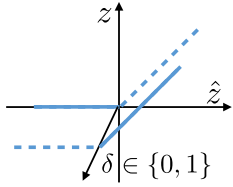

**Figure 6.2:** Illustration of linear encoding for given $\delta$. If $\delta$ is unknown, it corresponds to mixed integer encoding. The $\delta$ axis is perpendicular to the $z - \hat{z}$ plane in 3D.

We need enumerate through all nodes and methods to encode individual nodes. The code to enumerate through all nodes is shown in algorithm 13. There are different ways to encode individual nodes in (6.2). For given activations $\boldsymbol{\delta}_i$, equation (6.2) can be encoded as linear constraints as shown in algorithms 14 to 16. For given bounds $\boldsymbol{\ell}_i$ and $\mathbf{u}_i$, equation (6.2) can be encoded as linear constraints via triangle relaxation as shown in algorithm 17. Or (6.2) can be encoded as mixed integer linear constraints as shown in algorithms 18 and 19. In the following discussion, we use $\hat{z}_{i,j}$ for $\mathbf{w}_{i,j}\mathbf{z}_{i-1} + b_{i,j}$, though $\hat{z}_{i,j}$ is not a variable to be directly considered in the optimization.

**Linear constraints for given $\boldsymbol{\delta}_i$**  For a given activation $\boldsymbol{\delta}_i$ for any $i$, the network can be encoded as a set of linear constraints for $j \in \{1, \ldots, k_i\}$

$$\delta_{i,j} = 1 \Rightarrow z_{i,j} = \hat{z}_{i,j} \geq 0, \tag{6.3a}$$
$$\delta_{i,j} = 0 \Rightarrow z_{i,j} = 0, \ \hat{z}_{i,j} \leq 0. \tag{6.3b}$$

A function encoding this relationship is shown in algorithm 14. These constraints represent a subset of the original constraint (6.2), since the activation pattern is determined. This function is called in local search in Sherlock.

**Relaxed linear constraints for given $\boldsymbol{\delta}_i$**  In some cases, we drop the inequalities in (6.3) to get a relaxed encoding to ensure that the optimization problem is still feasible for infeasible activation $\boldsymbol{\delta}_i$'s. The relaxed encoding is

$$\delta_{i,j} = 1 \Rightarrow z_{i,j} = \hat{z}_{i,j}, \tag{6.4a}$$
$$\delta_{i,j} = 0 \Rightarrow z_{i,j} = 0. \tag{6.4b}$$

```julia
abstract type AbstractLinearProgram end

function encode_network!(model, network, encoding::AbstractLinearProgram)
    for (i, layer) in enumerate(network.layers)
        encode_layer!(encoding, model, layer,
                      model_params(encoding, model, i)...)
    end
end

function encode_layer!(LP::AbstractLinearProgram,
                       model,
                       layer::Layer{T},
                       ẑᵢ, zᵢ, args...)
    if T == Id
        @constraint(model, ẑᵢ .== zᵢ)
    elseif T == ReLU
        encode_relu.(LP, model, ẑᵢ, zᵢ, args...)
    end
end
```

**Algorithm 13:** The general process for encoding a linear program from a network. The function `encode_network!` applies the constraints layer by layer by calling `encode_layer!`. The encoding method is specified in the abstract type `AbstractLinearProgram`. Regardless of which encoding is used, if the layer has activation `Id` (identity), then we add the following constraint $\hat{\mathbf{z}}_i = \mathbf{z}_i$. For `ReLU` activation, each $\hat{\mathbf{z}}_{i,j}$ is related to $\mathbf{z}_{i,j}$ according to the specific encoding. The function `model_params` is responsible for extracting all of the relevant variables to encode the $i$th layer's constraints, which can vary depending on the encoding. The notation "..." allows "unpacking" a variable number of items. Recall that in Julia, a function called with a "." (as in `encode_relu.(·)`) broadcasts to each element of the collection(s) it is applied to.

```julia
struct StandardLP <: AbstractLinearProgram end

function encode_relu(::StandardLP, model, ẑᵢⱼ, zᵢⱼ, δᵢⱼ)
    if δᵢⱼ
        @constraint(model, ẑᵢⱼ >= 0.0)
        @constraint(model, zᵢⱼ == ẑᵢⱼ)
    else
        @constraint(model, ẑᵢⱼ <= 0.0)
        @constraint(model, zᵢⱼ == 0.0)
    end
end
```

**Algorithm 14:** Encoding the network as linear constraints for given $\boldsymbol{\delta}_i$. $\hat{z}_{ij}$ is a symbolic expression involving terms from $\mathbf{z}_{i-1}$, $W_i$, $b_i$.

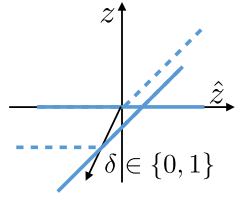

**Figure 6.3:** Illustration of relaxed linear encoding for given $\delta$.

The function is shown in algorithm 15 and illustrated in figure 6.3. This function is used in ILP.

```
struct LinearRelaxedLP <: AbstractLinearProgram end

function encode_relu(::LinearRelaxedLP, model, ẑᵢⱼ, zᵢⱼ, δᵢⱼ)
    @constraint(model, zᵢⱼ == (δᵢⱼ * ẑᵢⱼ))
end
```

**Algorithm 15:** Encoding the network as relaxed linear constraints for given $\delta_i$. Only the function `encode_relu` is shown. The functions `encode_network!` and `encode_layer!` are the same as in algorithm 13.

**Slack linear constraints for given $\delta_i$**  The previous relaxed linear encoding allows violation of the constraints imposed by ReLU's. We can use slack variables $s_{i,j}$'s to estimate the violations in the linear constraints (6.3):

$$\delta_{i,j} = 1 \Rightarrow \hat{z}_{i,j} + s_{i,j} \geq 0, \tag{6.5a}$$

$$\delta_{i,j} = 0 \Rightarrow \hat{z}_{i,j} - s_{i,j} \leq 0. \tag{6.5b}$$

The function is shown in algorithm 16. It is used in Planet to detect conflicts among different $\delta$'s.

**Triangle relaxation for given $\ell_i$ and $u_i$**  When we can bound the value of the nodes, we can use triangle relaxation to encode the constraint as

$$j \in \Gamma_i^+ \Rightarrow z_{i,j} = \hat{z}_{i,j}, \hat{z}_{i,j} \geq 0, \tag{6.6a}$$

$$j \in \Gamma_i^- \Rightarrow z_{i,j} = 0, \hat{z}_{i,j} \leq 0, \tag{6.6b}$$

$$j \in \Gamma_i \Rightarrow z_{i,j} \geq \hat{z}_{i,j}, z_{i,j} \geq 0, z_{i,j} \leq \frac{\hat{u}_{i,j}(\hat{z}_{i,j} - \hat{\ell}_{i,j})}{\hat{u}_{i,j} - \hat{\ell}_{i,j}}, \tag{6.6c}$$

where activated nodes $\Gamma_i^+$, unactivated nodes $\Gamma_i^-$, and undetermined nodes $\Gamma_i$ are introduced in (4.14). The function for triangle relaxation is shown in algorithm 17. The function `linear_transform` corresponds to (4.1). Triangle relaxation is used in ConvDual and Planet.

```
struct SlackLP <: AbstractLinearProgram end

function encode_relu(::SlackLP, model, ẑᵢⱼ, zᵢⱼ, δᵢⱼ, sᵢⱼ)
    if δᵢⱼ
        @constraint(model, zᵢⱼ == ẑᵢⱼ + sᵢⱼ)
        @constraint(model, ẑᵢⱼ + sᵢⱼ >= 0.0)
    else
        @constraint(model, zᵢⱼ == sᵢⱼ)
        @constraint(model, ẑᵢⱼ <= sᵢⱼ)
    end
end
```

**Algorithm 16:** Encoding the network as slack linear constraints for given $\delta_i$. Only the function encode_layer! is shown. The functions encode_network! and encode_layer! are the same as in algorithm 13.

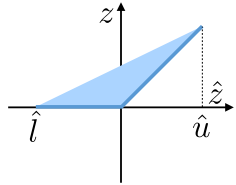

**Figure 6.4:** Illustration of triangle relaxation.

Note that node-wise triangle relaxation is not the tightest convex relaxation when there are multiple nodes in one layer, due to the fact that the dependencies of the nodes are dropped during relaxation. Singh *et al.* introduced k-ReLU to consider the dependencies and make the relaxation tighter [53]. Anderson *et al.* introduced tight mixed integer encoding and showed that their encoding is the tightest among all convex relaxations [4].

**Parallel relaxation for given $\ell_i$ and $u_i$**  Parallel relaxation is very similar to triangle relaxation, except that (6.6c) becomes

$$j \in \Gamma_i \Rightarrow \frac{\hat{u}_{i,j}}{\hat{u}_{i,j} - \hat{\ell}_{i,j}} \hat{z}_{i,j} \leq z_{i,j} \leq \frac{\hat{u}_{i,j}}{\hat{u}_{i,j} - \hat{\ell}_{i,j}} (\hat{z}_{i,j} - \hat{\ell}_{i,j}). \tag{6.7}$$

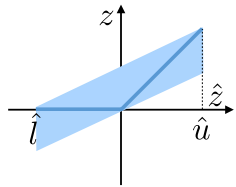

**Figure 6.5:** Illustration of parallel relaxation.

```
struct TriangularRelaxedLP <: AbstractLinearProgram end

function encode_relu(::TriangularRelaxedLP, model, ẑᵢⱼ, zᵢⱼ, l̂ᵢⱼ, ûᵢⱼ)
    if l̂ᵢⱼ > 0.0
        @constraint(model, zᵢⱼ == ẑᵢⱼ)
    elseif ûᵢⱼ < 0.0
        @constraint(model, zᵢⱼ == 0.0)
    else
        @constraints(model, begin
                        zᵢⱼ >= 0.0
                        zᵢⱼ >= ẑᵢⱼ
                        zᵢⱼ <= (ẑᵢⱼ - l̂ᵢⱼ) * ûᵢⱼ / (ûᵢⱼ - l̂ᵢⱼ)
                    end)
    end
end
```

**Algorithm 17:** Encoding the network as linear constraints for given $\ell_i$ and $\mathbf{u}_i$ via triangle relaxation. Only the function encode_relu is shown. The functions encode_network! and encode_layer! are the same as in algorithm 13.

Parallel relaxation is used in FastLin for network relaxation before applying reachability methods. To the best of our knowledge, it has not been used directly in any optimization method yet. The implementation of constraint encoding under parallel relaxation is not provided in the paper.

**Naive mixed integer linear constraints** The nonlinear constraint (6.2) can be formulated as a set of linear constraints:

$$z_{i,j} \geq \hat{z}_{i,j}, \tag{6.8a}$$
$$z_{i,j} \geq 0, \tag{6.8b}$$
$$z_{i,j} \leq \hat{z}_{i,j} + m(1 - \delta_{i,j}), \tag{6.8c}$$
$$z_{i,j} \leq m\delta_{i,j}, \tag{6.8d}$$

where $m$ should be sufficiently large. If $m$ is not large enough, the encoding may lead to error. The function is shown in algorithm 18. NSVerify calls this function.

**Mixed integer linear constraints for given $\ell_i$ and $\mathbf{u}_i$** When we have the bounds, the constraints can be more tightly encoded. When $\hat{\ell}_{i,j} \geq 0$, $z_{i,j} = \hat{z}_{i,j}$. When $\hat{u}_{i,j} \leq 0$,

```julia
struct MixedIntegerLP <: AbstractLinearProgram end

function encode_relu(::MixedIntegerLP, model, ẑᵢⱼ, zᵢⱼ, δᵢⱼ, m)
    @constraints(model, begin
                        zᵢⱼ >= 0.0
                        zᵢⱼ >= ẑᵢⱼ
                        zᵢⱼ <= m * δᵢⱼ
                        zᵢⱼ <= ẑᵢⱼ + m * (1 - δᵢⱼ)
                    end)
end
```

**Algorithm 18:** Encoding as mixed integer linear constraints using a sufficiently large number $m$. Only the function `encode_relu` is shown. The functions `encode_network!` and `encode_layer!` are the same as in algorithm 13.

$z_{i,j} = 0$. Otherwise, we have

$$z_{i,j} \geq \hat{z}_{i,j}, \tag{6.9a}$$
$$z_{i,j} \geq 0, \tag{6.9b}$$
$$z_{i,j} \leq \hat{z}_{i,j} - \hat{\ell}_{i,j}(1 - \delta_{i,j}), \tag{6.9c}$$
$$z_{i,j} \leq \hat{u}_{i,j}\delta_{i,j}. \tag{6.9d}$$

The function is shown in algorithm 19. The encoding (6.9) has the same structure as (6.8) except that $m$ in (6.8) is substituted by $-\hat{\ell}_{i,j}$ in the third inequality and $\hat{u}_{i,j}$ in the fourth inequality. The encoding (6.9) is sound for nodes that satisfy $\hat{\ell}_{i,j} \leq \hat{z}_{i,j} \leq \hat{u}_{i,j}$. As a result, it provides a way to determine $m$ in the naive encoding (6.8). Essentially, in order for the naive encoding to be sound, $m$ should be chosen such that

$$m \geq \max_{i,j}\{|\hat{u}_{i,j}|, |\hat{\ell}_{i,j}|\}. \tag{6.10}$$

If the binary variable $\delta_{i,j}$ is relaxed to be in the interval [0, 1] then the constraints in equation (6.9) become equivalent to the triangular relaxation between variables $z_{i,j}$ and $\hat{z}_{i,j}$. One approach to solve a mixed integer program is to relax each binary variable and then fix these binary variables one by one, performing a tree search over the activation space. This is similar to the Reluplex algorithm, presented in section 9.4, since both approaches would perform a tree search over the activation space of undetermined nodes. However, they differ in the subproblem that they solve at each node in the search. Each undetermined ReLU may be relaxed by some mixed integer program solvers to the triangle relaxation while Marabou will relax each node with the following constraints: $z_{i,j} \geq \hat{z}_{i,j}$, $z_{i,j} \geq 0$, and $\hat{\ell}_{i,j} \leq \hat{z}_{i,j} \leq \hat{u}_{i,j}$.

```julia
struct BoundedMixedIntegerLP <: AbstractLinearProgram end

function encode_relu(::BoundedMixedIntegerLP, model,
                    ẑᵢⱼ, zᵢⱼ, δᵢⱼ, l̂ᵢⱼ, ûᵢⱼ)
    if l̂ᵢⱼ >= 0.0
        @constraint(model, zᵢⱼ == ẑᵢⱼ)
    elseif ûᵢⱼ <= 0.0
        @constraint(model, zᵢⱼ == 0.0)
    else
        @constraints(model, begin
                        zᵢⱼ >= 0.0
                        zᵢⱼ >= ẑᵢⱼ
                        zᵢⱼ <= ûᵢⱼ * δᵢⱼ
                        zᵢⱼ <= ẑᵢⱼ - l̂ᵢⱼ * (1 - δᵢⱼ)
                     end)
    end
end
```

**Algorithm 19:** Encoding as mixed integer linear constraints using the bounds on node values. Only the function `encode_relu` is shown. The functions `encode_network!` and `encode_layer!` are the same as in algorithm 13.

## 6.2 Objective Functions

In primal optimization, there are multiple ways to design the objective. Some of them are listed in algorithm 20.

**Violation of linear constraints** In many cases, the objective function is chosen to measure the violation of the constraints. For example, when $\mathcal{Y}$ is represented by a halfspace (2.11), we maximize the following objective function

$$o := \mathbf{c}^\top \mathbf{y} - d. \tag{6.11}$$

Such design directly tells how much the output constraints can be violated in the problem (2.4). This objective is used in dual optimization methods, *e.g.*, Duality, ConvDual, and Certify.

**Maximum disturbance** The objective can also measure the maximum allowable disturbance. The disturbance with respect to a given input $\mathbf{x}_0$ is computed as

$$o := \|\mathbf{x} - \mathbf{x}_0\|_\infty. \tag{6.12}$$

The maximum allowable disturbance is computed by minimizing $o$ with respect to the constraint that $\mathbf{f}(\mathbf{x}) \notin \mathcal{Y}$. This objective is used in MIPVerify and ILP.

**Summation of variables**  The objective can also be

$$o := \sum_{i,j} s_{i,j}. \tag{6.13}$$

The summation of slack variables is used in Planet.

```julia
function max_disturbance!(model::Model, var)
    o = symbolic_infty_norm(var)
    @objective(model, Min, o)
    return o
end

function min_sum!(model::Model, var)
    o = sum(sum.(var))
    @objective(model, Min, o)
    return o
end

function max_sum!(model::Model, var)
    o = sum(sum.(var))
    @objective(model, Max, o)
    return o
end

# This is the default when creating a model. Only used for clarity.
feasibility_problem!(model::Model) = nothing
```

**Algorithm 20:** Objective functions: minimax disturbance, minimal summation, and maximal summation.

## 6.3 NSVerify

NSVerify[40] takes any linear constraints $\mathcal{X}$ and $\mathcal{Y}$, and considers networks with only ReLU activations. NSVerify encodes the ReLU activation functions as a set of mixed integer linear constraints. It does not need an objective. The method is sound and complete.

Our implementation is shown in algorithm 21. The solver needs to specify $m$ in (6.8). The solver is sound and complete only if $m$ is sufficiently big, *i.e.*, $m$ that satisfies (6.10). The solver first initializes variables $\mathbf{z}_i$ and $\boldsymbol{\delta}_i$ for all $i$. Then the solver adds the input constraint $\mathbf{z}_0 \in \mathcal{X}$ as well as the complement of the output constraint $\mathbf{z}_n \notin \mathcal{Y}$. Then it encodes the network as a set of mixed integer linear constraints. If there is a solution to the optimization, then we get a counter example. If not, the property is satisfied. In the implementation, for simplicity, we require that $\mathcal{X}$ be an `HPolytope` and $\mathcal{Y}$ a `PolytopeComplement`.

The authors of NSVerify later introduced Venus [21], which combines the MILP formulation with dependency-based pruning (*i.e.*, removing unnecessary constraints) and input refinement, and has been demonstrated to be more computationally efficient.

```julia
struct NSVerify
    optimizer
    m::Float64
end

function solve(solver::NSVerify, problem::Problem)
    model = Model(solver)
    z = init_vars(model, problem.network, :z, with_input=true)
    δ = init_vars(model, problem.network, :δ, binary=true)
    # If m is not already set, calculate it by finding an upper
    # bound on the possible variable values.
    if isnothing(solver.m)
        model[:M] = set_automatic_M(problem)
    else
        model[:M] = solver.m
    end

    add_set_constraint!(model, problem.input, first(z))
    add_complementary_set_constraint!(model, problem.output, last(z))
    encode_network!(model, problem.network, MixedIntegerLP())
    feasibility_problem!(model)
    optimize!(model)

    if termination_status(model) == OPTIMAL
        return CounterExampleResult(:violated, value(first(z)))
    else
        return CounterExampleResult(:holds)
    end
end
```

**Algorithm 21:** NSVerify. The verification problem is encoded as a mixed integer linear program. The naive encoding discussed in algorithm 18 is used in NSVerify. The parameter $m$ can specified by the solver, or a safe value can be chosen by computing the bounds of all of the node values, *e.g.*, using interval arithmetic or another technique.

## 6.4 MIPVerify

MIPVerify [59] can be viewed as a direct extension of NSVerify. It also encodes the network as mixed integer constraints, but the encoding is more efficient since MIPVerify pre-computes the bounds of the problem. In this way, the solver does not need to specify $m$ as in NSVerify.[3]

Our implementation is shown in algorithm 22. Similar to NSVerify, the solver first initializes variables $\mathbf{z}_i$ and $\boldsymbol{\delta}_i$ for all $i$. Then the solver adds the complement of the output constraint $\mathbf{z}_n \notin \mathcal{Y}$. Then it computes the bounds for the neurons, and encodes the network as a set of mixed integer linear constraints. The objective is to compute the maximum allowable disturbance. The satisfiability is determined by comparing the allowable range of disturbance with the input set. In the implementation, for simplicity, we require that $\mathcal{X}$ is a `Hyperrectangle` and $\mathcal{Y}$ is a `PolytopeComplement`. MIPVerify supports the max activation function and the $\ell_p$ norm in the objective, although these have not yet been supported in our implementation.

---

[3]The original code can be found at https://github.com/vtjeng/MIPVerify.jl.

```
struct MIPVerify
    optimizer
end

function solve(solver::MIPVerify, problem::Problem)
    model = Model(solver)
    z = init_vars(model, problem.network, :z, with_input=true)
    δ = init_vars(model, problem.network, :δ, binary=true)
    # get the pre-activation bounds:
    model[:bounds] = get_bounds(problem, false)

    add_set_constraint!(model, problem.input, first(z))
    add_complementary_set_constraint!(model, problem.output, last(z))
    encode_network!(model, problem.network, BoundedMixedIntegerLP())
    o = max_disturbance!(model, first(z) - problem.input.center)
    optimize!(model)
    if termination_status(model) == OPTIMAL
        return AdversarialResult(:violated, value(o))
    end
    return AdversarialResult(:holds)
end
```

**Algorithm 22:** MIPVerify. The verification problem is encoded as a mixed integer linear program. The bounds of node values are considered in the encoding. MIPVerify computes the maximum allowable disturbance.

## 6.5 ILP

ILP [8] encodes ReLU networks as linear constraints. It only considers a linear portion of the network that has the same activation pattern as the reference input. In our implementation, the reference input is chosen as the center of the input constraint set. The resulting problem can be solved by simply encoding a linear program using (6.3), where $\delta_{i,j}$'s denote the activation status of the reference input. To speed up the computation, ILP introduces iterative constraint solving. It first drops all inequality constraints with respect to $\hat{z}_{i,j}$ in

(6.3), *i.e.*, for all $i$ and $j$, the following constraints are dropped

$$(2\delta_{i,j} - 1)\hat{z}_{i,j} \geq 0. \tag{6.14}$$

The above expression is a compact version of $\hat{z}_{i,j} \geq 0$ for $\delta_{i,j} = 1$ and $\hat{z}_{i,j} \leq 0$ for $\delta_{i,j} = 0$. Without (6.14), the linear encoding reduces to the relaxed linear encoding in (6.4). The inequality constraint with respect to $\hat{z}_{i,j}$ is iteratively added, if the solution at the current iteration violates (6.14) for any $i$ and $j$.

Our implementation is shown in algorithm 24. For simplicity, we require that $\mathcal{X}$ be a hyperrectangle and $\mathcal{Y}$ the complement of a polytope, `PolytopeComplement`. The solver first computes the activation $\boldsymbol{\delta}_i$'s according to the reference input, *i.e.*, the center of $\mathcal{X}$. Then it initializes neuron variables $\mathbf{z}_i$'s, adds the complement of the output constraint $\mathbf{z}_n \notin \mathcal{Y}$, and adds an objective function for maximum allowable disturbance. We provide both the iterative implementation and the non-iterative implementation to solve the LP problem, where the iterative version corresponds to ILP.[4] In the non-iterative approach, the solver simply encodes the network using the linear constraints in (6.3). In the iterative approach, the solver first encodes the network using the relaxed linear constraints in (6.4) and solves the relaxed problem. If the resulting solution violates any inequality constraint (6.14), we add the constraint to the problem and solve the problem again. The process is repeated until all constraints (6.14) are satisfied. The process is guaranteed to converge in a finite number of steps since there are only finitely many constraints. The number of constraints equals the number of neurons.

# 7 Dual Optimization

Primal optimization needs to deal with complicated constraints. Another approach is to consider the dual problem of (6.1), which can be relaxed to many independent optimization problems [20, 68]. The objective considered in these methods is the violation of output constraints (6.11). The dual problem provides a valid bound on the violation. In particular, Duality [20] uses Lagrangian relaxation, which handles general activation functions such as ReLU, tanh, sigmoid, and maxpool. ConvDual [68] solves the dual problem of convexified (6.1), which handles ReLU only. Certify [50] uses semidefinite relaxation, which handles networks with one hidden layer, whose activation functions are differentiable almost everywhere.

---

[4] It is claimed that the iterative approach computes faster than the non-iterative approach [8].

```
struct ILP
    optimizer
    iterative::Bool
end

function solve(solver::ILP, problem::Problem)
    nnet = problem.network
    x = problem.input.center
    model = Model(solver)
    model[:δ] = δ = get_activation(nnet, x)
    z = init_vars(model, nnet, :z, with_input=true)
    add_complementary_set_constraint!(model, problem.output, last(z))
    o = max_disturbance!(model, first(z) - problem.input.center)

    if !solver.iterative
        encode_network!(model, nnet, StandardLP())
        optimize!(model)
        if termination_status(model) != OPTIMAL
            return AdversarialResult(:unknown)
        end
        x = value(first(z))
        return interpret_result(solver, x, problem.input)
    end

    encode_network!(model, nnet, LinearRelaxedLP())
    while true
        optimize!(model)
        if termination_status(model) != OPTIMAL
            return AdversarialResult(:unknown)
        end
        x = value(first(z))
        matched, index = match_activation(nnet, x, δ)
        if matched
            return interpret_result(solver, x, problem.input)
        end
        add_constraint!(model, nnet, z, δ, index)
    end
end
```

**Algorithm 23:** ILP. ILP computes the maximum allowable disturbance and returns adversarial results. Both iterative and non-iterative approaches are provided. The iterative approach first relaxes inequality constraints in `encode_network!` using linear relaxation, and then iteratively adds those inequality constraints in `add_constraint!`. The functions `interpret_result` and `add_constraint!` are shown in algorithm 24. The function `match_activation` finds the node that violates the inequality constraint equation (6.14). Its implementation is not shown.

```
function interpret_result(solver::ILP, x, input)
    radius = abs.(x .- center(input))
    if all(radius .>= radius_hyperrectangle(input))
        return AdversarialResult(:holds, minimum(radius))
    else
        return AdversarialResult(:violated, minimum(radius))
    end
end

function add_constraint!(model, nnet, z, δ, (i, j))
    layer = nnet.layers[i]
    ẑᵢⱼ = layer.weights[j, :]' * z[i] + layer.bias[j]
    if δ[i][j]
        @constraint(model, ẑᵢⱼ >= 0.0)
    else
        @constraint(model, ẑᵢⱼ <= 0.0)
    end
end
```

**Algorithm 24:** The functions interpret_result and add_constraint! in ILP. The solver returns :holds if the counter example **x** lies outside the hyperrectangle; otherwise :violated. The function add_constraint! iteratively adds the constraint (6.14).

## 7.1 Dual Network

This section introduces the concept of *dual network*, which is deeply related to the dual problem of the optimization with respect to a neural network.[1] The term "dual problem" refers to the Lagrangian dual problem, which is obtained by forming the Lagrangian of the optimization, using Lagrange multipliers to add the constraints to the objective function, and then solving for some primal variable values that optimize the Lagrangian. This process will be discussed in detail in section 7.2. The Lagrange multipliers are called dual variables. It will be shown that those dual variables form a dual network, whose structure is similar to the original network but which propagates in the opposite direction. Moreover, the dual variables encode the weights of corresponding nodes in a value function in the context of dynamic programming, if we regard the layer by layer propagation in a neural network as a dynamic system. In this context, the optimization-based verification problem can be understood as an optimal control problem, while the dual variables are the shadow prices [51]. In the following discussion, we first introduce the dual network in the context of dynamic programming, then point out its relationship with the Lagrangian dual problem.

---

[1] The dual network discussed here is not the dual neural network (DNN) [77], which is a recurrent neural network (RNN) to solve quadratic programming.

**Dynamic programming: General formulation** Many algorithms optimize an objective function that depends on non-input variables in the neural network, but constrained on the input $\mathbf{x}$. There is a nonlinear relationship between the objective function and the input $\mathbf{x}$. As feedforward neural networks are considered, we can use dynamic programming to simplify the nonlinear optimization problem and obtain the dual problem. Suppose the problem under consideration is

$$\max_{\mathbf{x} \in \mathcal{X}} o, \text{ where } o := \sum_{i=0}^{n} o_i(\hat{\mathbf{z}}_i), \tag{7.1}$$

where $o_i$ is the objective for different layers.[2] Define the value function at layer $i$ as

$$o_{i \to n}(\hat{\mathbf{z}}_i) := \max_{\hat{\mathbf{z}}_{i+1}, \cdots, \hat{\mathbf{z}}_n} \sum_{k \geq i} o_k(\hat{\mathbf{z}}_k). \tag{7.2}$$

The value function represents the optimal value that can be achieved given certain initial state $\hat{\mathbf{z}}_i$. Hence, the value function $o_i$ only has one variable $\hat{\mathbf{z}}_i$. The Bellman equation for dynamic programming can be written[3]

$$o_{i \to n}(\hat{\mathbf{z}}_i) = \max_{\hat{\mathbf{z}}_{i+1} = \mathbf{W}_{i+1}\boldsymbol{\sigma}_i(\hat{\mathbf{z}}_i) + \mathbf{b}_{i+1}} o_i(\hat{\mathbf{z}}_i) + o_{(i+1) \to n}(\hat{\mathbf{z}}_{i+1}). \tag{7.3}$$

The Bellman equation can be solved by backward dynamic programming. Then the original optimization (7.1) is reduced to the following one,

$$\max_{\mathbf{x} \in \mathcal{X}} o_{0 \to n}(\mathbf{x}), \tag{7.4}$$

whose objective only depends on $\mathbf{x}$.

This approach is widely used in discrete-time optimal control for dynamic systems where $\hat{\mathbf{z}}_i$ are states at step $i$. In the following discussion, we derive the explicit solution for linear objectives.

**Dynamic programming: Linear objective** Assume that $o_i(\hat{\mathbf{z}}_i) = \mathbf{c}_i^\mathsf{T} \hat{\mathbf{z}}_i - d_i$ where $\mathbf{c}_i \in \mathbb{R}^{k_i}$ and $d_i \in \mathbb{R}$. Define $\mathbf{v}_i \in \mathbb{R}^{k_i}$ to be the dual variable for $\hat{\mathbf{z}}_i$, and $\hat{\mathbf{v}}_i \in \mathbb{R}^{k_i}$ to be the dual variable for $\mathbf{z}_i$. The dual variables encode the weights of the corresponding nodes in a value function, *i.e.*,

$$o_{i \to n}(\hat{\mathbf{z}}_i) = \mathbf{v}_i^\mathsf{T} \hat{\mathbf{z}}_i + \gamma_i, \tag{7.5}$$

---

[2]The objective in (7.1) is nonlinear and depends on hidden variables. In most cases, the objective is linear that only depends on the output layer, *e.g.*, $o = \mathbf{c}^\mathsf{T} \hat{\mathbf{z}}_n - d$.

[3]The Bellman equation is not in its standard form in that the right-hand-side does not involve any actual optimization. In conventional dynamic programming, the relationship between $\hat{\mathbf{z}}_i$ and $\hat{\mathbf{z}}_{i+1}$ is non-deterministic. There should be a "decision variable" that affects the relationship, which can be optimized over. However, there is not such a variable in a neural network.

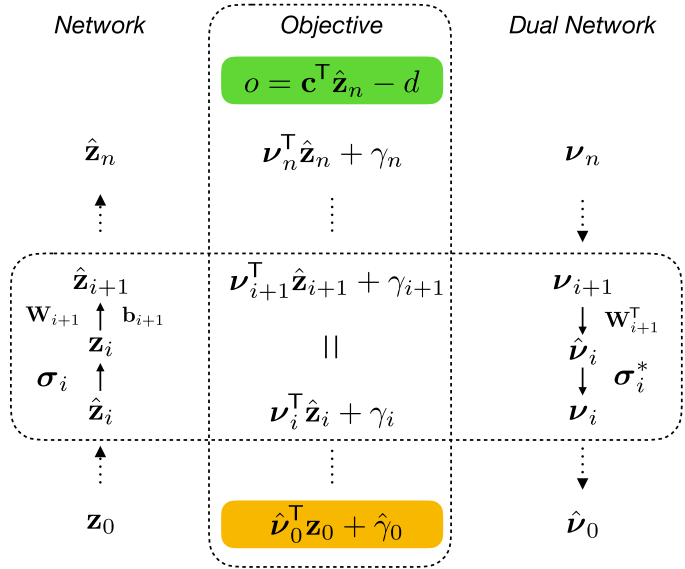

**Figure 7.1:** Illustration of dynamic programming and the dual network. Given an objective function $o$ that depends on the network output, dynamic programming can transform $o$ to a function that only depends on the network input. In the process, a dual network is constructed. The dual network propagates in the reverse order to the original network. All corresponding mappings are dual functions of the mappings in the original network. Moreover, the nodes $\nu_i$ in the dual network encode the weights of $z_i$ to the objective $o$. According to the stationary condition in dynamic programming, all expressions in the "objective" block are equivalent to one another. Then the objective function $o$ that depends on the output (shaded in green) is transformed to a function that only depends on the input (shaded in yellow).

where $\gamma_i \in \mathbb{R}$ is a bias term. For $o_n$, we have the boundary constraint

$$\mathbf{v}_n = \mathbf{c}_n, \gamma_n = -d_i. \tag{7.6}$$

Plugging (7.5) into the Bellman equation (7.3), we have

$$\mathbf{v}_i^\mathsf{T} \hat{\mathbf{z}}_i + \gamma_i = \mathbf{c}_i^\mathsf{T} \hat{\mathbf{z}}_i - d_i + \mathbf{v}_{i+1}^\mathsf{T} \mathbf{W}_{i+1} \boldsymbol{\sigma}_i(\hat{\mathbf{z}}_i) + \mathbf{v}_{i+1}^\mathsf{T} \mathbf{b}_{i+1} + \gamma_{i+1}. \tag{7.7}$$

Matching the coefficient of $\hat{\mathbf{z}}_i$ and the constant term on both sides, we obtain the backward relationship between $\mathbf{v}_i, \gamma_i$ and $\mathbf{v}_{i+1}, \gamma_{i+1}$ as

$$\mathbf{v}_i = \mathbf{c}_i + \boldsymbol{\sigma}_i^*(\hat{\mathbf{v}}_i), \tag{7.8a}$$
$$\hat{\mathbf{v}}_i = \mathbf{W}_{i+1}^\mathsf{T} \mathbf{v}_{i+1}, \tag{7.8b}$$
$$\gamma_i = \gamma_{i+1} + \mathbf{v}_{i+1}^\mathsf{T} \mathbf{b}_{i+1} - d_i, \tag{7.8c}$$

where $\boldsymbol{\sigma}_i^*$ is the dual function of $\boldsymbol{\sigma}_i$, defined to satisfy[4]

$$\boldsymbol{\sigma}_i^*(\hat{\mathbf{v}}_i)^\mathsf{T} \hat{\mathbf{z}}_i \equiv \hat{\mathbf{v}}_i^\mathsf{T} \boldsymbol{\sigma}_i(\hat{\mathbf{z}}_i). \tag{7.9}$$

In this way, the dual variables $\mathbf{v}_i$ and $\hat{\mathbf{v}}_i$ indeed form a backward dual network. Figure 7.1 illustrates the original network, the value function, and the dual network in the case that $o_i = 0$ for $i < n$. Hence, the original optimization problem (7.1) can be solved by 1) backward computing the dual network from the boundary constraint at layer $n$ to layer 0, 2) solving the reduced problem (7.4) that depends on the dual variables.

When $\boldsymbol{\sigma}_i$ is nonlinear, the dual function $\boldsymbol{\sigma}_i^*$ is difficult to handle, as it indeed depends on the value of $\hat{\mathbf{z}}_i$. In the case of ReLU activation, different approximations of $\boldsymbol{\sigma}_i$ are introduced to simplify the dual network. For example, ConvDual uses triangle relaxation and FastLin uses parallel relaxation. These approaches are to be introduced in section 7.3 and section 8.3.

**Dual network and Lagrange multipliers** The dual variables for dynamic programming are deeply related to Lagrange multipliers in Lagrangian dual problem. Let $\boldsymbol{\mu}_i$ be the multiplier for the constraint $\hat{\mathbf{z}}_i = \mathbf{W}_i \mathbf{z}_{i-1} + \mathbf{b}_i$ and $\boldsymbol{\lambda}_i$ the multiplier for the constraint $\mathbf{z}_i = \boldsymbol{\sigma}_i(\hat{\mathbf{z}}_i)$. Then the Bellman equation (7.3) can be rewritten as[5]

$$o_{i \to n}(\hat{\mathbf{z}}_i) = \min_{\boldsymbol{\mu}_{i+1}, \boldsymbol{\lambda}_i} \max_{\hat{\mathbf{z}}_{i+1}, \mathbf{z}_i} o_i(\hat{\mathbf{z}}_i) + o_{(i+1) \to n}(\hat{\mathbf{z}}_{i+1}) \tag{7.10}$$
$$+ \boldsymbol{\mu}_{i+1}^\mathsf{T} (\hat{\mathbf{z}}_{i+1} - \mathbf{W}_{i+1} \mathbf{z}_i - \mathbf{b}_{i+1}) + \boldsymbol{\lambda}_i^\mathsf{T} (\mathbf{z}_i - \boldsymbol{\sigma}_i(\hat{\mathbf{z}}_i)).$$

---

[4]To obtain an upper bound of the primal problem (7.1), the dual function only needs to satisfy $\boldsymbol{\sigma}_i^*(\hat{\mathbf{v}}_i)^\mathsf{T} \hat{\mathbf{z}}_i \geq \hat{\mathbf{v}}_i^\mathsf{T} \boldsymbol{\sigma}_i(\hat{\mathbf{z}}_i)$.

[5]The solution of the unconstrained problem $\min_\mathbf{a} \max_\mathbf{b} \mathbf{a}^\mathsf{T} \mathbf{b}$ is always $\mathbf{0}$ with $\mathbf{a} = \mathbf{b} = \mathbf{0}$. Hence, the optimal solution of the Bellman equation (7.10) is always $\boldsymbol{\mu}_{i+1} = \hat{\mathbf{z}}_{i+1} - \mathbf{W}_{i+1} \mathbf{z}_i - \mathbf{b}_{i+1} = \mathbf{0}$ and $\boldsymbol{\lambda}_i = \mathbf{z}_i - \boldsymbol{\sigma}_i(\hat{\mathbf{z}}_i) = \mathbf{0}$.

Using the linearity assumption $o_{(i+1) \to n}(\hat{\mathbf{z}}_{i+1}) = \mathbf{v}_{i+1}^\mathsf{T} \hat{\mathbf{z}}_{i+1} + \gamma_{i+1}$ in the above equation, we have

$$o_{i \to n}(\hat{\mathbf{z}}_i) = o_i^* + \min_{\boldsymbol{\mu}_{i+1}, \boldsymbol{\lambda}_i} \max_{\hat{\mathbf{z}}_{i+1}, \mathbf{z}_i} \mathbf{v}_{i+1}^\mathsf{T} \hat{\mathbf{z}}_{i+1} \qquad (7.11)$$
$$+ \boldsymbol{\mu}_{i+1}^\mathsf{T} (\hat{\mathbf{z}}_{i+1} - \mathbf{W}_{i+1} \mathbf{z}_i - \mathbf{b}_{i+1}) + \boldsymbol{\lambda}_i^\mathsf{T} (\mathbf{z}_i - \boldsymbol{\sigma}_i(\hat{\mathbf{z}}_i)),$$

where $o_i^* = \mathbf{c}_i^\mathsf{T} \hat{\mathbf{z}}_i - d_i + \gamma_{i+1}$. Rearrange the minimax problem,

$$\min_{\boldsymbol{\mu}_{i+1}, \boldsymbol{\lambda}_i} \max_{\hat{\mathbf{z}}_{i+1}, \mathbf{z}_i} (\mathbf{v}_{i+1} + \boldsymbol{\mu}_{i+1})^\mathsf{T} \hat{\mathbf{z}}_{i+1} + (\boldsymbol{\lambda}_i - \mathbf{W}_{i+1}^\mathsf{T} \boldsymbol{\mu}_{i+1})^\mathsf{T} \mathbf{z}_i - \boldsymbol{\mu}_{i+1}^\mathsf{T} \mathbf{b}_{i+1} - \boldsymbol{\lambda}_i^\mathsf{T} \boldsymbol{\sigma}_i(\hat{\mathbf{z}}_i). \qquad (7.12)$$

By applying the result in footnote 5 to the first two terms in (7.12), we conclude that $\mathbf{v}_{i+1} + \boldsymbol{\mu}_{i+1} = \mathbf{0}$ and $\boldsymbol{\lambda}_i - \mathbf{W}_{i+1}^\mathsf{T} \boldsymbol{\mu}_{i+1} = \mathbf{0}$. Hence, we have the following relationship

$$\boldsymbol{\mu}_{i+1} = -\mathbf{v}_{i+1}, \boldsymbol{\lambda}_i = -\hat{\mathbf{v}}_i. \qquad (7.13)$$

The nodes in the dual network are indeed the Lagrange multipliers if we do not consider the bounds on $\mathbf{z}_i$'s. Duality in section 7.2 provides a formulation that considers the bounds. In this case, the conclusion from footnote 5 does not hold. As a result, the relationship between the dual network and the Lagrange multipliers in (7.13) breaks.

**Dual network and backpropagation** When the objectives for hidden layers are all zero, i.e., $o_i = 0$ for all $i < n$, the dual variable $\mathbf{v}_i$ is indeed the gradient from the objective $o$ to the hidden variable $\hat{\mathbf{z}}_i$. The gradients are usually computed in backpropagation to train deep neural networks [26]. LeCun *et al.* pointed out the relationship between backpropagation and optimal control in their early paper [36].

## 7.2 Duality

Duality [20] takes a hyperrectangle as its input set and has a halfspace as its output set. The input hyperrectangle is denoted $|\mathbf{x} - \mathbf{x}_0| \leq \mathbf{r}$, where $\mathbf{x}_0$ and $\mathbf{r}$ are the center and radius of the hyperrectangle. The output halfspace is denoted $\mathbf{c}^\mathsf{T} \mathbf{y} \leq d$. Then the optimization problem becomes

$$\max_{\mathbf{z}_0, \ldots, \mathbf{z}_n, \hat{\mathbf{z}}_1, \ldots, \hat{\mathbf{z}}_n} \mathbf{c}^\mathsf{T} \mathbf{z}_n - d, \qquad (7.14\text{a})$$
$$\text{s.t.} \quad \mathbf{z}_i = \boldsymbol{\sigma}_i(\hat{\mathbf{z}}_i), \forall i \in \{1, \ldots, n\}, \qquad (7.14\text{b})$$
$$\hat{\mathbf{z}}_i = \mathbf{W}_i \mathbf{z}_{i-1} + \mathbf{b}_i, \forall i \in \{1, \ldots, n\}, \qquad (7.14\text{c})$$
$$|\mathbf{z}_0 - \mathbf{x}_0| \leq \mathbf{r}. \qquad (7.14\text{d})$$

Given the bounds on $\mathbf{z}_i$ and $\hat{\mathbf{z}}_i$, the optimal value of (7.14) is bounded by Lagrangian relaxation of the constraints

$$\max_{\mathbf{z}_0,\ldots,\mathbf{z}_n,\hat{\mathbf{z}}_1,\ldots,\hat{\mathbf{z}}_n} \mathbf{c}^\mathsf{T}\mathbf{z}_n - d + \sum_{i=1}^n \boldsymbol{\mu}_i^\mathsf{T}\left(\hat{\mathbf{z}}_i - \mathbf{W}_i\mathbf{z}_{i-1} - \mathbf{b}_i\right) + \sum_{i=1}^n \boldsymbol{\lambda}_i^\mathsf{T}\left(\mathbf{z}_i - \boldsymbol{\sigma}_i(\hat{\mathbf{z}}_i)\right), \quad (7.15\text{a})$$

$$\text{s.t. } \boldsymbol{\ell}_i \leq \mathbf{z}_i \leq \mathbf{u}_i, \forall i \in \{1,\ldots,n\}, \quad (7.15\text{b})$$

$$\hat{\boldsymbol{\ell}}_i \leq \hat{\mathbf{z}}_i \leq \hat{\mathbf{u}}_i, \forall i \in \{1,\ldots,n\}, \quad (7.15\text{c})$$

$$|\mathbf{z}_0 - \mathbf{x}_0| \leq \mathbf{r}, \quad (7.15\text{d})$$

where $\boldsymbol{\lambda}_i \in \mathbb{R}^{k_i}$ and $\boldsymbol{\mu} \in \mathbb{R}^{k_i}$ are Lagrange multipliers. For any choice of $\boldsymbol{\lambda}$ and $\boldsymbol{\mu}$, (7.15) provides a valid upper bound on the optimal value of (7.14). This property is known as weak duality.

Since the objective and constraints are separable in the layers, the variables in each layer can be optimized independently. The boundary condition is $\boldsymbol{\lambda}_n = -\mathbf{c}$. The objective function in (7.15) can be decomposed into the following three parts:

- Input layer value with respect to $\mathbf{z}_0$.

$$f_0(\boldsymbol{\mu}_1) = \max_{|\mathbf{z}_0 - \mathbf{x}_0| \leq \mathbf{r}} -\boldsymbol{\mu}_1^\mathsf{T}\left(\mathbf{W}_1\mathbf{z}_0 + \mathbf{b}_1\right), \quad (7.16\text{a})$$

$$= -\boldsymbol{\mu}_1^\mathsf{T}\mathbf{W}_1\mathbf{x}_0 - \boldsymbol{\mu}_1^\mathsf{T}\mathbf{b}_1 + \left|\boldsymbol{\mu}_1^\mathsf{T}\mathbf{W}_1\right|\mathbf{r}. \quad (7.16\text{b})$$

- Layer value with respect to $\mathbf{z}_i$. For $i \in \{1,\ldots,n-1\}$,

$$f_i(\boldsymbol{\lambda}_i, \boldsymbol{\mu}_{i+1}) = \max_{\boldsymbol{\ell}_i \leq \mathbf{z}_i \leq \mathbf{u}_i} -\boldsymbol{\mu}_{i+1}^\mathsf{T}\left(\mathbf{W}_{i+1}\mathbf{z}_i + \mathbf{b}_{i+1}\right) + \boldsymbol{\lambda}_i^\mathsf{T}\mathbf{z}_i, \quad (7.17\text{a})$$

$$= \left(\boldsymbol{\lambda}_i - \mathbf{W}_{i+1}^\mathsf{T}\boldsymbol{\mu}_{i+1}\right)^\mathsf{T} \frac{\boldsymbol{\ell}_i + \mathbf{u}_i}{2} - \boldsymbol{\mu}_{i+1}^\mathsf{T}\mathbf{b}_{i+1}$$

$$+ \left|\boldsymbol{\lambda}_i - \mathbf{W}_{i+1}^\mathsf{T}\boldsymbol{\mu}_{i+1}\right|^\mathsf{T} \frac{\mathbf{u}_i - \boldsymbol{\ell}_i}{2}. \quad (7.17\text{b})$$

- Activation value with respect to $\hat{\mathbf{z}}_i$. For $i \in \{1,\ldots,n\}$,

$$\tilde{f}_i(\boldsymbol{\lambda}_i, \boldsymbol{\mu}_i) = \max_{\hat{\boldsymbol{\ell}}_i \leq \hat{\mathbf{z}}_i \leq \hat{\mathbf{u}}_i} \boldsymbol{\mu}_i^\mathsf{T}\hat{\mathbf{z}}_i - \boldsymbol{\lambda}_i^\mathsf{T}\boldsymbol{\sigma}_i(\hat{\mathbf{z}}_i), \quad (7.18\text{a})$$

$$\leq \sum_j \max\{\mu_{i,j}\hat{\ell}_{i,j}, \mu_{i,j}\hat{u}_{i,j}\}$$

$$+ \sum_j \max\{-\lambda_{i,j}\sigma_{i,j}(\hat{\ell}_{i,j}), -\lambda_{i,j}\sigma_{i,j}(\hat{u}_{i,j})\}, \quad (7.18\text{b})$$

where the inequality is taken by considering element-wise maximum. For ReLU activations, the optimization objective in (7.18a) is piecewise linear in $\hat{\mathbf{z}}_i$. On each linear piece, the optimum is attached at one of the endpoints of the input domain.

Define $g_{i,j}(\hat{z}) := \mu_{i,j}\hat{z} - \lambda_{i,j}[\hat{z}]_+$. Then the activation value with respect to $\hat{\mathbf{z}}_i$ under ReLU activations can be computed as

$$\tilde{f}_i(\boldsymbol{\lambda}_i, \boldsymbol{\mu}_i) = \sum_j \max_{\hat{\ell}_{i,j} \leq \hat{z} \leq \hat{u}_{i,j}} g_{i,j}(\hat{z}), \tag{7.19a}$$

$$\max_{\hat{\ell}_{i,j} \leq \hat{z} \leq \hat{u}_{i,j}} g_{i,j}(\hat{z}) = \begin{cases} \max\{g_{i,j}(\hat{\ell}_{i,j}), g_{i,j}(\hat{u}_{i,j}), 0\} & \text{if } \hat{\ell}_{i,j} < 0 < \hat{u}_{i,j} \\ \max\{g_{i,j}(\hat{\ell}_{i,j}), g_{i,j}(\hat{u}_{i,j})\} & \text{otherwise} \end{cases}. \tag{7.19b}$$

Note (7.19) is tighter than the bound obtained in (7.18).

Any choice of the dual variables $\boldsymbol{\lambda}_i, \boldsymbol{\mu}_i$ in (7.15) provides an upper bound of the primal problem (7.14). To obtain a tight bound, we need to minimize (7.15) with respect to the dual variables. Hence, the dual problem can be constructed as

$$\min_{\boldsymbol{\lambda}_1,\ldots,\boldsymbol{\lambda}_n,\boldsymbol{\mu}_1,\ldots,\boldsymbol{\mu}_n} f_0(\boldsymbol{\mu}_1) + \sum_{i=1}^{n-1} f_i(\boldsymbol{\lambda}_i, \boldsymbol{\mu}_{i+1}) + \sum_{i=1}^{n} \tilde{f}_i(\boldsymbol{\lambda}_i, \boldsymbol{\mu}_i) - d, \tag{7.20a}$$

$$\text{s.t.} \quad \boldsymbol{\lambda}_n = -\mathbf{c}. \tag{7.20b}$$

The problem is satisfied if the optimal solution of (7.20) is negative; otherwise, it is not satisfied. Algorithm 25 provides an implementation.

## 7.3 ConvDual

ConvDual [68] takes a hypercube input set and a halfspace output set. The input hypercube is denoted $\|\mathbf{x} - \mathbf{x}_0\|_\infty \leq \epsilon$, where $\mathbf{x}_0$ and $\epsilon$ are the center and radius of the hypercube. The output halfspace is denoted $\mathbf{c}^\mathsf{T}\mathbf{y} \leq d$. ConvDual only considers ReLU networks. It first relaxes the constraint using triangle relaxation. The primal optimization becomes a convex problem:[6]

$$\max_{\mathbf{z}_0,\ldots,\mathbf{z}_n,\hat{\mathbf{z}}_1,\ldots,\hat{\mathbf{z}}_n} \mathbf{c}^\mathsf{T}\hat{\mathbf{z}}_n - d, \tag{7.21a}$$

$$\text{s.t.} \quad z_{i,j} = \hat{z}_{i,j}, \forall i \in \{1,\ldots,n-1\}, j \in \Gamma_i^+, \tag{7.21b}$$

$$z_{i,j} = 0, \forall i \in \{1,\ldots,n-1\}, j \in \Gamma_i^-, \tag{7.21c}$$

$$z_{i,j} \geq \hat{z}_{i,j}, z_{i,j} \geq 0, z_{i,j} \leq \frac{\hat{u}_{i,j}(\hat{z}_{i,j} - \hat{\ell}_{i,j})}{\hat{u}_{i,j} - \hat{\ell}_{i,j}},$$

$$\forall i \in \{1,\ldots,n-1\}, j \in \Gamma_i, \tag{7.21d}$$

$$\hat{\mathbf{z}}_i = \mathbf{W}_i \mathbf{z}_{i-1} + \mathbf{b}_i, \forall i \in \{1,\ldots,n\}, \tag{7.21e}$$

$$\|\mathbf{z}_0 - \mathbf{x}_0\|_\infty \leq \epsilon. \tag{7.21f}$$

---

[6]Convdual considers the pre-activation bounds of the last layer, which is equivalent to the case that the activation in the last layer is identity.

```julia
struct Duality
    optimizer
end

function solve(solver::Duality, problem::Problem)
    model = Model(solver)
    c, d = tosimplerep(problem.output)
    λ = init_vars(model, problem.network, :λ)
    μ = init_vars(model, problem.network, :μ)
    o = dual_value(solver, problem, model, λ, μ)
    @constraint(model, last(λ) .== -c)
    optimize!(model)
    if termination_status(model) != OPTIMAL
        return BasicResult(:unknown)
    elseif value(o) - d[1] <= 0.0
        return BasicResult(:holds)
    else
        return BasicResult(:violated)
    end
end

function dual_value(solver::Duality, problem, model, λ, μ)
    bounds = get_bounds(problem)
    layers = problem.network.layers
    λ = [zeros(dim(bounds[1])), λ...]
    o = 0
    for i in 1:length(layers)
        Lᵢ, Bᵢ = layers[i], bounds[i]
        W, b = Lᵢ.weights, Lᵢ.bias
        c, r = Bᵢ.center, Bᵢ.radius
        B̂ᵢ₊₁ = approximate_affine_map(Lᵢ, Bᵢ)
        o += λ[i]'*c - μ[i]'*(W*c + b)
        o += sum(symbolic_abs.(λ[i] .- W'*μ[i]) .* r)
        o += activation_value(Lᵢ.activation, μ[i], λ[i+1],
                              low(B̂ᵢ₊₁), high(B̂ᵢ₊₁))
    end
    @objective(model, Min, o)
    return o
end
```

**Algorithm 25:** Duality. Lagrangian relaxation is considered in Duality. The variables to be optimized are the Lagrange multipliers. The dual value is computed layer-wise. The function `dual_value` computes the dual value layer-by-layer using (7.16), (7.17), and `activation_value` implemented in algorithm 26. Duality returns basic satisfiability results.

```
function activation_value(σ::ReLU, μ, λ, l̂, û)
    gl̂ = @. μ*l̂ - λ*σ(l̂)
    gû = @. μ*û - λ*σ(û)
    max = symbolic_max
    return sum(@. ifelse(l̂ < 0 < û, max(gl̂, gû, 0), max(gl̂, gû)))
end

function activation_value(σ::Id, μ, λ, l̂, û)
    max = symbolic_max
    sum(@. max(μ*l̂ - λ*l̂, μ*û - λ*û))
end

function activation_value(σ::Any, μ, λ, l̂, û)
    max = symbolic_max
    sum(@. max(μ*l̂, μ*û) + max(-λ*σ(l̂), -λ*σ(û)))
end
```

**Algorithm 26:** The activation value in Duality. For general activation, (7.18) is implemented. For ReLU activation, (7.19) is implemented. For identity activation, the optimization in (7.18a) becomes linear and equals to $\max\{\boldsymbol{\mu}_i^\mathsf{T}\hat{\boldsymbol{\ell}}_i - \boldsymbol{\lambda}_i^\mathsf{T}\hat{\boldsymbol{\ell}}_i, \boldsymbol{\mu}_i^\mathsf{T}\hat{\mathbf{u}}_i - \boldsymbol{\lambda}_i^\mathsf{T}\hat{\mathbf{u}}_i\}$.

The optimization (7.21) can be regarded as an $n$-step dynamic program. We derive the dual problem and the dual network following the procedure of dynamic programming discussed in section 7.1. The original paper [68] directly applies the dual problem formulation.

Recall that $\boldsymbol{\nu}_i$ and $\hat{\boldsymbol{\nu}}_i$ are the dual variables for $\hat{\mathbf{z}}_i$ and $\mathbf{z}_i$. The boundary condition is $\boldsymbol{\nu}_n = \mathbf{c}$ and $\gamma_n = -d$. The Bellman equation (7.7) is reduced to

$$\boldsymbol{\nu}_i^\mathsf{T}\hat{\mathbf{z}}_i + \gamma_i = \hat{\boldsymbol{\nu}}_i^\mathsf{T}\mathbf{z}_i + \boldsymbol{\nu}_{i+1}^\mathsf{T}\mathbf{b}_{i+1} + \gamma_{i+1}. \tag{7.22}$$

To obtain the backward relationship between $\boldsymbol{\nu}_i, \gamma_i$ and $\boldsymbol{\nu}_{i+1}, \gamma_{i+1}$ similar to (7.8), we consider the following three cases. The sets $\Gamma_i^+$, $\Gamma_i^-$, and $\Gamma_i$ encode activation conditions as defined in (4.14).

- For active node $j \in \Gamma_i^+$.

  $z_{i,j} = \hat{z}_{i,j}$. Hence, $\nu_{i,j} = \hat{\nu}_{i,j}$.

- For inactive node $j \in \Gamma_i^-$.

  $z_{i,j} = 0$. Hence, $\nu_{i,j} = 0$.

- For undetermined node $j \in \Gamma_i$. Consider the triangle relaxation in (7.21d), the following conditions are satisfied,

$$\hat{\nu}_{i,j}z_{i,j} \leq \frac{\hat{\nu}_{i,j}\hat{u}_{i,j}}{\hat{u}_{i,j} - \hat{\ell}_{i,j}}\hat{z}_{i,j} - \frac{\hat{\nu}_{i,j}\hat{u}_{i,j}\hat{\ell}_{i,j}}{\hat{u}_{i,j} - \hat{\ell}_{i,j}} \quad \text{for } \hat{\nu}_{i,j} \geq 0, \tag{7.23a}$$

$$\hat{\nu}_{i,j}z_{i,j} \leq \alpha_{i,j}\hat{\nu}_{i,j}\hat{z}_{i,j} \quad \text{for } \hat{\nu}_{i,j} < 0, \tag{7.23b}$$

where $\alpha_{i,j} \in [0,1]$. (7.23) implies the following condition:

$$\hat{\nu}_{i,j} z_{i,j} \leq \frac{\hat{u}_{i,j}}{\hat{u}_{i,j} - \hat{\ell}_{i,j}} [\hat{\nu}_{i,j}]_+ \hat{z}_{i,j} + \alpha_{i,j} [\hat{\nu}_{i,j}]_- \hat{z}_{i,j} - \frac{\hat{u}_{i,j} \hat{\ell}_{i,j}}{\hat{u}_{i,j} - \hat{\ell}_{i,j}} [\hat{\nu}_{i,j}]_+. \tag{7.24}$$

Hence, the relationship among the dual variables that maximizes the value is

$$\gamma_i = \mathbf{v}_{i+1}^\mathsf{T} \mathbf{b}_{i+1} + \gamma_{i+1} - \sum_{j \in \Gamma_i} \frac{\hat{u}_{i,j} \hat{\ell}_{i,j}}{\hat{u}_{i,j} - \hat{\ell}_{i,j}} [\hat{\nu}_{i,j}]_+, \tag{7.25a}$$

$$\nu_{i,j} = \begin{cases} \hat{\nu}_{i,j} & j \in \Gamma_i^+ \\ 0 & j \in \Gamma_i^- \\ \frac{\hat{u}_{i,j}}{\hat{u}_{i,j} - \hat{\ell}_{i,j}} [\hat{\nu}_{i,j}]_+ + \alpha_{i,j} [\hat{\nu}_{i,j}]_- & j \in \Gamma_i \end{cases}, \tag{7.25b}$$

for $i \in \{1, \ldots, n-1\}$. Note that when $j \in \Gamma_i$ and $\nu_{i,j} > 0$, $\nu_{i,j} = \frac{\hat{u}_{i,j}}{\hat{u}_{i,j} - \hat{\ell}_{i,j}} [\hat{\nu}_{i,j}]_+$. Then the term $\frac{\hat{u}_{i,j} \hat{\ell}_{i,j}}{\hat{u}_{i,j} - \hat{\ell}_{i,j}} [\hat{\nu}_{i,j}]_+$ can be simplified as $\hat{\ell}_{i,j} [\nu_{i,j}]_+$.

Then optimizing $\mathbf{v}_n^T \hat{\mathbf{z}}_n + \gamma_n$ is equivalent to optimizing $\mathbf{v}_1^T \hat{\mathbf{z}}_1 + \gamma_1 = \hat{\mathbf{v}}_0^T \mathbf{z}_0 + \mathbf{v}_1^\mathsf{T} \mathbf{b}_1 + \gamma_1$. Hence,

$$\max_{\|\mathbf{z}_0 - \mathbf{x}_0\|_\infty \leq \epsilon} \hat{\mathbf{v}}_0^T \mathbf{z}_0 + \mathbf{v}_1^\mathsf{T} \mathbf{b}_1 + \gamma_1, \tag{7.26a}$$

$$= \hat{\mathbf{v}}_0^T \mathbf{x}_0 + \epsilon \|\hat{\mathbf{v}}_0\|_1 + \underbrace{\sum_{i=1}^{n} \mathbf{v}_i^\mathsf{T} \mathbf{b}_i - \sum_{i=1}^{n-1} \sum_{j \in \Gamma_i} \hat{\ell}_{i,j} [\nu_{i,j}]_+ - d}_{\mathbf{v}_1^\mathsf{T} \mathbf{b}_1 + \gamma_1}, \tag{7.26b}$$

where $\gamma_1$ is computed according to (7.25a).

Then the dual problem of (7.21) is

$$\min_{\alpha_{i,j} \in [0,1]} \hat{\mathbf{v}}_0^T \mathbf{x}_0 + \epsilon \|\hat{\mathbf{v}}_0\|_1 + \sum_{i=1}^{n} \mathbf{v}_i^\mathsf{T} \mathbf{b}_i - \sum_{i=1}^{n-1} \sum_{j \in \Gamma_i} \hat{\ell}_{i,j} [\nu_{i,j}]_+ - d, \tag{7.27a}$$

s.t. $\quad \mathbf{v}_n = \mathbf{c}, \tag{7.27b}$

$\quad \hat{\mathbf{v}}_i = \mathbf{W}_{i+1}^\mathsf{T} \mathbf{v}_{i+1}, \forall i \in \{0, \ldots, n-1\}, \tag{7.27c}$

$$\nu_{i,j} = \begin{cases} 0 & j \in \Gamma_i^- \\ \hat{\nu}_{i,j} & j \in \Gamma_i^+ \\ \frac{\hat{u}_{i,j}}{\hat{u}_{i,j} - \hat{\ell}_{i,j}} [\hat{\nu}_{i,j}]_+ + \alpha_{i,j} [\hat{\nu}_{i,j}]_- & j \in \Gamma_i \end{cases},$$

$$\forall i \in \{1, \ldots, n-1\}. \tag{7.27d}$$

The optimal solution of (7.27) provides an upper bound of the primal problem (7.21). If the dual value is smaller than 0, then the problem is satisfiable. The dual network consists of $\mathbf{v}_i$'s and is almost identical to the back-propagation network. The difference is that for nodes $j \in \mathcal{I}_i$, there is the additional free variable $\alpha_{i,j}$ that we can optimize over. One fixed and dual feasible solution of (7.27) is

$$\alpha_{i,j} = \frac{\hat{u}_{i,j}}{\hat{u}_{i,j} - \hat{\ell}_{i,j}}. \tag{7.28}$$

Under the condition, the dual network is linear. We will show in section 8.3 that the resulting dual network is identical to the case where we use parallel relaxation instead of triangle relaxation. Bunel *et al.* proposed two methods to obtain a solution to (7.27) without further relaxation: supergradient ascent and proximal maximization [13]. These two algorithms are not reviewed in this survey. It is also proved in [13] that the dual form in (7.27) is tighter than the dual form in (7.20).

Our implementation is in algorithm 27. The value is computed in `dual_value`. The dual network is computed in `backprop!` layer by layer. For simplicity, we directly use (7.28) in the implementation. The bounds $\hat{\ell}_{i,j}$ and $\hat{u}_{i,j}$ and the sets $\Gamma_i$, $\Gamma_i^-$, and $\Gamma_i^+$ can be computed using different methods, *e.g.*, interval arithmetic introduced in section 4.1 and section 5.4. The original paper computes the bounds by formulating the bounding problem as several optimization problems similar to (7.21). The objectives are $\min z_{i,j}$ and $\max z_{i,j}$ for all $i$ and $j$. Those problems can also be solved by dynamic programming. Moreover, the bounds can be computed inductively layer by layer. The details of the computation will be discussed in FastLin in section 8.3.

## 7.4 Certify

Certify [50] computes over-approximated certificates for a neural network with only one hidden layer. Similar to ConvDual, Certify also takes a hypercube input set $\|\mathbf{x} - \mathbf{x}_0\|_\infty \leq \epsilon$ and a halfspace output set $\mathbf{c}^\mathsf{T} \mathbf{y} \leq d$. It works for any activation function as long as the function is differentiable almost everywhere and its gradient is bounded. For simplicity, we assume that the derivative of the activation function is bounded by 0 and 1, *i.e.*, $\sigma'(\hat{z}) \in [0, 1]$ for any $\hat{z}$. If not, we can always scale it by a factor $\max_{\hat{z}} \sigma'(\hat{z})$ and add the factor in the following derivation. The primal optimization problem considered in Certify is

$$\max_{\|\mathbf{x} - \mathbf{x}_0\|_\infty \leq \epsilon} o(\mathbf{x}) = \mathbf{c}^\mathsf{T} \mathbf{f}(\mathbf{x}) - d. \tag{7.29}$$

Since $o$ is differentiable almost-everywhere, then

$$o(\mathbf{x}) \leq o(\mathbf{x}_0) + \epsilon \max_{\|\mathbf{x} - \mathbf{x}_0\|_\infty \leq \epsilon} \|\nabla o(\mathbf{x})\|_1. \tag{7.30}$$

For a neural network with only one hidden layer, $o(\mathbf{x}) = \sum_j c_j \sigma_{1,j}(\mathbf{w}_{1,j} \mathbf{x} + b_{1,j})$. For

```julia
struct ConvDual end

function solve(solver::ConvDual, problem::Problem)
    o = dual_value(solver, problem.network, problem.input, problem.output)
    if o >= 0.0
        return BasicResult(:holds)
    end
    return BasicResult(:unknown)
end

function dual_value(solver::ConvDual, network, input, output)
    layers = network.layers
    L, U  = get_bounds(network, input.center, input.radius[1])
    ν₀, d = tosimplehrep(output)
    ν = vec(ν₀)
    o = d[1]
    for i in reverse(1:length(layers))
        o -= ν'*layers[i].bias
        ν = layers[i].weights'*ν
        if i>1
            o += backprop!(ν, U[i-1], L[i-1])
        end
    end
    o -= input.center' * ν + input.radius[1] * sum(abs.(ν))
    return o
end

function backprop!(ν, u, l)
    o = 0.0
    for j in 1:length(ν)
        val = relaxed_relu_gradient(l[j], u[j])
        if val < 1.0
            ν[j] = ν[j] * val
            o += ν[j] * l[j]
        end
    end
    return o
end
```

**Algorithm 27:** ConvDual. The dual problem of the verification problem is considered. Triangle relaxation is applied. The dual problem is formulated by backward dynamic programing. The function `dual_value` explicitly computes the dual value equation (7.27) under the relaxation equation (7.28).

simplicity, we omit the first index (index for layer 1) in $\sigma$, $\mathbf{w}$ and $b$. Then

$$\nabla o(\mathbf{x}) = \sum_j c_j \sigma'_j(\mathbf{w}_j \mathbf{x} + b_j)\mathbf{w}_j^\mathsf{T} = \mathbf{W}^\mathsf{T} \mathrm{diag}(\mathbf{c})\nabla\boldsymbol{\sigma}(\mathbf{W}\mathbf{x} + \mathbf{b}), \tag{7.31}$$

where $\nabla\boldsymbol{\sigma} \in [0,1]^{k_1}$ is the vertical stack of $\sigma'_j$. Let $\mathbf{s} = \sigma'(\mathbf{W}\mathbf{x} + \mathbf{b})$. Then

$$\|\nabla o(\mathbf{x})\|_1 \leq \max_{\mathbf{s} \in [0,1]^{k_1}, \mathbf{t} \in [-1,1]^{k_0}} \mathbf{t}^\mathsf{T} \mathbf{W}^\mathsf{T} \mathrm{diag}(\mathbf{c}) \mathbf{s}, \tag{7.32a}$$

$$= \max_{\mathbf{s} \in [-1,1]^{k_1}, \mathbf{t} \in [-1,1]^{k_0}} \frac{1}{2}\mathbf{t}^\mathsf{T} \mathbf{W}^\mathsf{T} \mathrm{diag}(\mathbf{c})(\mathbf{1} + \mathbf{s}), \tag{7.32b}$$

$$= \max_{\mathbf{p} \in [-1,1]^{1+k_1+k_0}} \frac{1}{4}\mathbf{p}^\mathsf{T} \underbrace{\begin{bmatrix} 0 & \mathbf{0} & \mathbf{1}^\mathsf{T}\mathbf{W}^\mathsf{T}\mathrm{diag}(\mathbf{c}) \\ \mathbf{0} & \mathbf{0} & \mathbf{W}^\mathsf{T}\mathrm{diag}(\mathbf{c}) \\ \mathrm{diag}(\mathbf{c})\mathbf{W}\mathbf{1} & \mathrm{diag}(\mathbf{c})\mathbf{W} & \mathbf{0} \end{bmatrix}}_{\mathbf{M}(\mathbf{c},\mathbf{W})} \mathbf{p}, \tag{7.32c}$$

$$= \max_{\mathbf{p} \in [-1,1]^{1+k_1+k_0}} \frac{1}{4}\langle \mathbf{M}(\mathbf{c},\mathbf{W}), \mathbf{p}\mathbf{p}^\mathsf{T}\rangle, \tag{7.32d}$$

$$\leq \max_{\mathbf{P}\succeq \mathbf{0}, P_{jj}\leq 1} \frac{1}{4}\langle \mathbf{M}(\mathbf{c},\mathbf{W}), \mathbf{P}\rangle. \tag{7.32e}$$

The inequality in (7.32a) is due to $\|\mathbf{q}\|_1 \leq \max_{\|\mathbf{p}\|_\infty \leq 1} \mathbf{p}^\mathsf{T}\mathbf{q}$. (7.32b) changes the range of $\mathbf{s}$. (7.32c) packs all variables into one vector $\mathbf{p} := [1, \mathbf{t}, \mathbf{s}]$ and packs all parameters into the matrix $\mathbf{M}(\mathbf{c}, \mathbf{W})$. (7.32d) exploits the equivalence between matrix trace and quadratic form. The inner product between two matrices is defined as $\langle \mathbf{M}, \mathbf{P}\rangle := \mathrm{tr}(\mathbf{M}^\mathsf{T}\mathbf{P})$. (7.32e) is obtained by taking $\mathbf{P} := \mathbf{p}\mathbf{p}$. The matrix $\mathbf{P}$ is positive semidefinite, *i.e.*, $\mathbf{P}^\mathsf{T} \succeq 0$. Moreover, the diagonal term $P_{jj} = p_j^2 \leq 1$ for $j \in \{1, \ldots, 1 + k_0 + k_1\}$. Hence, the convex semidefinite relaxation of the value $o(\mathbf{x})$ is

$$\max_{\|\mathbf{x}-\mathbf{x}_0\|_\infty \leq \epsilon} o(\mathbf{x}) \leq o(\mathbf{x}_0) + \frac{\epsilon}{4} \max_{\mathbf{P}\succeq \mathbf{0}, P_{jj}\leq 1} \langle \mathbf{M}(\mathbf{c},\mathbf{W}), \mathbf{P}\rangle. \tag{7.33}$$

The right-hand side of (7.33) provides an upper bound of (7.29). If the bound is smaller than zero, then the problem is satisfiable. It is worth noting that the semidefinite program (the max function) only depends on $\mathbf{c}$ and $\mathbf{W}$. As it does not depend on the input $\mathbf{x}$, it only needs to be computed once for a problem.

Our implementation is shown in algorithm 28, which directly constructs and solves the semidefinite program in (7.33). Although the semidefinite relaxation approach can only deal with networks with one-hidden layer, Fazlyab *et al.* introduced a semidefinite programming formulation that can verify networks with multiple layers [24], which draws insight from robust control [11].

# 8 Search and Reachability

This chapter discusses methods that combine layer-by-layer reachability analysis with search. ReluVal [65] uses symbolic intervals for reachability analysis and then searches the input

```julia
struct Certify
    optimizer
end

function solve(solver::Certify, problem::Problem)
    model = Model(solver)
    c, d = tosimplehrep(problem.output)
    c, d = c[1, :], d[1]
    v = c' * problem.network.layers[2].weights
    W = problem.network.layers[1].weights
    M = get_M(v[1, :], W)
    n = size(M, 1)
    P = @variable(model, [1:n, 1:n], PSD)
    output = c' * compute_output(problem.network, problem.input.center) - d
    epsilon = maximum(problem.input.radius)
    o = output + epsilon/4 * tr(M*P)
    @constraint(model, diag(P) .<= ones(n))
    @objective(model, Max, o)
    optimize!(model)
    if value(o) <= 0
        return BasicResult(:holds)
    else
        return BasicResult(:violated))
    end
end
```

**Algorithm 28:** Certify. Solves the semidefinite program equation (7.33).

domain for potential violations using iterative interval refinement. Neurify [64] uses symbolic linear relaxation for reachability analysis and then performs the search by direct constraint refinement. FastLin [66] uses network approximation for reachability analysis and then uses binary search to estimate a certified lower bound of maximum allowable disturbance. FastLip [66] is built upon FastLin, which further estimates the local Lipschitz constant. DLV [30] performs layer-by-layer search in hidden layers for potential counter examples. Note that the return types of these methods are not necessarily reachability results.

The discussion in this chapter focuses on how reachability analysis is combined with search. Recent work studies the efficiency of different search strategies [6, 5].[1]

## 8.1 ReluVal

ReluVal [65] takes a hyperrectangle input set $|\mathbf{x}-\mathbf{x}_0| \leq \mathbf{r}$ and any output set that implements the abstract polytope type. The reachability analysis is done symbolically, while the search is done through iterative interval refinement. Our implementation is shown in algorithms 29 to 31.

```
struct SymbolicInterval{F<:AbstractPolytope}
    Low::Matrix{Float64}
    Up::Matrix{Float64}
    domain::F
end

struct SymbolicIntervalGradient{F<:AbstractPolytope, N<:Real}
    sym::SymbolicInterval{F}
    LΛ::Vector{Vector{N}}
    UΛ::Vector{Vector{N}}
end

const SymbolicIntervalMask = SymbolicIntervalGradient{<:Hyperrectangle, Bool}
```

**Algorithm 29:** The data structure in ReluVal. `SymbolicInterval{Hyperrectangle}` represents the symbolic interval defined in equation (8.1). `SymbolicIntervalMask` further includes the binary lower and upper bounds of $\nabla \boldsymbol{\sigma}_i$, *i.e.*, diagonal entries of the gradient bounds $\underline{\boldsymbol{\Lambda}}_i$ and $\overline{\boldsymbol{\Lambda}}_i$ introduced in section 4.3.

**Symbolic interval propagation** Reachability methods that use interval arithmetic usually provide very loose bounds, as they do not keep track of dependencies among the hidden nodes. Symbolic interval propagation can provide tighter bounds by keeping track of those dependancies layer by layer. Define the extended input as $\mathbf{x}^e := [\mathbf{x}, 1]$. Then a symbolic interval at layer $i$ is defined as

$$\mathbf{z}_i \in [\mathbf{L}_i \mathbf{x}^e, \mathbf{U}_i \mathbf{x}^e], \text{ for } \mathbf{x} \in [\mathbf{x}_0 - \mathbf{r}, \mathbf{x}_0 + \mathbf{r}], \tag{8.1}$$

where $\mathbf{L}_i, \mathbf{U}_i \in \mathbb{R}^{k_i \times (k_0+1)}$ are coefficients in the symbolic interval. For example, the $j$th node at layer $i$ is bounded by

$$\boldsymbol{\ell}_{i,j} \mathbf{x}^e \leq z_{i,j} \leq \mathbf{u}_{i,j} \mathbf{x}^e, \tag{8.2}$$

where $\boldsymbol{\ell}_{i,j}$ and $\mathbf{u}_{i,j}$ are the $j$th row in $\mathbf{L}_i$ and $\mathbf{U}_i$ respectively. Note that both the symbolic lower bound and the symbolic upper bound correspond to star sets (2.10) where the free parameter is $\mathbf{x}$ and the generators are the columns in $\mathbf{L}_i$ or $\mathbf{U}_i$.

---

[1] The search strategies can range from depth first search to breadth first search or from abstract to concrete, *i.e.*, counter example-guided abstraction refinement (CEGAR) or from concrete to abstract, *i.e.*, execution-guided overapproximation (EGO).

The data structure `SymbolicInterval` in algorithm 29 is introduced to keep track of symbolic intervals, where the field `Low` corresponds to $\mathbf{L}_i$, `Up` to $\mathbf{U}_i$. The `domain` can be any abstract polytope. In ReluVal, the domain is limited to the hyperrectangle $[\mathbf{x}_0 - \mathbf{r}, \mathbf{x}_0 + \mathbf{r}]$. In Neurify, the domain can be an H-polytope.

Given the symbolic interval with hyperrectangle domain, $\mathbf{L}_i \mathbf{x}^e$ belongs to a hyperrectangle centered at $\mathbf{L}_i [\mathbf{x}_0, 1]$ with radius $|\mathbf{L}_i|[\mathbf{r}, 0]$, where $|\mathbf{L}_i|$ is a matrix containing the element-wise absolute values of $\mathbf{L}_i$. Similarly, $\mathbf{U}_i \mathbf{x}^e$ belongs to a hyperrectangle centered at $\mathbf{U}_i [\mathbf{x}_0, 1]$ with radius $|\mathbf{U}_i|[\mathbf{r}, 0]$. Let $\mathbf{a} \mathbf{x}^e$ be a symbolic representation for a scalar variable, where $\mathbf{a} \in \mathbb{R}^{k_0+1}$. Let $\underline{h}, \overline{h} : \mathbb{R}^{k_0+1} \to \mathbb{R}$ be functions that map the symbolic representation $\mathbf{a} \mathbf{x}^e$ to its concrete lower bound and upper bound respectively. Recall from (4.1), these two functions $\underline{h}$ and $\overline{h}$ solve two optimizations (4.4). Since the domain is a hyperrectangle, we have analytical solutions

$$\underline{h}(\mathbf{a}) := \mathbf{a}[\mathbf{x}_0, 1] - |\mathbf{a}|[\mathbf{r}, 0], \tag{8.3a}$$
$$\overline{h}(\mathbf{a}) := \mathbf{a}[\mathbf{x}_0, 1] + |\mathbf{a}|[\mathbf{r}, 0]. \tag{8.3b}$$

The symbolic intervals are computed layer by layer by calling the function `forward_network` in algorithm 9. The function `forward_layer` is shown in algorithm 30, which consists of the following two steps:

- Symbolic interval propagation through the following linear map $\hat{\mathbf{z}}_i = \mathbf{W}_i \mathbf{z}_{i-1} + \mathbf{b}_i$ for $i \in \{2, \ldots, n\}$.

$$\hat{\mathbf{L}}_i = [\mathbf{W}_i]_+ \mathbf{L}_{i-1} + [\mathbf{W}_i]_- \mathbf{U}_{i-1} + [\mathbf{0}\ \mathbf{b}_i], \tag{8.4a}$$
$$\hat{\mathbf{U}}_i = [\mathbf{W}_i]_+ \mathbf{U}_{i-1} + [\mathbf{W}_i]_- \mathbf{L}_{i-1} + [\mathbf{0}\ \mathbf{b}_i]. \tag{8.4b}$$

For the first layer, the symbolic bounds are defined as horizontal concatenation of $\mathbf{W}_1$ and $\mathbf{b}_1$,

$$\hat{\mathbf{L}}_1 = \hat{\mathbf{U}}_1 = [\mathbf{W}_1\ \mathbf{b}_1]. \tag{8.5}$$

This corresponds to `forward_linear` in algorithm 30.

- Symbolic interval propagation through the ReLU activation function $\mathbf{z}_i = [\hat{\mathbf{z}}_i]_+$. For each node $j$, there are three possibilities: always active ($j \in \Gamma_i^+$), never active ($j \in \Gamma_i^-$), undetermined ($j \in \Gamma_i$). Similar to (4.14), we have

$$\Gamma_i^+ = \{j : \underline{h}(\hat{\boldsymbol{\ell}}_{i,j}) \geq 0\}, \tag{8.6a}$$
$$\Gamma_i^- = \{j : \overline{h}(\hat{\mathbf{u}}_{i,j}) \leq 0\}, \tag{8.6b}$$
$$\Gamma_i = \{j : j \notin \Gamma_i^+ \cup \Gamma_i^-\}. \tag{8.6c}$$

Then the symbolic interval for node $j$ is computed as

$$j \in \Gamma_i^+ \Rightarrow \boldsymbol{\ell}_{i,j} = \hat{\boldsymbol{\ell}}_{i,j}, \mathbf{u}_{i,j} = \hat{\mathbf{u}}_{i,j}, \tag{8.7a}$$
$$j \in \Gamma_i^- \Rightarrow \boldsymbol{\ell}_{i,j} = \mathbf{u}_{i,j} = \mathbf{0}, \tag{8.7b}$$
$$j \in \Gamma_i \Rightarrow \boldsymbol{\ell}_{i,j} = \mathbf{0}, \mathbf{u}_{i,j} = \begin{cases} \hat{\mathbf{u}}_{i,j} & \text{if } \underline{h}(\hat{\mathbf{u}}_{i,j}) \geq 0 \\ \begin{bmatrix} \mathbf{0} \ \overline{h}(\hat{\mathbf{u}}_{i,j}) \end{bmatrix} & \text{if } \underline{h}(\hat{\mathbf{u}}_{i,j}) < 0 \end{cases}. \tag{8.7c}$$

When the node is always active in (8.7a), we keep the symbolic dependency on the input variables. When the node is never active in (8.7b), all outputs should be 0. When the activation is undetermined in (8.7c), the lower bound is set to $\mathbf{0}$. If it is possible for the symbolic upper bound to be smaller than 0, the input dependencies will be thrown away. The upper bound is set to be its concrete maximum, *i.e.*, the first $k_0$ entries in $\mathbf{u}_{i,j}$ are set to zero. Otherwise, we keep the symbolic dependency of the upper bound on input variables.

This corresponds to `forward_act` in algorithm 30.

Given the symbolic interval propagation, the output reachable set is a hyper-rectangle such that

$$\tilde{\mathcal{R}} = \{\mathbf{y} : y_j \in [\underline{h}(\boldsymbol{\ell}_{n,j}), \overline{h}(\mathbf{u}_{n,j})], \forall j = 1, \ldots, k_n\}. \tag{8.8}$$

The output reachable set computed by symbolic interval propagation is tighter than simple interval arithmetic as illustrated in figure 5.3.

Given the reachable set $\tilde{\mathcal{R}}$, its relationship with respect to the output set $\mathcal{Y}$ is examined in the function `check_inclusion` in algorithm 31, which may return the following four different results:

- The return status is `:holds` if $\tilde{\mathcal{R}} \subset \mathcal{Y}$.

- The return status is `:violated` if $\tilde{\mathcal{R}} \cap \mathcal{Y} = \emptyset$.

- In addition, we sample the input interval to check for counter examples.[2] If the output with respect to any sample point does not belong to $\mathcal{Y}$, the sample point is returned as an unsatisfied `CounterExampleResult`.

---

[2]Heuristically, only the middle point of the input interval is checked in both the original implementation and our implementation.

- Otherwise, the return status is `:unknown`.

If the status is undetermined, ReluVal performs iterative interval refinement to minimize over-approximation in $\tilde{\mathcal{R}}$.

**Iterative interval refinement**  Although symbolic interval propagation can provide us with tighter bounds than simple interval arithmetic, it may still have significant over-approximation, especially when the input intervals are comparably large. Recall that MaxSens partitions the input intervals into smaller sets to minimize over approximations. ReluVal performs iterative interval refinement instead, which splits the intervals of input nodes according to their influence on the output.[3]

We evaluate the influence by considering the bounds on gradients $\underline{\mathbf{G}}_n$ and $\overline{\mathbf{G}}_n$ defined in chapter 4, which essentially measure the sensitivity of the output with respect to each input feature. The bounds $\underline{\mathbf{G}}_n$ and $\overline{\mathbf{G}}_n$ can be obtained by calling the third `get_gradient` in algorithm 8. The bounds $\underline{\mathbf{\Lambda}}_i$ and $\overline{\mathbf{\Lambda}}_i$ on $\nabla \boldsymbol{\sigma}_i$ are computed in the forward propagation, and recorded in the data structure `SymbolicIntervalGradient` in algorithm 29.

For each refinement step, ReluVal bisects the interval for input node $j$ that has the highest smear value

$$S_j \coloneqq \sum_k \max\{|UG_{n,j,k}|, |LG_{n,j,k}|\} r_k, \tag{8.9}$$

where $UG_{n,j,k}$ and $LG_{n,j,k}$ are the $j$th row and $k$th column entries in $\overline{\mathbf{G}}_n$ and $\underline{\mathbf{G}}_n$, respectively. Let $j^* = \arg\max_j S_j$. The smear values are computed in `get_smear_index` in algorithm 31, which returns the index to split. The split is performed by `interval_split` in algorithm 7.

The main procedure is shown in algorithm 31 and illustrated in figure 8.1. We first compute the reachable set without splitting the interval. If the status of the reachability analysis is undetermined, iterative interval refinement will be performed. A list of reachable sets is maintained. The reachable sets correspond to different input intervals. At each iteration, a set in the list is picked out according to the tree search strategy specified by the solver (default depth first search). For the picked set, the bounds of its gradient are computed. Then it is split into two intervals according to the smear values. The reachable sets for the two smaller intervals are then computed. If a reachable set belongs to the output constraint, we drop the corresponding interval. If a counter example is found in an interval, we conclude that the problem is violated. Otherwise, we need to further split the interval, whose reachable set is then pushed back to the list. If the list becomes empty, the property is verified to hold.

---

[3] The ReluVal paper discusses two approaches: baseline iterative refinement and optimizing iterative refinement. We consider optimizing iterative refinement.

```julia
function forward_layer(solver::ReluVal, layer::Layer, input)
    return forward_act(solver, forward_linear(solver, input, layer), layer)
end

function forward_linear(solver::ReluVal, input::SymbolicIntervalMask, layer)
    sym = input.sym
    output_Low, output_Up = interval_map(layer.weights, sym.Low, sym.Up)
    output_Up[:, end] += layer.bias
    output_Low[:, end] += layer.bias
    sym = SymbolicInterval(output_Low, output_Up, domain(input))
    return SymbolicIntervalGradient(sym, input.LΛ, input.UΛ)
end

function forward_act(::ReluVal, input::SymbolicIntervalMask, layer)
    output_Low, output_Up = copy(input.sym.Low), copy(input.sym.Up)
    n_node = n_nodes(layer)
    LΛᵢ, UΛᵢ = falses(n_node), trues(n_node)
    for j in 1:n_node
        if upper_bound(upper(input), j) <= 0
            LΛᵢ[j], UΛᵢ[j] = 0, 0
            output_Low[j, :] .= 0
            output_Up[j, :] .= 0
        elseif lower_bound(lower(input), j) >= 0
            LΛᵢ[j], UΛᵢ[j] = 1, 1
        else
            LΛᵢ[j], UΛᵢ[j] = 0, 1
            output_Low[j, :] .= 0
            if lower_bound(upper(input), j) < 0
                output_Up[j, :] .= 0
                output_Up[j, :][end] = upper_bound(upper(input), j)
            end
        end
    end
    sym = SymbolicInterval(output_Low, output_Up, domain(input))
    LΛ = push!(input.LΛ, LΛᵢ)
    UΛ = push!(input.UΛ, UΛᵢ)
    return SymbolicIntervalGradient(sym, LΛ, UΛ)
end
```

**Algorithm 30:** Symbolic interval propagation in ReluVal. The function `forward_layer` is called by `forward_network` in the layer-by-layer propagation. The function `forward_layer` consists of `forward_linear` for the linear mapping and `forward_act` for the nonlinear activation.

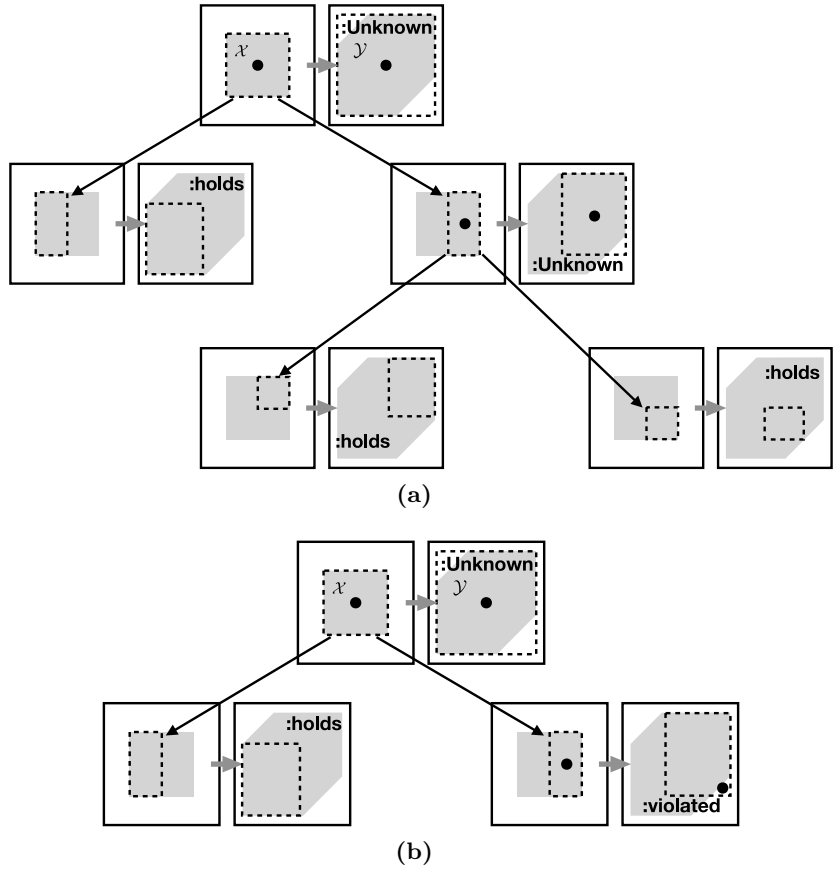

**Figure 8.1:** Illustration of iterative interval refinement. The two subfigures illustrate two search trees in two different scenarios. Each node in the search tree consists of two boxes connected by an arrow. The left box corresponds to the input space and the right box corresponds to the output space. The input constraint and the output constraint are shaded in corresponding spaces. The dashed box in the input space represents the refined interval, while the dashed box in the output space represents the reachable set for that interval. If the reachable set belongs to $\mathcal{Y}$, then the status is holds. If the reachable set overlaps with $\mathcal{Y}$ but is not a subset of $\mathcal{Y}$, the status is unknown. In this case, a point is sampled in the input space (black dot). If the output of that sample point does not belong to $\mathcal{Y}$, the status is violated. Otherwise, the interval is split into two. (a) Scenario 1: property holds as the reachable sets all belong to $\mathcal{Y}$. (b) Scenario 2: property violated as a counter example is found.

```julia
struct ReluVal
    max_iter::Int64
    tree_search::Symbol
end

function solve(solver::ReluVal, problem::Problem)
    reach_list = []
    interval, nnet = problem.input, problem.network
    for i in 1:solver.max_iter
        if i > 1
            interval = select!(reach_list, solver.tree_search)
        end
        reach = forward_network(solver, nnet, init_symbolic_mask(interval))
        result = check_inclusion(reach.sym, problem.output, nnet)
        if result.status === :violated
            return result
        elseif result.status === :unknown
            intervals = bisect_interval_by_max_smear(nnet, reach)
            append!(reach_list, intervals)
        end
        isempty(reach_list) && return CounterExampleResult(:holds)
    end
    return CounterExampleResult(:unknown)
end

function check_inclusion(reach::SymbolicInterval{<:Hyperrectangle},
                        output, nnet)
    reachable = Hyperrectangle(low = low(reach), high = high(reach))
    issubset(reachable, output) && return CounterExampleResult(:holds)
    middle_point = center(domain(reach))
    y = compute_output(nnet, middle_point)
    y ∈ output || return CounterExampleResult(:violated, middle_point)
    return CounterExampleResult(:unknown)
end
```

**Algorithm 31:** The main function in ReluVal. The reachable set for a given input interval is computed by calling `forward_network`, which then calls `forward_layer`. In order to perform iterative interval refinement, the solve function keeps track of a list of reachable sets that correspond to different input intervals. At each iteration, a set in the list is picked out according to the tree search strategy specified by the solver by `pick_out!`. For the picked set, the bounds of its gradient are computed by `get_gradient`. Then the index to be split is determined by `get_smear_index`. The interval is split into two by `split_interval`. The reachable sets for the two smaller intervals are then computed. The `check_inclusion` checks the status of each reachable set. If a counter example is found, the result is directly returned. If the status is unknown, we need to further split the input interval. The corresponding reachable set is pushed back to the list. If the list becomes empty, the problem is verified.

## 8.2 Neurify

Neurify [64] is a direct extension of ReluVal. It takes a polytope input set and any output set that implements the abstract polytope type. The algorithm follows the same structure as in ReluVal where the reachability analysis is done symbolically and the search is done through iterative interval refinement. There are two major differences between Neurify and ReluVal. The first difference is that Neurify uses a tighter concretization method for symbolic analysis, called *symbolic linear relaxation*. The second difference is that Neurify uses *direct constraint refinement* to iteratively refine the over approximated sets. The data structures used in Neurify are `SymbolicInterval{HPolytope}` and `SymbolicIntervalGradient{HPolytope, Float64}`.

**Symbolic linear relaxation**  Symbolic linear relaxation follows the procedures for symbolic interval propagation discussed in section 8.1. The difference exists in the symbolic interval propagation through the ReLU activation. Triangle relaxation is used for both the symbolic lower bound and the symbolic upper bound. For the symbolic lower bound, we have

$$\boldsymbol{\ell}_{i,j} = \begin{cases} \hat{\boldsymbol{\ell}}_{i,j} & \text{if } \underline{h}(\hat{\boldsymbol{\ell}}_{i,j}) \geq 0 \\ \mathbf{0} & \text{if } \bar{h}(\hat{\boldsymbol{\ell}}_{i,j}) \leq 0 \\ \frac{\bar{h}(\hat{\boldsymbol{\ell}}_{i,j})}{\bar{h}(\hat{\boldsymbol{\ell}}_{i,j}) - \underline{h}(\hat{\boldsymbol{\ell}}_{i,j})} \hat{\boldsymbol{\ell}}_{i,j} & \text{if } \underline{h}(\hat{\boldsymbol{\ell}}_{i,j}) < 0 < \bar{h}(\hat{\boldsymbol{\ell}}_{i,j}) \end{cases}. \tag{8.10}$$

For the symbolic upper bound, we have

$$\mathbf{u}_{i,j} = \begin{cases} \hat{\mathbf{u}}_{i,j} & \text{if } \underline{h}(\hat{\mathbf{u}}_{i,j}) \geq 0 \\ \mathbf{0} & \text{if } \bar{h}(\hat{\mathbf{u}}_{i,j}) \leq 0 \\ \frac{\bar{h}(\hat{\mathbf{u}}_{i,j})}{\bar{h}(\hat{\mathbf{u}}_{i,j}) - \underline{h}(\hat{\mathbf{u}}_{i,j})} (\hat{\mathbf{u}}_{i,j} - \underline{h}(\hat{\mathbf{u}}_{i,j})) & \text{if } \underline{h}(\hat{\mathbf{u}}_{i,j}) < 0 < \bar{h}(\hat{\mathbf{u}}_{i,j}) \end{cases}. \tag{8.11}$$

The lower and upper bounds $\underline{h}(\cdot)$ and $\bar{h}(\cdot)$ for symbolic intervals with polytope domains are computed using linear programming as defined in (4.4). For simplicity, define

$$\tilde{\lambda}(\hat{\boldsymbol{\ell}}_{i,j}) = \frac{\bar{h}(\hat{\boldsymbol{\ell}}_{i,j})}{\bar{h}(\hat{\boldsymbol{\ell}}_{i,j}) - \underline{h}(\hat{\boldsymbol{\ell}}_{i,j})}, \quad \tilde{\lambda}(\hat{\mathbf{u}}_{i,j}) = \frac{\bar{h}(\hat{\mathbf{u}}_{i,j})}{\bar{h}(\hat{\mathbf{u}}_{i,j}) - \underline{h}(\hat{\mathbf{u}}_{i,j})} \tag{8.12}$$

Note the third cases in (8.10) and (8.11) are different. The third case in (8.10) is a lower bound of the symbolic lower bound, while the third case in (8.11) is an upper bound of the symbolic upper bound. By comparing (8.10) and (8.11) with the symbolic interval propagation (8.7c) in ReluVal, it is easy to see that the only difference exists in the concretization step for nodes with undetermined ReLU activation status (i.e., node $j \in \Gamma_i$ in layer $i$) which also satisfy $\underline{h}(\hat{\boldsymbol{\ell}}_{i,j}) < 0 < \bar{h}(\hat{\boldsymbol{\ell}}_{i,j})$ and $\underline{h}(\hat{\mathbf{u}}_{i,j}) < 0 < \bar{h}(\hat{\mathbf{u}}_{i,j})$. In ReluVal, the symbolic interval for node $j \in \Gamma_i$ is concretized as shown in (8.7c), which may result in big over-approximation. For better comparison, we rewrite the conditions in (8.7) in the following forms:

$$\boldsymbol{\ell}_{i,j} = \begin{cases} \hat{\boldsymbol{\ell}}_{i,j} & \text{if } \underline{h}(\hat{\boldsymbol{\ell}}_{i,j}) \geq 0 \\ \mathbf{0} & \text{if } \bar{h}(\hat{\boldsymbol{\ell}}_{i,j}) \leq 0 \\ \mathbf{0} & \text{if } \underline{h}(\hat{\boldsymbol{\ell}}_{i,j}) < 0 < \bar{h}(\hat{\boldsymbol{\ell}}_{i,j}) \end{cases}, \quad (8.13a)$$

$$\mathbf{u}_{i,j} = \begin{cases} \hat{\mathbf{u}}_{i,j} & \text{if } \underline{h}(\hat{\mathbf{u}}_{i,j}) \geq 0 \\ \mathbf{0} & \text{if } \bar{h}(\hat{\mathbf{u}}_{i,j}) \leq 0 \\ \begin{bmatrix} \mathbf{0} \ \bar{h}(\hat{\mathbf{u}}_{i,j}) \end{bmatrix} & \text{if } \underline{h}(\hat{\mathbf{u}}_{i,j}) < 0 < \bar{h}(\hat{\mathbf{u}}_{i,j}) \end{cases}. \quad (8.13b)$$

The relaxed set $[\boldsymbol{\ell}_{i,j}, \mathbf{u}_{i,j}]$ in the third cases in (8.10) and (8.11) is tighter than the concretized set $[0, \bar{h}(\hat{\mathbf{u}}_{i,j})]$ in the third cases in (8.13). The symbolic linear relaxation keeps part of the input dependencies during interval propagation, while the concretization in (8.13) drops the dependencies. It is reported in [64] that the symbolic interval relaxation cuts 59.64% more over-approximation error than the symbolic interval propagation in ReluVal. However, it is worth noting that the set $[\boldsymbol{\ell}_{i,j}, \mathbf{u}_{i,j}]$ obtained from symbolic linear relaxation is not a subset of the concretized set $[0, \bar{h}(\hat{\mathbf{u}}_{i,j})]$. The upper bound is valid, *i.e.*, any concrete value of $\tilde{\lambda}(\hat{\mathbf{u}}_{i,j})(\hat{\mathbf{u}}_{i,j} - \underline{h}(\hat{\mathbf{u}}_{i,j}))$ is smaller than $\bar{h}(\hat{\mathbf{u}}_{i,j})$. However, it is possible for the concrete value of $\tilde{\lambda}(\hat{\boldsymbol{\ell}}_{i,j})\hat{\boldsymbol{\ell}}_{i,j}$ to be smaller than zero.

The symbolic linear relaxation is implemented in `forward_act` in algorithm 32. Similar to ReluVal, during the forward symbolic propagation, the bounds on $\nabla \boldsymbol{\sigma}_i$ are computed for subsequent backward analysis. The bounds on $\nabla \boldsymbol{\sigma}_i$ in ReluVal follow (4.15), while the bounds on $\nabla \boldsymbol{\sigma}_i$ in Neurify follow (4.16) due to triangle relaxation.

Given the symbolic linear relaxation, the symbolic output reachable set is $[\boldsymbol{\ell}_n, \mathbf{u}_n]$. We may directly concretize the symbolic output reachable set and compare it with output constraint $\mathcal{Y}$ as in ReluVal. However, the concretization uses a hyperrectangle to bound the symbolic reachable set, which will introduce more over approximations. Instead of concretization, Neurify checks the satisfiability of the output constraint using optimization. Suppose the output constraint is a H-polytope $\mathbf{Cy} \leq \mathbf{d}$, which consists of $K$ halfspace constraints $\mathbf{c}_k \mathbf{y} \leq d_k$ for $k \in \{1, \ldots, K\}$. Then we check the satisfiability of all halfspace constraints by solving the following linear program for all $k$

$$\max_{\mathbf{x} \in \mathcal{X}} [\mathbf{c}_k]_+ \mathbf{u}_n + [\mathbf{c}_k]_- \boldsymbol{\ell}_n - d_k. \quad (8.14)$$

The optimal value of (8.14) is denoted $o_k^*$ while the decision variable that attains the optimal value is denoted $\mathbf{x}_k^*$. If $o_k^* \leq 0$, then the output constraint is satisfied. If $o_k^* > 0$, then we need to check whether there is truly a violation or it is a false alarm due to over approximation of the symbolic bounds. If $\mathbf{f}(\mathbf{x}_k^*) \notin \mathcal{Y}$, then $x_k^*$ is returned as an unsatisfied `CounterExampleResult`; otherwise, we cannot draw any conclusion of the constraint satisfiability before further constraint refinement. If all $K$ constraints are satisfied, then

```
function forward_act(::Neurify, input::SymbolicIntervalGradient, layer)
    n_node = n_nodes(layer)
    output_Low, output_Up = copy(input.sym.Low), copy(input.sym.Up)
    LΛᵢ, UΛᵢ = zeros(n_node), ones(n_node)
    for j in 1:n_node
        up_low, up_up = bounds(upper(input), j)
        low_low, low_up = bounds(lower(input), j)
        up_slope = relaxed_relu_gradient(up_low, up_up)
        low_slope = relaxed_relu_gradient(low_low, low_up)
        output_Up[j, :] .*= up_slope
        output_Up[j, end] += up_slope * max(-up_low, 0)
        output_Low[j, :] .*= low_slope
        LΛᵢ[j], UΛᵢ[j] = low_slope, up_slope
    end
    sym = SymbolicInterval(output_Low, output_Up, domain(input))
    LΛ = push!(input.LΛ, LΛᵢ)
    UΛ = push!(input.UΛ, UΛᵢ)
    return SymbolicIntervalGradient(sym, LΛ, UΛ)
end
```

**Algorithm 32:** Symbolic linear relaxation in Neurify. The function `forward_act` implements (8.10) and (8.11). The concrete lower and upper bounds of the symbolic interval is computed in `bounds` using linear programming as defined in (4.4). The bounds on $\nabla \sigma_i$ is computed in `relaxed_relu_graident` which implements (4.16).

the solver returns `:holds`. If there is some constraint that is not satisfied and no counter example is found, the solver returns `:unknown`. The index of the most violated constraint, *i.e.*, $k^* = \arg\max_k o_k^*$, will be recorded and used to guide the constraint refinement to be discussed below. The optimization-based constraint checking is implemented in algorithm 33.

**Direct constraint refinement** Direct constraint refinement splits the intermediate nodes that operate in the nonlinear region of the ReLU activation function, which is more efficient than the naive interval refinement which simply split input nodes.

The direct constraint refinement first uses influence analysis to choose the node to split. The influence analysis will only check overestimated nodes that operate in the nonlinear region of the ReLU activation function, *i.e.*, either $\underline{h}(\hat{\ell}_{i,j}) < 0 < \bar{h}(\hat{\ell}_{i,j})$ or $\underline{h}(\hat{\mathbf{u}}_{i,j}) < 0 < \bar{h}(\hat{\mathbf{u}}_{i,j})$. Among all overestimated nodes, we choose the one that has the largest influence of the verification result. The influence of a node (denoted $I_{i,j}$) is determined by the gradient from the most violated constraint $k^*$ to the node multiplying the range of the node, *i.e.*,

$$I_{i,j} := \max\{|UG_{k^*,i,j}|, |LG_{k^*,i,j}|\} \left[\bar{h}(\mathbf{u}_{i,j}) - \underline{h}(\ell_{i,j})\right], \tag{8.15}$$

where $UG_{k^*,i,j}$ and $LG_{k^*,i,j}$ are the upper and lower bounds of the gradient from the

```julia
function check_inclusion(solver::Neurify, nnet::Network,
                         reach::SymbolicInterval, output)
    input_domain = domain(reach)
    model = Model(solver); set_silent(model)
    x = @variable(model, [1:dim(input_domain)])
    add_set_constraint!(model, input_domain, x)
    max_violation = 0.0
    max_violation_con = nothing
    for (i, cons) in enumerate(constraints_list(output))
        a, b = cons.a, cons.b
        c = max.(a, 0)'*reach.Up + min.(a, 0)'*reach.Low
        @objective(model, Max, c * [x; 1] - b)
        optimize!(model)
        if termination_status(model) == OPTIMAL
            if compute_output(nnet, value(x)) ∉ output
                return CounterExampleResult(:violated, value(x)), nothing
            end
            viol = objective_value(model)
            if viol > max_violation
                max_violation = viol
                max_violation_con = a
            end
        end
    end
    if max_violation > 0.0
        return CounterExampleResult(:unknown), max_violation_con
    else
        return CounterExampleResult(:holds), nothing
    end
end
```

**Algorithm 33:** The function to check the satisfiability of the output constraint given the symbolic interval in Neurify. The function solves the linear programming (8.14) for all constraints. If a concrete counter example is found, the function returns `:violated` with the counter example. If some constraint is violated but no concrete counter example is found, the function returns `:unknown` with the most violated constraint. Otherwise, the function returns `:hold`.

expression $\mathbf{c}_{k^*}\mathbf{y}$ to the $j$th node in the $i$th layer. The bounds on $\nabla \boldsymbol{\sigma}_i$ computed in the forward pass will be used to compute $UG$ and $LG$ in (8.15). Note the influence is different from the smear value in (8.9). In the smear value, the gradient is computed directly from the output nodes, while the gradients from different output nodes are weighted by their respective output ranges and then summed together. We pick the node that has the largest influence, *i.e.*, $(i^*, j^*) = \arg\max_{i,j} I_{i,j}$, and split the node into the following three conditions:

- Active: $\hat{\boldsymbol{\ell}}_{i^*,j^*} \geq 0$ and $\hat{\mathbf{u}}_{i^*,j^*} \geq 0$;
- Inactive: $\hat{\boldsymbol{\ell}}_{i^*,j^*} \leq 0$ and $\hat{\mathbf{u}}_{i^*,j^*} \leq 0$;
- Both $\hat{\boldsymbol{\ell}}_{i^*,j^*} \leq 0$ and $\hat{\mathbf{u}}_{i^*,j^*} \geq 0$.

Note that in either of the cases, the symbolic linear relaxation in (8.10) and (8.11) will not hit the third condition, hence will not introduce any over approximation. And the three cases fully cover all possible situations of an overestimated node, since we implicitly have the condition that $\hat{\boldsymbol{\ell}}_{i^*,j^*} \leq \hat{\mathbf{u}}_{i^*,j^*}$. The three conditions will split the input set $\mathcal{X}$ into three subsets (called split sets).

For a split set, we compute its symbolic reachable set using symbolic linear relaxation, and check the satisfiability of the symbolic reachable set by solving the linear program in (8.14) for all output constraints. The split sets that get unknown returns (violating the output constraints but no concrete counter example found) will be pushed into the search queue for further splits. The main procedure of Neurify is shown in algorithm 34, which is similar to that of ReluVal.

## 8.3 FastLin

FastLin [66] computes the certified lower bound of the maximum allowable disturbance.[4] The method combines reachability analysis with binary search to estimate the bound, and returns an adversarial result. FastLin only works for ReLU activation functions. A general method that works for any activation function is called CROWN [75], but it is not reviewed here.

The binary search procedure shown in the main function in algorithm 36 can be combined with any reachability method. In particular, FastLin computes the bounds based on linear approximation of the neurons. The method takes a hyperrectangle as input set and a polytope or the complement of a polytope as output set.

---

[4] An earlier work [67] called CLEVER also estimates the lower bound. However, CLEVER is not sound. Hence, it is not reviewed here.

```
struct Neurify
    max_iter::Int64
    tree_search::Symbol
    optimizer = GLPK.Optimizer
end

function solve(solver::Neurify, problem::Problem)
    nnet, output = problem.network, problem.output
    reach_list = []
    domain = init_symbolic_grad(problem.input)
    splits = Set()
    for i in 1:solver.max_iter
        if i > 1
            domain, splits = select!(reach_list, solver.tree_search)
        end
        reach = forward_network(solver, nnet, domain)
        result, max_violation_con = check_inclusion(solver, nnet,
                                    last(reach).sym, output)
        if result.status === :violated
            return result
        elseif result.status === :unknown
            subdomains = constraint_refinement(nnet, reach,
                                    max_violation_con, splits)
            for domain in subdomains
                push!(reach_list, (init_symbolic_grad(domain), copy(splits)))
            end
        end
        isempty(reach_list) && return CounterExampleResult(:holds)
    end
    return CounterExampleResult(:unknown)
end
```

**Algorithm 34:** The main function in Neurify. The reachable set for a given input domain is computed by calling `forward_network`, which then calls `forward_layer`, then `forward_linear` and `forward_act`. In order to perform constraint refinement, the solve function keeps track of a list of reachable sets that correspond to different input domains. At each iteration, a set in the list is picked out according to the tree search strategy specified by the solver by `select!`. For the picked set, `constraint_refinement` first picks a node with largest influence (8.15), then splits the node according to different activation status. The reachable sets for the split domains are then computed. The `check_inclusion` checks the status of each reachable set. If a counter example is found, the result is directly returned. If the status is unknown, we need to further split the input interval. The corresponding reachable set is pushed back to the list together with its maximum violation. If the list becomes empty, the problem is verified.

**Reachability via network relaxation**  Given the bounds from interval arithmetic, a ReLU node that has an undetermined activation can be approximated using parallel relaxation.[5] Given the bounds $\boldsymbol{\ell}_i$ and $\mathbf{u}_i$ for $i \leq k$, the bounds $\hat{\boldsymbol{\ell}}_{k+1}$ and $\hat{\mathbf{u}}_{k+1}$ can be computed by directly optimizing the node values. Similar to the derivation in ConvDual, we derive the lower and upper bounds using backward dynamic programming.[6] The optimization and dynamic programming approach will provide tighter bounds than interval arithmetic discussed in section 4.1, since the correlation among nodes in the hidden layers is considered in the optimization problem, but not in interval arithmetic algorithms.

The bounds are initiated as $\boldsymbol{\ell}_0 = \mathbf{x}_0 - \epsilon \mathbf{1}$ and $\mathbf{u}_0 = \mathbf{x}_0 + \epsilon \mathbf{1}$. Given the bounds $\boldsymbol{\ell}_i$ and $\mathbf{u}_i$ for $i \leq k$, we show the derivation for $\boldsymbol{\ell}_{k+1}$ and $\mathbf{u}_{k+1}$. For simplicity and without loss of generality, we assume that the $k+1$th layer only contains one node. The derivation below can be easily extended to the case with multiple nodes. The lower and upper bounds of node $\hat{z}_{k+1}$ are given by

$$\hat{\ell}_{k+1} = \min_{\|\mathbf{x}-\mathbf{x}_0\|\leq \epsilon} \hat{z}_{k+1}, \hat{u}_{k+1} = \max_{\|\mathbf{x}-\mathbf{x}_0\|\leq \epsilon} \hat{z}_{k+1}. \tag{8.16}$$

For simplicity, we only show the detailed derivation of $\hat{u}_{k+1}$. Recall that $\mathbf{v}_i$ and $\hat{\mathbf{v}}_i = \mathbf{W}_{i+1}\mathbf{v}_{i+1}$ are the dual variables for $\hat{\mathbf{z}}_i$ and $\mathbf{z}_i$, and $\gamma_i$ is the bias in the value function. The boundary conditions are $\nu_{k+1} = 1$ and $\gamma_{k+1} = 0$. Then the equation in (7.22) follows. The sets $\Gamma_i^+$, $\Gamma_i^-$, and $\Gamma_i$ for $i \leq k$ can be constructed using the known bounds according to (4.14). For $j \in \Gamma_i^+$, $\nu_{i,j} = \hat{\nu}_{i,j}$. For $\Gamma_i^-$, $\nu_{i,j} = 0$. These two cases are identical to the cases in ConvDual in section 7.3. For $j \in \Gamma_i$, consider the parallel relaxation (6.7) instead of triangle relaxation, we have

$$\hat{\nu}_{i,j} z_{i,j} \leq \frac{\hat{\nu}_{i,j}\hat{u}_{i,j}}{\hat{u}_{i,j} - \hat{\ell}_{i,j}} \hat{z}_{i,j} - \frac{\hat{\nu}_{i,j}\hat{u}_{i,j}\hat{\ell}_{i,j}}{\hat{u}_{i,j} - \hat{\ell}_{i,j}} \qquad \text{for } \hat{\nu}_{i,j} \geq 0, \tag{8.17a}$$

$$\hat{\nu}_{i,j} z_{i,j} \leq \frac{\hat{\nu}_{i,j}\hat{u}_{i,j}}{\hat{u}_{i,j} - \hat{\ell}_{i,j}} \hat{z}_{i,j} \qquad \text{for } \hat{\nu}_{i,j} < 0, \tag{8.17b}$$

which is different from (7.23) in that there is no degree of freedom for a slack variable $\alpha$. In this way, the relationship among the dual variables that maximizes the value is that for $i \in \{1, \ldots, k\}$,

$$\gamma_i = \mathbf{v}_{i+1}^\mathsf{T} \mathbf{b}_{i+1} + \gamma_{i+1} - \sum_{j \in \Gamma_i} \hat{\ell}_{i,j} [\nu_{i,j}]_+, \tag{8.18a}$$

$$\nu_{i,j} = \begin{cases} \hat{\nu}_{i,j} & j \in \Gamma_i^+ \\ 0 & j \in \Gamma_i^- \\ \frac{\hat{u}_{i,j}}{\hat{u}_{i,j} - \hat{\ell}_{i,j}} \hat{\nu}_{i,j} & j \in \Gamma_i \end{cases}. \tag{8.18b}$$

---

[5]Parallel relaxation is looser than triangle relaxation.

[6]The FastLin paper provides a different derivation.

The resulting dual network defined by $\nu_{i,j}$'s is identical to the dual network in ConvDual when $\alpha_{i,j}$ is chosen according to (7.28). For simplification, define a diagonal matrix $\mathbf{D}_i \in \mathbb{R}^{k_i \times k_i}$ whose diagonal entries $d_{i,j,j}$ for all $j$ satisfy

$$d_{i,j,j} = \begin{cases} 1 & j \in \Gamma_i^+ \\ 0 & j \in \Gamma_i^- \\ \frac{\hat{u}_{i,j}}{\hat{u}_{i,j} - \hat{\ell}_{i,j}} & j \in \Gamma_i \end{cases}. \tag{8.19}$$

Then the dual network satisfies

$$\mathbf{v}_i = \mathbf{D}_i \mathbf{W}_{i+1}^\mathsf{T} \mathbf{v}_{i+1}, \forall i \leq k. \tag{8.20}$$

The dual variables provide an upper bound of $\hat{z}_{k+1}$ such that $\hat{z}_{k+1} \leq \mathbf{v}_1^T \hat{\mathbf{z}}_1 + \gamma_1 = \hat{\mathbf{v}}_0^T \mathbf{x} + \mathbf{v}_1^\mathsf{T} \mathbf{b}_1 + \gamma_1$ where $\mathbf{x}$ satisfies the $\ell_p$ bounds $\|\mathbf{x} - \mathbf{x}_0\|_p \leq \epsilon$. According to (8.18a) and the boundary condition $\gamma_{k+1} = 0$,

$$\mathbf{v}_1^\mathsf{T} \mathbf{b}_1 + \gamma_1 = \sum_{i=1}^{k+1} \mathbf{v}_i^\mathsf{T} \mathbf{b}_i - \sum_{i=1}^{k} \sum_{j \in \Gamma_i} \hat{\ell}_{i,j} [\nu_{i,j}]_+ =: \mu^+. \tag{8.21}$$

Hence, the upper bound of $\hat{z}_{k+1}$ is

$$\hat{u}_{k+1} := \max_{\|\mathbf{x} - \mathbf{x}_0\|_p \leq \epsilon} \hat{\mathbf{v}}_0^\mathsf{T} \mathbf{x} + \mu^+ = \hat{\mathbf{v}}_0^\mathsf{T} \mathbf{x}_0 + \epsilon \|\hat{\mathbf{v}}_0\|_q + \mu^+, \tag{8.22}$$

where $q$ is the dual variable for $p$, i.e., $p^{-1} + q^{-1} = 1$. Our implementation only considers the case where $p = \infty$ and $q = 1$.

Similarly, the lower bound of $\hat{z}_{k+1}$ can be computed as

$$\hat{\ell}_{k+1} := \min_{\|\mathbf{x} - \mathbf{x}_0\|_p \leq \epsilon} \hat{\mathbf{v}}_0^\mathsf{T} \mathbf{x} + \mu^- = \hat{\mathbf{v}}_0^\mathsf{T} \mathbf{x}_0 - \epsilon \|\hat{\mathbf{v}}_0\|_q + \mu^-, \tag{8.23}$$

where

$$\mu^- = \sum_{i=1}^{k+1} \mathbf{v}_i^\mathsf{T} \mathbf{b}_i - \sum_{i=1}^{k} \sum_{j \in \Gamma_i} \hat{\ell}_{i,j} [\nu_{i,j}]_-. \tag{8.24}$$

We are using the same dual network for both the upper bound and the lower bound.

The above process should be performed iteratively from $k = 0$ to $k = n - 1$. If there are multiple nodes in layer $k + 1$, we need to combine all dual variables together. Denote the variables in the dual network and the value function for the $j$th node in layer $k + 1$ as $\mathbf{v}_i^{k+1,j}$ and $\gamma_i^{k+1,j}$ for $i \leq k + 1$. The derivation of $\mathbf{v}_i^{k+1,j}$ and $\gamma_i^{k+1,j}$ follows the discussion above, except for the boundary conditions. The boundary condition for the dual network is now $\mathbf{v}_{k+1}^{k+1,j} = [0, \ldots, 0, 1, 0, \ldots, 0]$ where the value is 1 for the $j$th entry and 0 otherwise. When we optimize the value for the $j$the node, the objective is $(\mathbf{v}_{k+1}^{k+1,j})^\mathsf{T} \mathbf{z}_{k+1}$. The boundary condition for the bias is kept unchanged $\gamma_{k+1}^{k+1,j} = 0$. Define a vector $\boldsymbol{\gamma}_i^{k+1} := [\gamma_i^{k+1,1}, \ldots, \gamma_i^{k+1,k_{k+1}}] \in \mathbb{R}^{k_{k+1}}$. Define a matrix $\mathbf{V}_i^{k+1} \in \mathbb{R}^{k_i \times k_{k+1}}$ to be the horizontal concatenation of $\mathbf{v}_i^{k+1,j}$ for all $j \in \{1, \ldots, k_{k+1}\}$, *i.e.*,

$$\mathbf{V}_i^{k+1} = [\ \mathbf{v}_i^{k+1,1}\ \ \mathbf{v}_i^{k+1,2}\ \cdots\ \mathbf{v}_i^{k+1,k+1}\ ]. \tag{8.25}$$

The matrices $\mathbf{V}_i^{k+1}$ for all $i \leq k+1$ also have a network structure. According to (8.20), the network structure can be described by

$$\mathbf{V}_i^{k+1} = \mathbf{D}_i \mathbf{W}_{i+1}^\mathsf{T} \mathbf{V}_{i+1}^{k+1}, \forall i \leq k, \tag{8.26}$$

with boundary condition

$$\mathbf{V}_{k+1}^{k+1} = \mathbf{I}. \tag{8.27}$$

Following (8.22) and (8.23), the upper and lower bounds for $\hat{\mathbf{z}}_{k+1}$ can be computed as

$$\hat{\mathbf{u}}_{k+1} = \mathbf{m}_{k+1} + \epsilon \|(\mathbf{V}_1^{k+1})^\mathsf{T} \mathbf{W}_1\|_q - \sum_{i=1}^{k} [(\mathbf{V}_i^{k+1})]_+^\mathsf{T} \mathbf{D}_i^* \hat{\boldsymbol{\ell}}_i, \tag{8.28a}$$

$$\hat{\boldsymbol{\ell}}_{k+1} = \mathbf{m}_{k+1} - \epsilon \|(\mathbf{V}_1^{k+1})^\mathsf{T} \mathbf{W}_1\|_q - \sum_{i=1}^{k} [(\mathbf{V}_i^{k+1})]_-^\mathsf{T} \mathbf{D}_i^* \hat{\boldsymbol{\ell}}_i, \tag{8.28b}$$

where $\mathbf{m}_{k+1} := (\mathbf{V}_1^{k+1})^\mathsf{T} \mathbf{W}_1 \mathbf{x}_0 + \sum_{i=1}^{k+1} (\mathbf{V}_i^{k+1})^\mathsf{T} \mathbf{b}_i$ and $\mathbf{D}_i^*$ is a diagonal matrix whose $j$th diagonal entry is 1 for $j \in \Gamma_i$ and 0 otherwise. The $q$ norm is taken row-wise, *i.e.*, we obtain a column vector

$$\|(\mathbf{V}_1^{k+1})^\mathsf{T} \mathbf{W}_1\|_q := [\|(\mathbf{v}_1^{k+1,1})^\mathsf{T} \mathbf{W}_1\|_q; \ldots; \|(\mathbf{v}_1^{k+1,k+1})^\mathsf{T} \mathbf{W}_1\|_q]. \tag{8.29}$$

**Relationships among dual networks** For every $k$, we indeed have a dual network of $k$ hidden layers. The dual variables in the dual network are related. The correlations are derived below. Note that $\mathbf{V}_i^{k+1} = \mathbf{D}_i \mathbf{W}_{i+1}^\mathsf{T} \mathbf{V}_{i+1}^{k+1} = \mathbf{D}_i \mathbf{W}_{i+1}^\mathsf{T} \mathbf{D}_{i+1} \mathbf{W}_{i+2}^\mathsf{T} \cdots \mathbf{D}_k \mathbf{W}_{k+1}^\mathsf{T}$. And $\mathbf{V}_i^k = \mathbf{D}_i \mathbf{W}_{i+1}^\mathsf{T} \mathbf{V}_{i+1}^k = \mathbf{D}_i \mathbf{W}_{i+1}^\mathsf{T} \mathbf{D}_{i+1} \mathbf{W}_{i+2}^\mathsf{T} \cdots \mathbf{D}_{k-1} \mathbf{W}_k^\mathsf{T}$. Hence,

$$\mathbf{V}_i^{k+1} = \mathbf{V}_i^k \mathbf{D}_k \mathbf{W}_{k+1}^\mathsf{T}. \tag{8.30}$$

The relationships among all $\mathbf{V}_i^k$'s are shown below.

$$\begin{array}{ccccccccc}
 & & & & & & & & \mathbf{V}_n^n \\
 & & & & & & & & \downarrow \mathbf{D}_{n-1}\mathbf{W}_n^\mathsf{T} \\
 & & & & & & \mathbf{V}_{n-1}^{n-1} & \xrightarrow{\cdot \mathbf{D}_{n-1}\mathbf{W}_n^\mathsf{T}} & \mathbf{V}_{n-1}^n \\
 & & & & & & \vdots & & \vdots \\
 & & & & \mathbf{V}_3^3 & \cdots\cdots & \mathbf{V}_3^{n-1} & \xrightarrow{\cdot \mathbf{D}_{n-1}\mathbf{W}_n^\mathsf{T}} & \mathbf{V}_3^n \\
 & & & & \downarrow \mathbf{D}_2\mathbf{W}_3^\mathsf{T} & & \downarrow \mathbf{D}_2\mathbf{W}_3 & & \downarrow \mathbf{D}_2\mathbf{W}_3^\mathsf{T} \\
 & & \mathbf{V}_2^2 & \xrightarrow{\cdot \mathbf{D}_2\mathbf{W}_3^\mathsf{T}} & \mathbf{V}_2^3 & \cdots\cdots & \mathbf{V}_2^{n-1} & \xrightarrow{\cdot \mathbf{D}_{n-1}\mathbf{W}_n^\mathsf{T}} & \mathbf{V}_2^n \\
 & & \downarrow \mathbf{D}_1\mathbf{W}_2^\mathsf{T} & & \downarrow \mathbf{D}_1\mathbf{W}_2^\mathsf{T} & & \downarrow \mathbf{D}_1\mathbf{W}_2^\mathsf{T} & & \downarrow \mathbf{D}_1\mathbf{W}_2^\mathsf{T} \\
\mathbf{V}_1^1 & \xrightarrow{\cdot \mathbf{D}_1\mathbf{W}_2^\mathsf{T}} & \mathbf{V}_1^2 & \xrightarrow{\cdot \mathbf{D}_2\mathbf{W}_3^\mathsf{T}} & \mathbf{V}_1^3 & \cdots\cdots & \mathbf{V}_1^{n-1} & \xrightarrow{\cdot \mathbf{D}_{n-1}\mathbf{W}_n^\mathsf{T}} & \mathbf{V}_1^n.
\end{array} \tag{8.31}$$

There are two approaches to compute $\mathbf{V}_i^k$ for different $i$ and $k$.

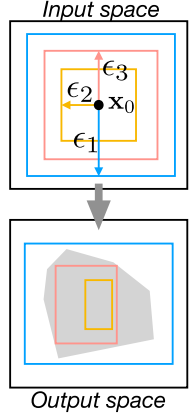

**Figure 8.2:** Illustration of the binary search in FastLin. The upper plot shows three different input sets with radius $\epsilon_1$, $\epsilon_2$, and $\epsilon_3$. The lower plot shows the output constraint $\mathcal{Y}$ (gray area), and the output reachable sets that correspond to the three input sets. The first reachable set $\mathcal{R}(\epsilon_1)$ does not belong to $\mathcal{Y}$. Then a smaller radius $\epsilon_2$ is chosen. The resulting reachable set $\mathcal{R}(\epsilon_2)$ is a subset of $\mathcal{Y}$. Then a larger radius $\epsilon_3 \in (\epsilon_2, \epsilon_1)$ is chosen. This process is repeated.

1. To compute the dual network from scratch for every $k$.

2. To compute the dual network for $k+1$ by reusing the dual network from the last $k$ following (8.30).

The first approach is used in the original implementation in FastLin. The second approach is used in the original implementation in ConvDual. The second approach is more computationally efficient, and is shown in algorithm 35.

**Binary search** The binary search process of the allowable disturbance $\epsilon$ is illustrated in figure 8.2. We keep track of lower and upper bounds of $\epsilon$, denoted $\underline{\epsilon}$ and $\bar{\epsilon}$. The output reachable set for input set with radius $\epsilon$ is denoted $\mathcal{R}(\epsilon)$, which satisfies the bounds computed in (8.28). The variables $\underline{\epsilon}$ and $\bar{\epsilon}$ need to satisfy that

$$\mathcal{R}(\underline{\epsilon}) \subseteq \mathcal{Y}, \mathcal{R}(\bar{\epsilon}) \not\subseteq \mathcal{Y}. \tag{8.32}$$

At every search step, the reachable set $\mathcal{R}(\epsilon)$ for $\epsilon := \frac{\underline{\epsilon}+\bar{\epsilon}}{2}$ is computed. There are two possibilities.

- If $\mathcal{R}(\epsilon) \subseteq \mathcal{Y}$, then $\underline{\epsilon}$ is updated to be $\epsilon$.

- If $\mathcal{R}(\epsilon) \not\subseteq \mathcal{Y}$, then $\bar{\epsilon}$ is updated to be $\epsilon$.

The search process can be terminated if either the maximum iteration is reached or the bounds are close enough to each other, *i.e.*, $\bar{\epsilon} - \underline{\epsilon}$ is small. The binary search is implemented in the main loop in algorithm 36.

```julia
function get_bounds(nnet::Network, input::Vector{Float64}, ϵ::Float64)
    layers  = nnet.layers
    l = Vector{Vector{Float64}}() # Lower bound
    u = Vector{Vector{Float64}}() # Upper bound
    b = Vector{Vector{Float64}}() # bias
    μ = Vector{Vector{Vector{Float64}}}() # Dual variables
    input_ReLU = Vector{Vector{Float64}}()
    v1 = layers[1].weights'
    push!(b, layers[1].bias)
    l1, u1 = input_layer_bounds(layers[1], input, ϵ)
    push!(l, l1)
    push!(u, u1)
    for i in 2:length(layers)
        n_input  = length(layers[i-1].bias)
        n_output = length(layers[i].bias)
        last_input_ReLU = relaxed_ReLU.(last(l), last(u))
        push!(input_ReLU, last_input_ReLU)
        D = Diagonal(last_input_ReLU)
        WD = layers[i].weights*D
        v1 = v1 * WD'
        map!(g -> WD*g,   b, b) # propagate bias
        for V in μ
            map!(m -> WD*m,   V, V)
        end
        push!(b, layers[i].bias)
        push!(μ, new_μ(n_input, n_output, last_input_ReLU, WD))
        ψ = v1' * input + sum(b)
        eps_v1_sum = ϵ * vec(sum(abs, v1, dims = 1))
        neg, pos = residual(input_ReLU, l, μ, n_output)
        push!(l,  ψ - eps_v1_sum - neg )
        push!(u,  ψ + eps_v1_sum - pos )
    end
    return l, u
end
```

**Algorithm 35:** Reachability analysis in FastLin via network relaxation. The variables in the $(k+1)$th dual network is computed by reusing the variables in the $k$th dual network.

```
struct FastLin
    maxIter::Int64
    ϵ0::Float64
    accuracy::Float64
end

function solve(solver::FastLin, problem::Problem)
    ϵ_upper = 2 * max(solver.ϵ0, maximum(problem.input.radius))
    ϵ = fill(maximum(problem.input.radius), solver.maxIter+1)
    ϵ_lower = 0.0
    n_input = dim(problem.input)
    for i = 1:solver.maxIter
        input_bounds = Hyperrectangle(problem.input.center, fill(ϵ[i], n_input))
        output_bounds = forward_network(solver, problem.network, input_bounds)
        if issubset(output_bounds, problem.output)
            ϵ_lower = ϵ[i]
            ϵ[i+1] = (ϵ[i] + ϵ_upper) / 2
            abs(ϵ[i] - ϵ[i+1]) > solver.accuracy || break
        else
            ϵ_upper = ϵ[i]
            ϵ[i+1] = (ϵ[i] + ϵ_lower) / 2
        end
    end
    if ϵ_lower > maximum(problem.input.radius)
        return AdversarialResult(:holds, ϵ_lower)
    else
        return AdversarialResult(:violated, ϵ_lower)
    end
end
```

**Algorithm 36:** FastLin. This main function shows the binary search process to determine the maximum allowable disturbance in the input space.

## 8.4 FastLip

FastLip [66] estimates the local Lipschitz constant, which is equivalent to what Certify computes in (7.32). Certify considers only networks with one hidden layer, while FastLip deals with networks with multiple layers. Same as in FastLin, FastLip takes in a hyperrectangle input set and a polytope output set. For simplicity, we assume the output set is only a halfspace defined by $\mathbf{c}^\mathsf{T}\mathbf{y} \leq d$. FastLip optimizes the following problem,

$$\min_{\mathbf{x}} \quad \epsilon, \tag{8.33a}$$

$$\text{s.t. } \|\mathbf{x}-\mathbf{x}_0\|_q \geq \epsilon, \mathbf{y}=\mathbf{f}(\mathbf{x}), \mathbf{c}^\mathsf{T}\mathbf{y} \geq d. \tag{8.33b}$$

Let $o(\mathbf{x}) = \mathbf{c}^\mathsf{T}\mathbf{y} - d = \mathbf{c}^\mathsf{T}\mathbf{f}(\mathbf{x}) - d$. Assume $\epsilon$ satisfies the above optimization, then $o(\mathbf{x}) \geq 0$ for some $\mathbf{x}$ with $\|\mathbf{x}-\mathbf{x}_0\|_q = \epsilon$. Similar to (7.29), we have the following relationship

$$0 \leq o(\mathbf{x}) \leq o(\mathbf{x}_0) + \epsilon \max_{\|\mathbf{x}-\mathbf{x}_0\|_q \leq \epsilon} \|\nabla o\|_p. \tag{8.34}$$

Hence, the solution of the problem (8.33) is bounded by

$$\min_{\|\mathbf{x}-\mathbf{x}_0\|_q \geq \epsilon, o(\mathbf{x}) \geq 0} \epsilon \geq -\frac{o(\mathbf{x}_0)}{\max_{\|\mathbf{x}-\mathbf{x}_0\|_q \leq \epsilon} \|\nabla o\|_p}. \tag{8.35}$$

The max over $\|\nabla o\|_p$ is hard to compute without knowing the optimal solution $\epsilon$ on the left hand side. In practice, the max over $\|\nabla o\|_p$ is taken over $\mathcal{X} = \{\mathbf{x} = \|\mathbf{x}-\mathbf{x}_0\|_q \leq \epsilon^{FastLin}\}$ where $\epsilon^{FastLin}$ is the solution obtained from FastLin. The adversarial bound is chosen as the minimum of $\epsilon^{FastLin}$ and the right hand side of (8.35). For simplicity, we only consider the norm with $p=1$ and $q=\infty$.

As shown in (8.35), to compute the adversarial bound $\epsilon$, we need to compute the gradient $\nabla o$. Recall that Certify uses semidefinite relaxation to compute $\|\nabla o\|_1$. FastLip uses a different approach to estimate element-wise derivative $\nabla_{x_j} o = \frac{\partial o}{\partial x_j}$ for $j \in \{1,\ldots,k_0\}$, then

$$\max_{\mathbf{x} \in \mathcal{X}} \|\nabla_{\mathbf{x}} o\|_1 \leq \sum_j \max_{\mathbf{x} \in \mathcal{X}} |\nabla_{x_j} o|. \tag{8.36}$$

The computation of the gradient follows from the method discussed in section 4.3, which uses interval arithmetic and considers the chain rule. Given the bounds computed in FastLin, the activation pattern can be determined. Hence, the lower and upper bounds of the gradients of the activation functions can be determined. The bounds on gradients $\underline{\mathbf{G}}_i$ and $\overline{\mathbf{G}}_i$ can be computed following (4.12) and (4.13).[7] Given the bounds on the gradients, the gradient of $o$ satisfies that

$$\underbrace{[\mathbf{c}]_+^\mathsf{T} \underline{\mathbf{G}}_n + [\mathbf{c}]_-^\mathsf{T} \overline{\mathbf{G}}_n}_{\mathbf{a}} \leq \nabla o \leq \underbrace{[\mathbf{c}]_+^\mathsf{T} \overline{\mathbf{G}}_n + [\mathbf{c}]_-^\mathsf{T} \underline{\mathbf{G}}_n}_{\mathbf{b}}. \tag{8.37}$$

---

[7] The original paper uses a different set of equations. The two derivations are indeed equivalent.

The maximum 1-norm gradient is

$$\max_{\mathbf{x} \in \mathcal{X}} \|\nabla_{\mathbf{x}} o\|_1 \leq \sum_j \max_{\mathbf{x} \in \mathcal{X}} |\nabla_{x_j} o|_1 = \sum_j \max\{|a_j|, |b_j|\}. \tag{8.38}$$

Then

$$\epsilon^{FastLip} := \min\left\{\frac{-o(\mathbf{x}_0)}{\sum_j \max\{|a_j|, |b_j|\}}, \epsilon^{FastLin}\right\}. \tag{8.39}$$

Our implementation is shown in algorithm 37.

```
struct FastLip
    maxIter::Int64
    ϵ0::Float64
    accuracy::Float64
end

function solve(solver::FastLip, problem::Problem)
    c, d = tosimplehrep(convert(HPolytope, problem.output))
    y = compute_output(problem.network, problem.input.center)
    o = (c * y - d)[1]
    if o > 0
        return AdversarialResult(:violated, -o)
    end
    result = solve(FastLin(solver), problem)
    result.status == :violated && return result
    ϵ_fastLin = result.max_disturbance
    LG, UG = get_gradient(problem.network, problem.input)
    a, b = interval_map(c, LG, UG)
    v = max.(abs.(a), abs.(b))
    ϵ = min(-o/sum(v), ϵ_fastLin)
    if ϵ > maximum(problem.input.radius)
        return AdversarialResult(:holds, ϵ)
    else
        return AdversarialResult(:violated, ϵ)
    end
end
```

**Algorithm 37:** FastLip. FastLip depends on FastLin and further estimates the allowable input range by computing a local Lipschitz constant of the network.

## 8.5 DLV

DLV [30] searches not only the input space but also the hidden layers for a possible counter example. It uses a layer-by-layer approach. At layer $i$, it tries to find an assignment of the hidden nodes $\mathbf{z}_i$ that satisfies the following two constraints:

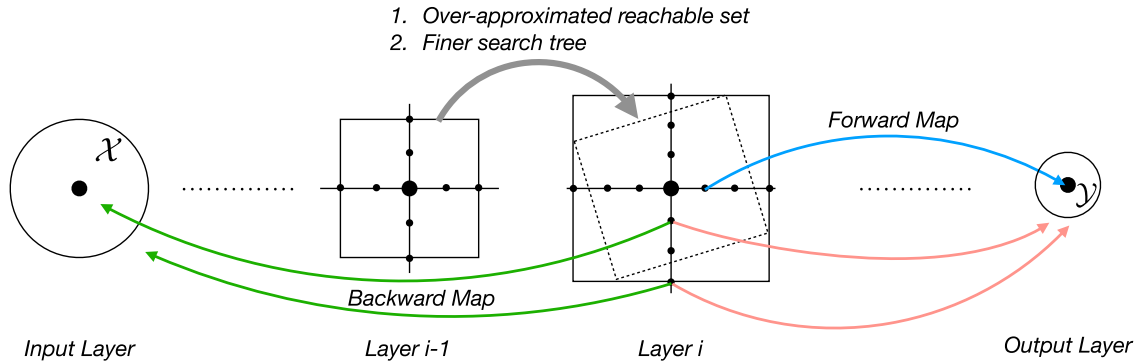

**Figure 8.3:** Illustration of the DLV approach. DLV uses layer-by-layer analysis. At layer $i$, it computes an over-approximated reachable set and a finer search tree based on the reachable set and the search tree in layer $i-1$. For each point on the search tree, DLV first checks whether it violates the output constraint $\mathcal{Y}$ using the forward map. For a point that violates the output constraint, DLV then checks whether it is reachable given the input constraint $\mathcal{X}$ using the backward map. If such a point is reachable, then the corresponding input value is taken as a counter example.

1. Under the forward mapping $\mathbf{f}_{i+1 \to n} := \mathbf{f}_n \circ \cdots \circ \mathbf{f}_{i+1}$, the output $\mathbf{f}_{i+1 \to n}(\mathbf{z}_i)$ does not belong to the desired constraint $\mathcal{Y}$.

2. Under the backward mapping $\mathbf{f}_{1 \to i} := \mathbf{f}_i \circ \cdots \circ \mathbf{f}_1$, there is a valid input $\mathbf{x} \in \mathcal{X}$ such that $\mathbf{f}_{1 \to i}(\mathbf{x}) = \mathbf{z}_i$.

The conditions can be written concisely as

$$(\mathbf{f}_{1 \to i})^{-1}(\mathbf{z}_i) \bigcap \mathcal{X} \neq \emptyset, \tag{8.40a}$$

$$\mathbf{f}_{i+1 \to n}(\mathbf{z}_i) \notin \mathcal{Y}. \tag{8.40b}$$

**Overall procedure** To search for $\mathbf{z}_i$ that satisfies (8.40), DLV does the following steps layer by layer.

1. Find a reachable set $\mathcal{R}_i$ at layer $i$ given the reachable set $\mathcal{R}_{i-1}$ at layer $i-1$. The reachable set at the input layer is $\mathcal{R}_0 := \mathcal{X}$.

2. Build a search tree inside $\mathcal{R}_i$, denoted $\mathcal{T}_i$. The search tree $\mathcal{T}_i$ should be a refinement of the search tree $\mathcal{T}_{i-1}$ in the previous layer.[8]

3. Search for a counter example in the search tree $\mathcal{T}_i$ that satisfies (8.40). If such an example is found, the property is violated. If not, we continue to the next layer.

---

[8]The branches in the search tree are called ladders in the original paper [30].

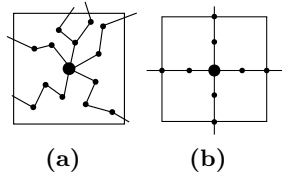

**Figure 8.4:** Illustration of a search tree in a 2D hidden space. (a) Complex search tree that explores all directions. (b) Simplified search tree for element-wise exploration.

The procedure is illustrated in figure 8.3. The key insight is that we can refine the search tree layer by layer. When it is closer to the output layer, it is easier to sample for counter examples. The original DLV implementation[9] uses SMT solvers to construct a reachable set in step 1 and build a search tree in step 2. For simplicity, our implementation in algorithm 38 does not involve SMT solvers. For step 1, we may compute the reachable set using any reachability method discussed in chapter 5. In algorithm 38, we use interval arithmetic by directly calling MaxSens using the function `get_bounds` in algorithm 5. The considerations in building the search tree in step 2 and the satisfiability check in step 3 are discussed below.

**Search tree**   To build a search tree, we may explore all possible directions as shown in figure 8.4a. However, the complexity grows exponentially with $k_i$, the number of nodes in layer $i$. The authors of DLV suggest that we decompose the high-dimensional search space into several perpendicular low-dimensional search spaces, and only build search trees in those low-dimensional search spaces. Here, we simplify the search to element-wise exploration, *i.e.*, we sample every hidden node separately as shown in figure 8.4b. The construction of the search tree for a new layer needs to ensure that the tree is finer in the new layer. A set of conditions that precisely define "finer" are discussed in the DLV paper [30]. The problem of finding a search tree is then solved by calling an SMT solver.

For simplicity, our implementation takes equidistant samples for every hidden node. The sampling interval for node $z_{i,j}$ is denoted $\epsilon_{i,j}$. The sample intervals are refined layer by layer. For the input node, the sampling intervals $\epsilon_{0,j}$ for all $j \in \{1, \ldots, k_0\}$ are set to a solver-specified value. The sample interval at layer $i-1$ indeed defines a hyperrectangle spanned by the following vectors

$$\mathbf{v}_{i-1,1} \coloneqq [\epsilon_{i-1,1}, 0, \ldots, 0], \mathbf{v}_{i-1,2} \coloneqq [0, \epsilon_{i-1,2}, 0, \ldots, 0], \ldots, \\ \mathbf{v}_{i-1,k_{i-1}} \coloneqq [0, \ldots, 0, \epsilon_{i-1,k_{i-1}}]. \tag{8.41}$$

No point, with the exception of the vertices of the hyperrectangle, are evaluated at layer $i-1$. In layer $i$, we need to ensure that we evaluate points inside the hyperrectangle, as

---

[9]https://github.com/VeriDeep/DLV

```julia
struct DLV
    optimizer
    ϵ::Float64
end

function solve(solver::DLV, problem::Problem)
    η = get_bounds(problem)
    δ = Vector{Vector{Float64}}(undef,length(η))
    δ[1] = fill(solver.ϵ, dim(η[1]))
    if issubset(last(η), problem.output)
        return CounterExampleResult(:holds)
    end
    output = compute_output(problem.network, problem.input.center)
    for (i, layer) in enumerate(problem.network.layers)
        δ[i+1] = get_manipulation(layer, δ[i], η[i+1])
        if i == length(problem.network.layers)
            mapping = x -> (x ∈ problem.output)
        else
            forward_nnet = Network(problem.network.layers[i+1:end])
            mapping = x -> (compute_output(forward_nnet, x) ∈ problem.output)
        end
        var, y = bounded_variation(η[i+1], mapping, δ[i+1])
        if var
            backward_nnet = Network(problem.network.layers[1:i])
            status, x = backward_map(solver, y, backward_nnet, η[1:i+1])
            if status
                return CounterExampleResult(:violated, x)
            end
        end
    end
    return ReachabilityResult(:violated, [last(η)])
end
```

**Algorithm 38:** Main code for DLV. First, the reachable set $\eta$ for each layer is computed by calling `get_bounds`. Then the sampling intervals for the input layer are initialized. The solver first checks for counter examples in the input layer by calling `bounded_variation`. If no counter example is found, the solver then proceeds to the hidden layers. For every hidden layer, the desired sampling intervals are computed by calling `get_manipulation`. Then the solver checks for hidden values that do not map to $\mathcal{Y}$ by calling `bounded_variation`. For those hidden values, the solver then maps back to the input layer by calling `backward_map`. If a corresponding input is found, it returns the input as a counter example for the problem. Otherwise, it goes to the next layer. The details of the functions `get_manipulation` and `bounded_variation` are not included.

shown in figure 8.3. The vertices of the hyperrectangle are mapped to $\mathbf{W}_i \mathbf{v}_{i-1,j'}$ at layer $i$ for all $j' \in \{1, \ldots, k_{i-1}\}$. For the $j$th node in layer $i$, the projected sampling interval given the hyperrectangle at layer $i-1$ is $\max_{j'} \sigma_{i,j}(|\mathbf{w}_{i,j}\mathbf{v}_{i-1,j'}|) = \max_{j'} \sigma_{i,j}(|w_{i,j,j'}|\epsilon_{i-1,j'})$.[10] We then set the sample interval to be a value smaller than the projected sampling interval:

$$\epsilon_{i,j} := \gamma \max_{j'} \sigma_{i,j}(|w_{i,j,j'}|\epsilon_{i-1,j'}), \tag{8.42}$$

where $\gamma \in (0, 1)$ is set by the solver. This function is implemented in `get_manipulation`.

Given the sampling interval $\boldsymbol{\epsilon}_i$, the search tree $\mathcal{T}_i$ consists of $2k_i$ chain branches. All branches are centered at $\mathbf{f}_{1 \to i}(\mathbf{x}_0)$. Then chains move along $\mathbf{v}_{i,j}$ in either the positive direction or the negative direction,

$$\mathbf{f}_{1 \to i}(\mathbf{x}_0) \to \mathbf{f}_{1 \to i}(\mathbf{x}_0) + 1 \cdot \mathbf{v}_{i,j} \to \mathbf{f}_{1 \to i}(\mathbf{x}_0) + 2 \cdot \mathbf{v}_{i,j} \to \cdots \tag{8.43a}$$

$$\mathbf{f}_{1 \to i}(\mathbf{x}_0) \to \mathbf{f}_{1 \to i}(\mathbf{x}_0) - 1 \cdot \mathbf{v}_{i,j} \to \mathbf{f}_{1 \to i}(\mathbf{x}_0) - 2 \cdot \mathbf{v}_{i,j} \to \cdots \tag{8.43b}$$

Since $j \in \{1, \ldots, k_i\}$, we then have $2k_i$ chain branches. For example, in figure 8.4b, the search tree in two-dimensional space contains four branches. We don't need to explicitly construct the search tree. The search tree will be implicitly traversed when we check for satisfiability in step 3.

The original implementation of DLV is always complete and is sound when the search tree obtained by SMT solvers is minimal [30]. The tree nodes on a minimal search tree effectively cover all branches of a network. However, the search tree in our implementation is built heuristically and may not be minimal. To still make it sound, we only return holds when the output reachable set belongs to the output constraint. Otherwise, even if a counter example is not found, we return a violated reachability result. Hence, our implementation is sound but not complete.

**Satisfiability check** For every sampled value $\mathbf{z}_i$ in the search tree, we first check if the corresponding output belongs to $\mathcal{Y}$. The function is implemented in `bounded_variation`, which outputs the points on the search tree that violate the output constraint. The original paper [30] indeed computes the maximum bounded variation of the search tree, *i.e.*, the maximum number of links that connect one satisfying node and one unsatisfying node along any tree path. As only one counter example is needed to falsify our problem, our implementation only checks for the existence of such a link.

If such a $\mathbf{z}_i$ is found, we solve the following optimization problem to determine whether

---

[10] The function $\sigma_{i,j}$ is the activation function for node $j$ at layer $i$. $\mathbf{w}_{i,j}$ is the $j$th row in the weight matrix $\mathbf{W}_i$. $w_{i,j,k}$ is the $k$th entry in $\mathbf{w}_{i,j}$.

$\mathbf{z}_i$ is reachable from $\mathcal{X}$:[11]

$$\min_{\mathbf{x}} \|\mathbf{x} - \mathbf{x}_0\|_\infty, \tag{8.44a}$$

$$\text{s.t. } \mathbf{x} \in \mathcal{X}, \mathbf{f}_{1 \to i}(\mathbf{x}) = \mathbf{z}_i. \tag{8.44b}$$

The optimization problem is solved using a MILP encoding in `backward_map`. The method is similar to NSVerify. If we find such an $\mathbf{x}$, then it is a counter example.

# 9 Search and Optimization

This chapter discusses methods that combine search with optimization approaches. These methods leverage the piecewise linearity in the activation functions to develop efficient algorithms. Sherlock [18] uses local and global search to compute bounds on the output, solving different optimization problems during the local search and the global search. BaB [12] uses branch and bound to compute bounds on the output, where the branch step corresponds to search and the bound step corresponds to optimization. Planet [22] formulates the verification problem as a satisfiability problem, searching for an activation pattern such that a feasible input is mapped to an infeasible output. Reluplex [32] uses a simplex algorithm that searches for a feasible activation pattern that leads to an infeasible output.

## 9.1 Sherlock

Sherlock [18] estimates the output range for a network with a single output, *i.e.*, $k_n = 1$. If $k_n > 1$, the method works for each individual variable. Given the input domain $\mathcal{X}$, we compute the tightest output bounds

$$\ell_n = \min_{\mathbf{x} \in \mathcal{X}} f(\mathbf{x}), \quad u_n = \max_{\mathbf{x} \in \mathcal{X}} f(\mathbf{x}). \tag{9.1}$$

Sherlock solves (9.1) by combining local search and global search. In local search, it solves a linear program to find the optimal value in a given line segment of the function $\mathbf{f}$. In global search, it solves a feasibility program to check whether the current local optimal bound can be improved. To find the global optimal bound, the solver iteratively does local search and global search. This approach is illustrated in figure 9.1. The main process is implemented in algorithm 39.

**Local search** Starting from a sampled point $\mathbf{x}^*$ in the input domain. Local search tries to find a better solution that has the same activation pattern as $\mathbf{x}^*$, *i.e.*, in the same line segment of $\mathbf{f}$. Denote the activation pattern with respect to $\mathbf{x}^*$ as $\delta_{i,j}^*$ for all layers $i$ and

---
[11]The point $\mathbf{z}_i$ may lie in the over-approximated area that is not reachable from the input set $\mathcal{X}$.

```
struct Sherlock
    optimizer
    ϵ::Float64
end

function solve(solver::Sherlock, problem::Problem)
    (x_u, u) = output_bound(solver, problem, :max)
    (x_l, l) = output_bound(solver, problem, :min)
    bound = Hyperrectangle(low = [l], high = [u])
    reach = Hyperrectangle(low = [l - solver.ϵ], high = [u + solver.ϵ])
    return interpret_result(reach, bound, problem.output, x_l, x_u)
end

function output_bound(solver::Sherlock, problem::Problem, type::Symbol)
    opt = solver.optimizer
    x = an_element(problem.input)
    while true
        (x, bound) = local_search(problem, x, opt, type)
        ϵ = bound + ifelse(type == :max, solver.ϵ, -solver.ϵ)
        (x_new, bound_new, feasible) = global_search(problem, ϵ, opt, type)
        feasible || return (x, bound)
        (x, bound) = (x_new, bound_new)
    end
end
```

**Algorithm 39:** The main function in Sherlock. Sherlock computes both the upper and the lower bound of the output node through the function `output_bound` and then interprets the result with respect to the given output constraint. The function `output_bound` performs iterative local and global search in the while loop. The global search step needs to improve the bound by at least $\epsilon$. The while loop is broken if the global search fails. The value $\epsilon$ is specified by the solver, which affects the tightness of the bound.

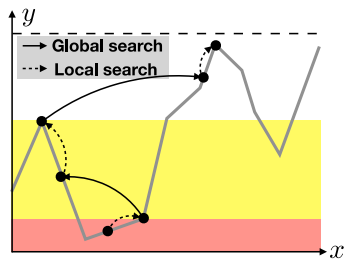

**Figure 9.1:** Illustration of the Sherlock approach. Sherlock combines local and global search. For a network with ReLU activations, the function represented by the network is piecewise linear. Local search improves the bound in a linear segment. Global search finds a point that improves the local bound by $\epsilon$, which lies in another linear segment.

nodes $j$. The network is encoded using (6.3). The local optimization problem for the upper bound is formulated as

$$\max_{\mathbf{z}_0,\ldots,\mathbf{z}_n,\hat{\mathbf{z}}_1,\ldots,\hat{\mathbf{z}}_n} \nabla \mathbf{f}(\mathbf{x}^*)^\mathsf{T} \mathbf{z}_0, \tag{9.2a}$$

$$\text{s.t. } \mathbf{z}_0 \in \mathcal{X}, \tag{9.2b}$$

$$z_{i,j} = \hat{z}_{i,j} \geq 0, \text{if } \delta^*_{i,j} = 1, \forall i \in \{1,\ldots,n\}, j \in \{1,\ldots,k_i\}, \tag{9.2c}$$

$$z_{i,j} = 0, \hat{z}_{i,j} \leq 0, \text{if } \delta^*_{i,j} = 0, \forall i \in \{1,\ldots,n\}, j \in \{1,\ldots,k_i\}, \tag{9.2d}$$

$$\hat{\mathbf{z}}_i = \mathbf{W}_i \mathbf{z}_{i-1} + \mathbf{b}_i, \forall i \in \{1,\ldots,n\}, \tag{9.2e}$$

where $\nabla \mathbf{f}(\mathbf{x}^*)$ is the gradient at $\mathbf{x}^*$ computed by (4.8). The optimal solution is denoted as $\mathbf{x}^l$. The corresponding bound is $u_n^l := f(\mathbf{x}^l)$. Similarly we can compute the lower bound by changing the max to min. The local search is implemented in algorithm 40.

```
function local_search(problem::Problem, x, optimizer, type::Symbol)
    nnet = problem.network
    act_pattern = get_activation(nnet, x)
    gradient = get_gradient(nnet, x)
    model = Model(with_optimizer(optimizer))
    neurons = init_neurons(model, nnet)
    add_set_constraint!(model, problem.input, first(neurons))
    encode_network!(model, nnet, neurons, act_pattern, StandardLP())
    o = gradient * neurons[1]
    index = ifelse(type == :max, 1, -1)
    @objective(model, Max, index * o[1])
    optimize!(model)
    x_new = value(neurons[1])
    bound_new = compute_output(nnet, x_new)
    return (x_new, bound_new[1])
end
```

**Algorithm 40:** Local search in Sherlock, which solves equation (9.2). The network is encoded as a set of linear constraints according to the given activation pattern. Those constraints confine the function $f$ to the line segment that $\mathbf{x}^*$ lies on. The slope of the line segment equals the gradient of $f$ at $\mathbf{x}^*$. Hence, the solution of the local search is a local optimum of the function $f$.

**Global search** Given the bound from local search, a global search aims to improve the bound by $\epsilon$, a solver-specified value. The global search problem is a feasibility problem. For example, for the upper bound, it tries to find a solution that satisfies

$$\mathbf{x} \in \mathcal{X}, f(\mathbf{x}) \geq u_n^l + \epsilon. \tag{9.3}$$

This problem is solved by calling NSVerify. If a solution $\mathbf{x}^g$ is found, then the upper bound is updated to $u_n^g = f(\mathbf{x}^g)$. The global search is implemented in algorithm 41.

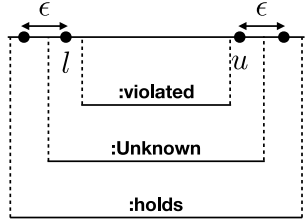

**Figure 9.2:** Interpreting results. The upper line illustrates the computed bound $[l, u]$ and the potential reachable set $[l - \epsilon, u + \epsilon]$. The bottom three lines illustrate three different output constraints $\mathcal{Y}$. If $\mathcal{Y}$ does not cover either $\ell$ or $u$, the solver returns violated. If $\mathcal{Y}$ covers the reachable set, the solver returns holds. Otherwise, the solver returns unknown.

Since global search only solves a feasibility problem, the solution is not guaranteed to be a local optimum. Once a new global bound is found, local search will be called again to obtain the local optimum with respect to the new result. The reference point in the local search is set to $\mathbf{x}^* := \mathbf{x}^g$. After the local search, the global search will be called again. If there is no solution for the global search, we conclude that the upper bound is the bound obtained by the last local search, *i.e.*, $u_n := u_n^l$. Similarly, we can compute the global lower bound.

```
function global_search(problem, bound::Float64, optimizer, type::Symbol)
    index = ifelse(type == :max, 1.0, -1.0)
    h = HalfSpace([index], index * bound)
    output_set = HPolytope([h])
    problem_new = Problem(problem.network, problem.input, output_set)
    solver  = NSVerify(optimizer = optimizer)
    result  = solve(solver, problem_new)
    if result.status == :violated
        x = result.counter_example
        bound = compute_output(problem.network, x)
        return (x, bound[1], true)
    else
        return ([], 0.0, false)
    end
end
```

**Algorithm 41:** Global search in Sherlock. It calls NSVerify to solve equation (9.3) to determine if the bound can be improved by $\epsilon$ or not. If there is such a point that improves the bound by at least $\epsilon$, the point and the new bound are returned. If there does not exist such a point, the problem is determined infeasible by returning a false flag.

**Result interpretation** The original purpose of Sherlock is only to obtain tight bounds. To fit into our problem formulation (2.4), the obtained bounds are compared against the output constraint $\mathcal{Y}$. The bounding set $\mathcal{B} := [\ell_n, u_n]$ is tight with at most $\epsilon$ uncertainty. The reachable set is defined as $\tilde{\mathcal{R}} := [\ell_n - \epsilon, u_n + \epsilon]$. The solution of the problem can be determined by

$$\tilde{\mathcal{R}} \in \mathcal{Y} \Rightarrow \text{holds},\tag{9.4a}$$

$$\mathcal{B} \notin \mathcal{Y} \Rightarrow \text{violated},\tag{9.4b}$$

$$\text{otherwise} \Rightarrow \text{unknown}.\tag{9.4c}$$

The three conditions are illustrated in figure 9.2 and implemented in algorithm 42. When the reachable set belongs to the output constraint, the property holds. When either bound $u_n$ or $\ell_n$ exceeds the output constraint, the property is violated. In this case, we can output the counter example, which is the point that achieves either $u_n$ or $\ell_n$ that violates the constraint. Otherwise, the problem is undetermined. The smaller the $\epsilon$, the more accurate the result, as the problem is less likely to be undetermined. It is worth noting that the key advantage of Sherlock is to output tight bounds. It calls optimization solvers multiple times during the search for bounds. If the purpose is only to verify a given output constraint, it would be more computationally efficient to solve a feasibility problem using NSVerify, which corresponds to one global search step in Sherlock.

```
function interpret_result(reach, bound, output, x_l, x_u)
    if high(reach) > high(output) && low(reach) < low(output)
        return ReachabilityResult(:holds, reach)
    end
    high(bound) > high(output)  && return CounterExampleResult(:violated, x_u)
    low(bound)  < low(output)   && return CounterExampleResult(:violated, x_l)
    return RechabilityResult(:unknown, reach)
end
```

**Algorithm 42:** Interpretation of results. The code implements equation (9.4) and outputs corresponding counter examples when there is a violation. The input `reach` corresponds to $[\ell_n - \epsilon, u_n + \epsilon]$. The input `bound` corresponds to $[\ell_n, u_n]$. The input `output` is the output constraint $\mathcal{Y}$. The last two inputs are the points that achieve $\ell_n$ and $u_n$ respectively.

## 9.2 BaB

BaB [12] uses branch and bound to estimate the output bounds of a network.[1] The main loop is shown in algorithm 43. For simplicity, we assume that the network only has one output node. Similar to Sherlock, the solver tries to compute the lower and upper bounds of the output in (9.1). The bounds are computed using the function `output_bound`. The results are interpreted by `interpret_result` in algorithm 42. BaB is different from Sherlock mainly in the method to estimate bounds. In the following discussion, we discuss the approach to estimate the upper bound $u_n$. Estimation of $\ell_n$ follows easily.

---

[1] It is observed in the paper that several methods, such as Reluplex and Planet, can also be viewed as branch and bound.

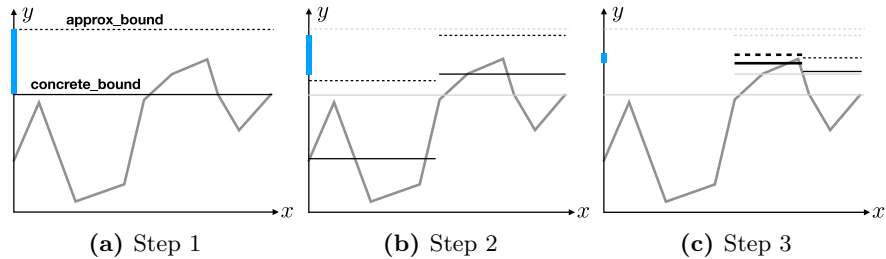

**Figure 9.3:** Illustration of branch and bound to estimate the upper bound of $y$. In step 1, an approximated upper bound (dashed line) and a concrete lower bound (solid line) are computed for the whole input interval. As there is a large gap between the two bounds (blue range), the input interval is split in two. In step 2, approximated upper bounds and concrete lower bounds are computed for the two intervals respectively. As the upper bound of the left interval is smaller than the lower bound of the right interval, the left interval is pruned out. The new global bounds are updated to be the bounds of the right interval. In step 3, the right interval is further split into two intervals. The previous process is repeated. The new bounds are close to each other. Hence, the global upper bound of $y$ is found.

We maintain a global upper bound $\bar{u}_n$ of $u_n$, and a global lower bound $\underline{u}_n$ of $u_n$. The lower bound $\underline{u}_n$ is computed by sampling for a concrete input, hence it is also called the concrete bound. The upper bound $\bar{u}_n$ is computed with respect to a relaxation of the problem, hence it is called the approximated bound. The concrete bound always provides an under-estimation, while the upper bound always provides an over-estimation. For the lower bound $\ell_n$, its lower bound $\underline{\ell}_n$ is the approximated bound, while its upper bound $\bar{\ell}_n$ is the concrete bound. These two possibilities are encoded by the `:max` and `:min` symbols in algorithm 43.

If $\bar{u}_n - \underline{u}_n < \epsilon$ for some small $\epsilon$ specified by the solver, the over-estimation is small, and we conclude $u_n$ is found. To minimize the over-approximation, the bounds $\bar{u}_n$ and $\underline{u}_n$ are improved by iterative interval refinement (a search process) similar to the procedure discussed in ReluVal in section 8.1.

**Search strategy** During the search process, a list of input subsets $\{\mathcal{X}_i\}_i$ are recorded. In addition, the approximated bound of that subset $\mathcal{X}_i$ is recorded, which is used to prioritize the list. For the estimation of $u_n$, we record $\bar{u}_n(\mathcal{X}_i)$. The list satisfies that $\bar{u}_n(\mathcal{X}_1) \geq \bar{u}_n(\mathcal{X}_2) \geq \cdots$. For the estimation of $\ell_n$, we record $\underline{\ell}_n(\mathcal{X}_i)$. The list satisfies that $\underline{\ell}_n(\mathcal{X}_1) \leq \underline{\ell}_n(\mathcal{X}_2) \leq \cdots$.

At each iteration, the first domain in the priority queue, *i.e.*, the domain with the largest upper bound $\bar{u}_n(\mathcal{X}_1)$ or the domain with the smallest lower bound $\underline{\ell}_n(\mathcal{X}_1)$, is picked out and split into two domains. In ReluVal, we split the index that corresponds to the greatest smear value. In BaB, we split the index that has the longest interval. The resulting two domains are denoted $\mathcal{X}_1^j$ for $j \in \{1, 2\}$.

For the two domains, we compute their approximated and concrete bounds $\bar{u}_n(\mathcal{X}_1^j)$ and $\underline{u}_n(\mathcal{X}_1^j)$ for $j \in \{1, 2\}$. The global lower bound $\underline{u}_n(\mathcal{X})$ is updated to $\max_j \underline{u}_n(\mathcal{X}_1^j)$ if it is greater than the previous global lower bound. If the upper bound $\bar{u}_n(\mathcal{X}_1^j)$ is greater than the global lower bound, the corresponding sub domain $\mathcal{X}_1^j$ needs further splits. Then $\mathcal{X}_1^j$ is pushed back to the priority list. The current global upper bound $\bar{u}_n(\mathcal{X})$ is $\bar{u}_n(\mathcal{X}_1)$ in the new list.

The search terminates if the global concrete and approximated bounds are close to each other, *i.e.*, $\bar{u}_n(\mathcal{X}) - \underline{u}_n(\mathcal{X}) < \epsilon$. The search process is illustrated in figure 9.3.

**Concrete bounds** The concrete bounds $\underline{u}_n(\mathcal{X}_i)$ and $\bar{\ell}_n(\mathcal{X}_i)$ are computed by sampling in the domain $\mathcal{X}_i$.[2] Heuristically, we only check three points in a domain, the upper and lower bounds of the domain, and the center point. Hence, an upper bound is always accompanied with a concrete sample input that achieves that bound. Computation of concrete bounds is implemented in the function `concrete_bound` in algorithm 44.

**Approximated bounds** The approximated bounds $\bar{u}_n(\mathcal{X}_i)$ and $\underline{\ell}_n(\mathcal{X}_i)$ are computed by minimizing the output given the triangle relaxation of the network. For example, $\bar{u}_n(\mathcal{X}_i)$ is computed by

$$\max_{\mathbf{z}_0,\ldots,\mathbf{z}_n,\hat{\mathbf{z}}_1,\ldots,\hat{\mathbf{z}}_n} z_n, \tag{9.5a}$$

$$\text{s.t.} \quad z_{i,j} = \hat{z}_{i,j}, \forall i \in \{1,\ldots,n-1\}, j \in \Gamma_i^+, \tag{9.5b}$$

$$z_{i,j} = 0, \forall i \in \{1,\ldots,n-1\}, j \in \Gamma_i^-, \tag{9.5c}$$

$$z_{i,j} \geq \hat{z}_{i,j}, z_{i,j} \geq 0, z_{i,j} \leq \frac{\hat{u}_{i,j}(\hat{z}_{i,j} - \hat{\ell}_{i,j})}{\hat{u}_{i,j} - \hat{\ell}_{i,j}},$$

$$\forall i \in \{1,\ldots,n-1\}, j \in \Gamma_i, \tag{9.5d}$$

$$\hat{\mathbf{z}}_i = \mathbf{W}_i \mathbf{z}_{i-1} + \mathbf{b}_i, \forall i \in \{1,\ldots,n-1\}, \tag{9.5e}$$

$$\mathbf{z}_0 \in \mathcal{X}_i. \tag{9.5f}$$

The approximated bound $\underline{\ell}_n(\mathcal{X}_i)$ can be computed by changing the maximization to minimization in (9.5). ConvDual solves a similar optimization problem in (7.21), though it transforms it to the dual problem and then obtains a heuristic solution. Due to relaxation, the above optimization overestimates the true bounds, *i.e.*, $\bar{u}_n(\mathcal{X}_i) \geq u_n(\mathcal{X}_i)$ and $\underline{\ell}_n(\mathcal{X}_i) \leq \ell_n(\mathcal{X}_i)$. The finer $\mathcal{X}_i$, the closer the approximated bound is to the true value.

---

[2]ReluVal uses a similar approach to search for violations in `check_inclusion`.

```julia
struct BaB
    optimizer
    ϵ::Float64
end

function solve(solver::BaB, problem::Problem)
    (u_approx, u, x_u) = output_bound(solver, problem, :max)
    (l_approx, l, x_l) = output_bound(solver, problem, :min)
    bound = Hyperrectangle(low = [l], high = [u])
    reach = Hyperrectangle(low = [l_approx], high = [u_approx])
    return interpret_result(reach, bound, problem.output, x_l, x_u)
end

function output_bound(solver::BaB, problem::Problem, type::Symbol)
    nnet = problem.network
    opt = solver.optimizer
    global_concrete, x_star = concrete_bound(nnet, problem.input, type)
    global_approx = approx_bound(nnet, problem.input, opt, type)
    doms = Tuple{Float64, Hyperrectangle}[(global_approx, problem.input)]
    index = ifelse(type == :max, 1, -1)
    while index * (global_approx - global_concrete) > solver.ϵ
        dom, doms = pick_out(doms) # pick_out implements the search strategy
        subdoms = split_dom(dom[2]) # split implements the branching rule
        for i in 1:length(subdoms)
            dom_concrete, x = concrete_bound(nnet, subdoms[i], type)
            dom_approx = approx_bound(nnet, subdoms[i], opt, type)
            if index * (dom_concrete - global_concrete) > 0
                (global_concrete, x_star) = (dom_concrete, x)
            end
            if index * (dom_approx - global_concrete) > 0
                add_domain!(doms, (dom_approx, subdoms[i]), type)
            end
        end
        length(doms) == 0 && return (global_approx, global_concrete, x_star)
        global_approx = doms[1][1]
    end
end
```

**Algorithm 43:** The main function in BaB. BaB computes both the upper and the lower bound of the output node through the function `output_bound` and then interprets the result with respect to the given output constraint. The function `output_bound` performs branch and bound in the while loop. The solver keeps track of a concrete under-estimation and an approximated over-estimation in the search process. The search terminates once the two bounds are close to each other. The value $\epsilon$ is specified by the solver, which affects the tightness of the bound.

Hence, the approximated bound is improved during iterative refinement. Computation of the approximated bounds is implemented in the function `approx_bound` in algorithm 44.

```julia
function concrete_bound(nnet::Network, subdom::Hyperrectangle, type::Symbol)
    points = [subdom.center, low(subdom), high(subdom)]
    values = Vector{Float64}(undef, 0)
    for p in points
        push!(values, sum(compute_output(nnet, p)))
    end
    value, index = ifelse(type == :min, findmin(values), findmax(values))
    return (value, points[index])
end

function approx_bound(nnet::Network, dom::Hyperrectangle, optimizer, type)
    bounds = get_bounds(nnet, dom)
    model = Model(with_optimizer(optimizer))
    neurons = init_neurons(model, nnet)
    add_set_constraint!(model, dom, first(neurons))
    encode_network!(model, nnet, neurons, bounds, TriangularRelaxedLP())
    index = ifelse(type == :max, 1, -1)
    o = sum(last(neurons))
    @objective(model, Max, index * o)
    optimize!(model)
    termination_status(model) == OPTIMAL && return value(o)
end
```

**Algorithm 44:** Compute bounds in BaB. The concrete bound is computed by sampling in the domain. Heuristically, we sample three points: lower bound of domain, upper bound of domain, and center of domain. The approximated bound is computed by minimizing the output constrained on triangle relaxation of the network.

**Result interpretation**  Similar to Sherlock, the original purpose of BaB is only to obtain tight bounds. To fit into our problem formulation (2.4), the obtained bounds are compared against the output constraint $\mathcal{Y}$. Define the bounding set $\mathcal{B} \coloneqq [\bar{\ell}_n, \underline{u}_n]$, which considers only the concrete bounds, and are tight with at most $\epsilon$ uncertainty. The reachable set is defined as $\tilde{\mathcal{R}} \coloneqq [\underline{\ell}_n, \bar{u}_n]$, which considers only the approximated bounds. Then the solution to the problem can be determined following the conditions in (9.4).

**Potential improvements**  Bunel *et al.* discussed several potential improvements of BaB though better bounding and better branching [14]. The node-wise bound is computed by `get_bounds` in `approx_bound` in algorithm 44. Our implementation uses interval arithmetic to compute the bounds and updates at every iteration. For better bounding, we may use other methods to compute tighter bounds. If such method is computationally expensive, it suffices to only update the bounds for essential nodes at every iteration. Meanwhile, for better branching, the function `split_dom` can use better heuristics to perform either input interval refinement or constraint refinement.[3] For example, BaBSB performs input interval refinement in a way that results in the tightest approximate output bounds after split. For computation efficiency, the approximate output bounds are evaluated analytically using the dual approach introduced in ConvDual and FastLin. BaBSR performs constraint refinement, which splits hidden nodes. The node to split is chosen such that the approximate output bounds are the tightest after split. Nonetheless, the evaluation of the heuristics can be computationally expensive when the size of the network grows. Lu and Kumar developed a method to learn the branching strategy using a graph neural network (GNN) by imitating the branching heuristics [41].

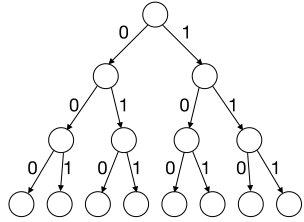

**Figure 9.4:** Illustration of binary tree search to assign values to $\delta_{i,j}$ for all nodes $j$ in layer $i$.

## 9.3 Planet

Planet [22] formulates the problem (2.4) as a satisfiability problem, which consists of trying to find an assignment for $\delta_i$ for all $i$ such that a point in the input set can be mapped to the complement of the output set. The search in Planet can be understood as a binary tree search as the assignment is either 0 or 1 as illustrated in figure 9.4. The novelty of Planet is that it uses linear programming to 1) infer tighter bounds on the nodes, 2) filter out conflicting assignments, and 3) infer more assignments given a partial assignment of $\delta_i$. In this way, the search can be more efficient.

The original implementation[4] is eager, which modifies the main loop in a state-of-the-art SAT solver (*i.e.*, MiniSAT) to impose network constraints during the search phase with partial assignments for $\delta_i$. If conflicts are detected in the partial assignments, the solver performs back-tracking.

For simplicity, we didn't follow the original eager implementation. Instead, we only check for conflicts when all $\delta$ have been assigned. We call PicoSAT.jl[5] to solve the SAT problem for assignment. In this way, we can leave the SAT solver untouched. The main loop in algorithm 45 is different from the Planet paper. In the non-eager implementation, the node-wise bounds are computed first. Planet introduces an optimization approach to compute tighter bounds, which will be discussed below. Then a clause $\Psi$ is initialized with respect to the bounds. The clause can be understood as a pruned search tree. A full assignment of all $\delta$'s according to the clause is computed by calling PicoSAT. The feasibility of this assignment is checked by elastic filtering, which returns conflicts in the assignment if the assignment is not feasible. Then the conflicts are added to the clause, which helps further pruning the search tree. If the assignment is feasible, that means there is an activation pattern that maps a desired input to an undesired output. Then the property is violated. If no such assignment exists, the property holds.

---

[3] Recall constraint refinement is discussed in section 4.2.
[4] https://github.com/progirep/planet
[5] https://github.com/jakebolewski/PicoSAT.jl

```
struct Planet
    optimizer
    eager::Bool
end

function solve(solver::Planet, problem::Problem)
    status, bounds = tighten_bounds(problem, solver.optimizer)
    status == OPTIMAL || return CounterExampleResult(:holds)
    ψ = init_ψ(problem.network, bounds)
    δ = PicoSAT.solve(ψ)
    opt = solver.optimizer
    while δ != :unsatisfiable
        status, conflict = elastic_filtering(problem, δ, bounds, opt)
        if status != INFEASIBLE
            return CounterExampleResult(:violated, conflict)
        end
        push!(ψ, conflict)
        δ = PicoSAT.solve(ψ)
    end
    return CounterExampleResult(:holds)
end
```

**Algorithm 45:** Main loop in Planet. This is a non-eager implementation. First, tight bounds of node values are computed through `tighten_bounds`. Then the clause $\Psi$ is initialized with respect to the bounds. A full assignment according to the clause is computed by calling `PicoSAT`. The feasibility of this assignment is checked by `elastic_filtering`, which returns conflicts in the assignment if the assignment is not feasible. Then the conflicts are added to the clause. The while loop repeats previous steps. If the assignment is feasible, that means there is an activation pattern that maps a desired input to an undesired output. Then the property is violated. If no such assignment exists, the property is violated.

**Computing tight bounds** We first compute the bounds through the function `tighten_bounds` in algorithm 46. This function first gets relatively loose bounds $\hat{\boldsymbol{\ell}}_i$ and $\hat{\mathbf{u}}_i$ from `get_bounds` in algorithm 5. Then the following optimization problem is solved for every layer $i \in \{1, \ldots, n\}$ and node $j \in \{1, \ldots, k_i\}$,

$$\min_{\mathbf{z}_0, \ldots, \mathbf{z}_n, \hat{\mathbf{z}}_1, \ldots, \hat{\mathbf{z}}_n} q z_{i,j}, \tag{9.6a}$$

$$\text{s.t. } \mathbf{z}_0 \in \mathcal{X}, \mathbf{z}_n \notin \mathcal{Y}, \tag{9.6b}$$

$$z_{i,j} = \hat{z}_{i,j}, \forall i \in \{1, \ldots, n-1\}, j \in \Gamma_i^+, \tag{9.6c}$$

$$z_{i,j} = 0, \forall i \in \{1, \ldots, n-1\}, j \in \Gamma_i^-, \tag{9.6d}$$

$$z_{i,j} \geq \hat{z}_{i,j}, z_{i,j} \geq 0, z_{i,j} \leq \frac{\hat{u}_{i,j}(\hat{z}_{i,j} - \hat{\ell}_{i,j})}{\hat{u}_{i,j} - \hat{\ell}_{i,j}},$$

$$\forall i \in \{1, \ldots, n-1\}, j \in \Gamma_i, \tag{9.6e}$$

$$\hat{\mathbf{z}}_i = \mathbf{W}_i \mathbf{z}_{i-1} + \mathbf{b}_i, \forall i \in \{1, \ldots, n\}, \tag{9.6f}$$

where $q \in \{-1, 1\}$. When $q = -1$, we compute the tighter upper bounds. When $q = 1$, we get the lower bounds. Here we use triangle relaxation in (6.6). Note that we update the encoding with newly computed bounds from the optimization to solve for even tighter bounds in later nodes. If there is no solution in (9.6), that means the constraints are infeasible. Hence, for all $\mathbf{x} \in \mathcal{X}$, $\mathbf{y} = \mathbf{f}(\mathbf{x}) \in \mathcal{Y}$. Then the property is satisfied. If there is a solution for (9.6), then we need to further solve the SAT problem.

**Solving the SAT problem** A clause $\Psi$ corresponds to a search tree, which encodes a set of binary conditions. According to the bounds computed earlier, the clause can be initialized as

$$\Psi = \bigwedge_{\hat{\ell}_{i,j} \leq 0 \leq \hat{u}_{i,j}} [\{\delta_{i,j} = 1\} \vee \{\delta_{i,j} = 0\}] \bigwedge_{0 < \hat{\ell}_{i,j}} [\delta_{i,j} = 1] \bigwedge_{\hat{u}_{i,j} < 0} [\delta_{i,j} = 0]. \tag{9.7}$$

Solving the SAT problem means finding an assignment of $\delta$'s such that the clause $\Psi$ is true. The case that $\Psi = \bigwedge_{i,j} [\{\delta_{i,j} = 1\} \vee \{\delta_{i,j} = 0\}]$ corresponds to a full search tree shown in figure 9.4. As more conditions are added to the clause, we are essentially pruning the search tree. Elastic filtering, to be discussed below is used to help the pruning. There may be more than one solution that satisfies the clause $\Psi$. We only consider one solution at a time. PicoSAT is called to find a feasible assignment for $\delta_{i,j}$'s.

```julia
function tighten_bounds(problem::Problem, optimizer; act=true)
    output = problem.output
    bounds = get_bounds(problem, false)
    post_activation_bounds = Vector{Hyperrectangle}(undef, length(bounds))
    for i in 2:length(problem.network.layers) + 1
        layer = problem.network.layers[i-1]
        lower = low(bounds[i])
        upper = high(bounds[i])
        model = Model(optimizer)
        neurons = init_neurons(model, problem.network)
        add_set_constraint!(model, problem.input, first(neurons))
        add_complementary_set_constraint!(model, output, last(neurons))
        encode_network!(model, problem.network, neurons, bounds,
                        TriangularRelaxedLP(); pre_activation=true)
        for j in 1:length(neurons[i])
            neuron = neurons[i][j]
            objective = dot(layer.weights[j, :], neurons[i-1]) + layer.bias[j]
            @objective(model, Min, objective)
            optimize!(model)
            termination_status(model) == OPTIMAL
                || return (INFEASIBLE, bounds)
            lower[j] = value(objective)
            @objective(model, Max, objective)
            optimize!(model)
            upper[j] = value(objective)
        end
        bounds[i] = Hyperrectangle(low = lower, high = upper)
        post_activation_bounds[i] = bounds[i]
        if (layer.activation == ReLU())
            post_activation_bounds[i] = rectify(bounds[i])
        end
    end
    return (OPTIMAL, act ? post_activation_bounds : bounds)
end
```

**Algorithm 46:** Obtaining tighter bounds using optimization. The solution is tighter than `get_bounds` in algorithm 5. The implementation encodes equation (9.6) and solves the optimization problem by calling the `JuMP` solver.

**Elastic filtering**   Given an assignment of $\delta_{i,j}$'s, we check for conflicts using `elastic_filtering` in algorithm 47. The problem is defined as

$$\min_{\mathbf{z}_0,\ldots,\mathbf{z}_n,\hat{\mathbf{z}}_1,\ldots,\hat{\mathbf{z}}_n,\mathbf{s}_1,\ldots,\mathbf{s}_n} q \sum_{i=1}^{n} \sum_{j=1}^{k_i} s_{i,j}, \tag{9.8a}$$

$$\text{s.t.} \quad \mathbf{z}_0 \in \mathcal{X}, \mathbf{z}_n \notin \mathcal{Y}, \tag{9.8b}$$

$$z_{i,j} = \hat{z}_{i,j}, \forall i \in \{1,\ldots,n-1\}, j \in \Gamma_i^+, \tag{9.8c}$$

$$z_{i,j} = 0, \forall i \in \{1,\ldots,n-1\}, j \in \Gamma_i^-, \tag{9.8d}$$

$$z_{i,j} \geq \hat{z}_{i,j}, z_{i,j} \geq 0, z_{i,j} \leq \frac{\hat{u}_{i,j}(\hat{z}_{i,j} - \hat{\ell}_{i,j})}{\hat{u}_{i,j} - \hat{\ell}_{i,j}},$$

$$\forall i \in \{1,\ldots,n-1\}, j \in \Gamma_i, \tag{9.8e}$$

$$\hat{\mathbf{z}}_i = \mathbf{W}_i \mathbf{z}_{i-1} + \mathbf{b}_i, \forall i \in \{1,\ldots,n\}, \tag{9.8f}$$

$$z_{i,j} = \hat{z}_{i,j} + s_{i,j} \geq 0, \text{ for } \delta_{i,j} = 1,$$

$$\forall i \in \{1,\ldots,n\}, j \in \{1,\ldots,k_i\}, \tag{9.8g}$$

$$z_{i,j} = 0, \hat{z}_{i,j} - s_{i,j} \leq 0, \text{ for } \delta_{i,j} = 0,$$

$$\forall i \in \{1,\ldots,n\}, j \in \{1,\ldots,k_i\}, \tag{9.8h}$$

which uses both triangle relaxation (given bounds) and slack relaxation (given activation). If the assignment of $\delta_{i,j}$'s are feasible, then all the slack variables should be non-positive. If some slack variables are positive, we fix the node with the largest slack value by adding a condition $s_{i,j} = 0$. Then the optimization problem is solved again and again until there is no feasible solution. The list of node $\{(i^{(1)}, j^{(1)}), (i^{(2)}, j^{(2)}), \ldots, (i^{(k)}, j^{(k)})\}$ and their $\delta$ assignments that we fixed during the process is a conflict. A conflict means that the $\delta$ assignments of those nodes cannot be satisfied simultaneously. The intuition behind elastic filtering is that we can find conflicts faster by fixing the most violated constraints.

Once a conflict is determined with respect to an assignment, we can add the conflict back to the clauses and solve the SAT problem again. If there is no feasible solution of $\Psi$, the property holds. If no conflict is found for a specific assignment, the property is violated. Our implementation is still sound and complete, though it is less efficient than the original eager implementation. Our implementation requires that input set $\mathcal{X}$ be a `Hyperrectangle` and the output set $\mathcal{Y}$ be a `PolytopeComplement`.

## 9.4   Reluplex

Reluplex [32] applies a simplex algorithm to ReLU networks. It searches for a counter example (2.12).

```
function elastic_filtering(problem, δ, bounds, optimizer)
    model = Model(optimizer)
    nnet = problem.network
    neurons = init_neurons(model, nnet)
    add_set_constraint!(model, problem.input, first(neurons))
    add_complementary_set_constraint!(model, problem.output, last(neurons))
    encode_network!(model, nnet, neurons, bounds, TriangularRelaxedLP())
    SLP = encode_network!(model, nnet, neurons, δ, SlackLP())
    slack = SLP.slack
    min_sum!(model, slack)
    conflict = Vector{Int64}()
    act = get_activation(nnet, bounds)
    while true
        optimize!(model)
        termination_status(model) == OPTIMAL || return (INFEASIBLE, conflict)
        (m, index) = max_slack(value.(slack), act)
        m > 0.0 || return (:Feasible, value.(neurons[1]))
        coeff = δ[index[1]][index[2]] ? -1 : 1
        node = coeff * get_node_id(nnet, index)
        push!(conflict, node)
        @constraint(model, slack[index[1]][index[2]] == 0.0)
    end
end
```

**Algorithm 47:** Elastic filtering is used to determine conflicts in an assignment of all $\delta$'s. It iteratively solves equation (9.8). At each iteration, it fixes the largest slack variable to zero. The conflicting sequence is found once the optimization becomes infeasible.

For each node $(i,j)$, Reluplex optimizes over two variables $z_{i,j}$ and $\hat{z}_{i,j}$. There are three possible statuses for each node $(i,j)$, active (denoted $(i,j) \in \mathcal{A}$), inactive (denoted $(i,j) \in \mathcal{N}$), and undetermined (denoted $(i,j) \in \mathcal{U}$). Reluplex tries to find a feasible assignment for undetermined nodes through depth-first search. The basic constraints in Reluplex without considering the undetermined nodes are

$$\mathcal{B} = \{\mathbf{z}_i, \hat{\mathbf{z}}_i : \mathbf{z}_0 \in \mathcal{X}, \mathbf{z}_n \notin \mathcal{Y}, \tag{9.9a}$$
$$z_{i,j} \geq 0, z_{i,j} \geq \hat{z}_{i,j}, \hat{\ell}_{i,j} \leq \hat{z}_{i,j} \leq \hat{u}_{i,j}, \tag{9.9b}$$
$$z_{i,j} = \hat{z}_{i,j}, \hat{z}_{i,j} \geq 0, \forall (i,j) \in \mathcal{A}, \tag{9.9c}$$
$$z_{i,j} = 0, \hat{z}_{i,j} < 0, \forall (i,j) \in \mathcal{N}, \tag{9.9d}$$
$$\hat{\mathbf{z}}_i = \mathbf{W}_i \mathbf{z}_{i-1} + \mathbf{b}_i, \forall i \in \{1, \ldots, n\}\}, \tag{9.9e}$$

At each search step, Reluplex either finds a counter example, or determines that there is no such counter example (then does backtracking), or assigns one node $(i^*, j^*) \in \mathcal{U}$ to be active or inactive. The set $\mathcal{B}^0 := \mathcal{B}$, $\mathcal{A}^0 := \mathcal{A}$, $\mathcal{N}^0 := \mathcal{N}$, and $\mathcal{U}^0 := \mathcal{U}$ are initialized at the beginning of the search. At search depth $k$, we have a set of constraints $\mathcal{B}^k$, which is an intersection of the basic constraint $\mathcal{B}$ and the constraint induced by the assignments of nodes in $\mathcal{U}^0$ during the search. $\mathcal{B}^k$ can be represented in the form of (9.9) by adding superscript $k$ to $\mathcal{B}$, $\mathcal{A}$, and $\mathcal{N}$. The relationships among the sets at different depths are

$$\mathcal{A}^0 \subseteq \mathcal{A}^1 \subseteq \cdots \subseteq \mathcal{A}^k, \tag{9.10a}$$
$$\mathcal{N}^0 \subseteq \mathcal{N}^1 \subseteq \cdots \subseteq \mathcal{N}^k, \tag{9.10b}$$
$$\mathcal{U}^0 \supset \mathcal{U}^1 \supset \cdots \supset \mathcal{U}^k. \tag{9.10c}$$

At each search step $k$, we first solve a feasibility problem to check if there is a solution of $\mathcal{B}^{k-1}$. If there is no feasible solution that satisfies $\mathcal{B}^{k-1}$, we do back tracking. If there exists a solution that satisfies $\mathcal{B}^{k-1}$, we consider the following two possibilities.

- If all ReLU constraints are satisfied, *i.e.*, $z_{i,j} = \max\{\hat{z}_{i,j}, 0\}$, then we indeed find a counter example for the problem (2.4).

- If some ReLU constraints are not satisfied, *i.e.*, $z_{i,j} \neq \max\{\hat{z}_{i,j}, 0\}$, then we need to fix those broken nodes during the search.[6] We pick such a broken node $(i^k, j^k)$ from $\mathcal{U}^{k-1}$. Then $\mathcal{U}^k = \mathcal{U}^{k-1} \setminus \{(i^k, j^k)\}$. We may assign the node to be either active or inactive. In the active case, $\mathcal{A}^k = \mathcal{A}^{k-1} \cup \{(i^k, j^k)\}$, $\mathcal{N}^k = \mathcal{N}^{k-1}$, and

$$\mathcal{B}^k := \mathcal{B}^{k-1} \cap \{\mathbf{z}_i, \hat{\mathbf{z}}_i : z_{i^k, j^k} = \hat{z}_{i^k, j^k} \geq 0\}. \tag{9.11}$$

---

[6]In the original Reluplex implementation, the solver tries a greedy fix (*i.e.*, change the variable value without doing a case split) for such broken ReLUs before doing the case split that we describe in (9.11) and (9.12). And this greedy approach is performed until some ReLU gets fixed at least $N$ times, for some threshold $N$. Once the threshold is reached, the solver then proceeds with a case split. Our implementation here is essentially an instance of the more general algorithm with $N = 1$.

In the inactive case, $\mathcal{A}^k = \mathcal{A}^{k-1}$, $\mathcal{N}^k = \mathcal{N}^{k-1} \cup \{(i^k, j^k)\}$, and

$$\mathcal{B}^k := \mathcal{B}^{k-1} \cap \{\mathbf{z}_i, \hat{\mathbf{z}}_i : z_{i^k, j^k} = 0, \hat{z}_{i^k, j^k} \leq 0\}. \tag{9.12}$$

During the search, we either end up finding a counter example or concluding that there is no such counter example. The worst case scenario complexity is $2^{|\mathcal{U}|}$, *i.e.*, we traverse a depth $|\mathcal{U}|$ binary tree, where $|\mathcal{U}|$ is the cardinality of the set $\mathcal{U}$. It is possible to use elastic filtering introduced in Planet to help detect conflicts faster.

Our implementation supports input sets $\mathcal{X}$ of type `Hyperrectangle` and output sets $\mathcal{Y}$ of type `PolytopeComplement` and is shown in algorithm 48.[7] The depth-first search is performed in `reluplex_step`. The node to fix is selected by `find_relu_to_fix` in algorithm 49. The construction of $\mathcal{B}^k$ at each search step is done by `encode` in algorithm 50. The inputs to `reluplex_step` include,

- `model`, which encodes the constraint $\mathcal{B}^k$ similar to (9.9).

- `bs` and `fs`, which are $\hat{\mathbf{z}}$'s and $\mathbf{z}$'s.

- `relu_status`, which can be either 0, 1 or 2. If it is 0, the corresponding node belongs $\mathcal{U}^k$. If it is 1, the corresponding node belongs $\mathcal{A}^k$. If it is 2, the corresponding node belongs to $\mathcal{N}^k$.

## 10  Comparison and Results

This chapter presents experimental results for our implementation of the algorithms.[1] We present results on the runtime performance of each algorithm as well as an empirical comparison of different bound tightening techniques employed in the approaches. Different algorithms handle problems with different specifications as shown in table 2.1. As described previously, the objects used to represent the input set $\mathcal{X}$ and the output set $\mathcal{Y}$ vary along with the characteristics of the sets that each of the algorithms support. Additionally, different algorithms output different types of results. We have, consequently, split the algorithms in six groups such that the same verification problem can be solved by all the algorithms within a group, facilitating a comparison of the implementations provided with this work. The groups are as follows:

1. Ai2, ExactReach, and maxSens.

---

[7] The original Reluplex uses a data structure called a tableau to encode new LP constraints. Conceptually, our Julia implementation is the same as the original implementation, but in a less efficient way.

[1] In our implementations, readability was favored over speed. Some of our implementations are simplified versions of the original ones that still output the same results but can be slower.

```julia
struct Reluplex
    optimizer
end
function solve(solver::Reluplex, problem::Problem)
    initial_model = Model(solver)
    bs, fs = encode(solver, initial_model, problem)
    layers = problem.network.layers
    status = [zeros(Int, n) for n in n_nodes.(layers)]
    insert!(initial_status, 1, zeros(Int, dim(problem.input)))
    return reluplex_step(solver, problem, initial_model, bs, fs, status)
end
function reluplex_step(solver::Reluplex, problem, model,
            ẑ::Vector{Vector{VariableRef}}, z::Vector{Vector{VariableRef}},
            relu_status::Vector{Vector{Int}})
    optimize!(model)
    if termination_status(model) == OPTIMAL
        i, j = find_relu_to_fix(ẑ, z)
        i == 0 && return CounterExampleResult(:violated, value.(first(ẑ)))
        for repair! in (type_one_repair!, type_two_repair!)
            new_constraints = repair!(model, ẑ[i][j], z[i][j])
            result = reluplex_step(solver, problem, model, ẑ, z, relu_status)
            delete.(model, new_constraints)
            result.status == :violated && return result
        end
    end
    return CounterExampleResult(:holds)
end
function type_one_repair!(model, ẑᵢⱼ, zᵢⱼ)
    con_one = @constraint(model, ẑᵢⱼ == zᵢⱼ)
    con_two = @constraint(model, ẑᵢⱼ >= 0.0)
    return con_one, con_two
end
function type_two_repair!(model, ẑᵢⱼ, zᵢⱼ)
    con_one = @constraint(model, ẑᵢⱼ <= 0.0)
    con_two = @constraint(model, zᵢⱼ == 0.0)
    return con_one, con_two
end
```

**Algorithm 48:** Main loop in Reluplex. The function `reluplex_step` performs depth-first search. At every search step, it first solves a feasibility problem encoded in `model`. If the problem is infeasible, meaning that no counter example can be found, we output `:holds`. If there is a solution of the problem, we find the first node such that the ReLU activation is broken, i.e., $z_{i,j} \neq [\hat{z}_{i,j}]_+$. If no such broken node is found, meaning that we have found a counter example, we output `:violated` with the counter example. For the broken node, we can either change its status to 1 or 2, which corresponds to the two branches in the search. For each branch, the model will be updated by adding new constraints. Then the search continues.

```julia
function find_relu_to_fix(ẑ, z)
    for i in 1:length(z), j in 1:length(z[i])
        ẑᵢⱼ = value(ẑ[i][j])
        zᵢⱼ = value(z[i][j])
        if type_one_broken(ẑᵢⱼ, zᵢⱼ) ||
            type_two_broken(ẑᵢⱼ, zᵢⱼ)
             return (i, j)
        end
    end
    return (0, 0)
end
type_one_broken(ẑᵢⱼ, zᵢⱼ) = (zᵢⱼ > 0.0)  &&
    (!(-0.0 < ẑᵢⱼ - zᵢⱼ < 0.0)) # (zᵢⱼ > 0) && (zˆᵢⱼ != zᵢⱼ)
type_two_broken(ẑᵢⱼ, zᵢⱼ) = (-0.0 < zᵢⱼ < 0.0) &&
    (ẑᵢⱼ > 0.0) # (zᵢⱼ == 0) && (zˆᵢⱼ > 0)
```

**Algorithm 49:** The function to select the next ReLu node to fix in Reluplex. It returns the first broken node by checking all nodes from the first layer to the last layer.

```julia
function encode(solver::Reluplex, model::Model, problem::Problem)
    layers = problem.network.layers
    ẑ = init_neurons(model, layers) # before activation
    z = init_neurons(model, layers) # after activation
    activation_constraint!(model, ẑ[1], z[1], Id())
    bounds = get_bounds(problem)
    for (i, L) in enumerate(layers)
        @constraint(model, affine_map(L, z[i]) .== ẑ[i+1])
        add_set_constraint!(model, bounds[i], z[i])
        activation_constraint!(model, ẑ[i+1], z[i+1], L.activation)
    end
    add_set_constraint!(model, last(bounds), last(z))
    add_complementary_set_constraint!(model, problem.output, last(z))
    feasibility_problem!(model)
    return ẑ, z
end
```

**Algorithm 50:** Encoding the optimization problem in Reluplex. The optimization problem has zero objective and is constrained on $\mathcal{B}$ in equation (9.9). In the search process, more constraints are to be added by repair!, which encode constraints according to relu_status.

Input: `HPolytope`.

Output: `HPolytope` (bounded).

2. ILP, MIPVerify, and NSVerify.

    Input: `Hyperrectangle`.

    Output: `PolytopeComplement`.

3. Duality and convDual.

    Input: `Hyperrectangle` (uniform radius).

    Output: `Halfspace`.

4. FastLin, FastLip, ILP, and MIPVerify.

    Input: `Hyperrectangle`.

    Output: `Halfspace`.

5. BaB, DLV, ReluVal, and Sherlock.

    Input: `Hyperrectangle`.

    Output: `Hyperrectangle` (1-D).

6. Planet, Reluplex, and ReluVal.

    Input: `Hyperrectangle`.

    Output: `PolytopeComplement` and `Hyperrectangle`.

Despite having different problem specifications, the exact same property can be encoded across Groups 2, 3, 4, and 6 by using their corresponding input and output sets to represent the same constraints.

Along with performance tests, we compared a variety of techniques to find bounds. The approaches to compute these bounds are listed below:

1. Interval arithmetic, described in section 4.1, which is used by NSVerify, MIPVerify, DLV, and Duality.

2. Planet's tighten bounds, which uses a linear relaxation of the problem which is described in section 9.3.

3. ReluVal's Symbolic Interval Propagation described in section 8.1. ReluVal's symbolic bounds at each layer can be concretized into a hyperrectangle to provide node-wise bounds.

4. Neurify's implementation of Symbolic Linear Relaxation described in section 8.2. Similar to ReluVal, Neurify's symbolic bounds at each layer can be concretized then overapproximated by a hyperrectangle to provide node-wise bounds.

5. The reachable sets propagated by Ai2, described in section 5.3 can be overapproximated by a hyperrectangle and interpeted as node-wise bounds. We see that Ai2 with zonotopes results in equivalent bounds to ConvDual, while Ai2 on the Box domain results in equivalent bounds to interval arithmetic.

6. ConvDual's get bounds function directly provides bounds on each node and is described in section 7.3. This form of bound propagation is also used by FastLin which is presented in section 8.3

7. A combination of the above techniques, where the bounds from Ai2z are used to perform the triangular relaxation for Planet's tighten bounds instead of interval arithmetic.

To compute the ground truth bounds for each node, MIPVerify's mixed-integer encoding with bound tightening from Ai2z was applied to maximize then minimize the output of each node. Gurobi was used to solve the resulting mixed integer program. We found that using Ai2z bounds instead of LP bounds sped up the calculation of these ground truth bounds. This process gave a tight upper and lower bound on each node, and provided a baseline against which to compare the other algorithms. Alternatively, we could have computed tight bounds using ExactReach since it provides the exact reachable set for each layer. However, in our implementation ExactReach quickly becomes intractable as the number of nodes increases.

## 10.1 Bound Experiments

First we describe how we arrived at each set of bounds. For interval arithmetic and ConvDual node-wise pre-activation bounds were already being computed. Planet's tighten bounds computed post-activation bounds, but with a slight modification of the implementation it could find pre-activation bounds instead. For ReluVal, Neurify, and Ai2 the pre-activation reachable set at each layer was overapproximated with a hyperrectangle in order to provide a bound on each node. For our combination technique where we combine Ai2z and Planet's tighten bounds, we first computed Ai2z's node-wise pre-activation bounds. We then used these instead of interval arithmetic for the triangular relaxation.

We performed bound tightening on two networks. The first was a network trained on the MNIST dataset [35] with 10 hidden layers, each with 20 nodes. The second was a network with randomly generated weights and biases, with a single output, 10 hidden layers with 20 nodes each, and a single output. Note that the dimensions of the input and output differ between these two networks, with the MNIST network having 784 inputs and 1 output. For each network, we applied each bound tightening technique. For the MNIST network the input set was a hyperrectangle of radius 0.004, approximately 1 pixel, centered around an example image of a handwritten digit. For the random network the input set was a hyperrectangle centered at the origin with radius 0.1. We found the size of the interval for each node from each algorithm, divided it by the ground truth's interval, and plotted this on a log scale. In figures figure 10.1 and figure 10.2 the entire first layer is plotted, then the second, and so on. In this way we can see trends in performance as a function of the depth of the node in the network. We exclude the input nodes for these plots, since the bounds of the input are directly given by the query and match for each algorithm. Ai2 box is equivalent to interval arithmetic, and as a result gives the same bounds and was is excluded from the plot for simplicity. Ai2z results in the the same bounds as ConvDual although the equivalence is more surprising than with Ai2 box. As a result, we included both Ai2z and ConvDual in the plots although they do lead to equivalent bounds.

The runtime for each algorithm for both the MNIST and randomly generated networks is shown in table 10.1.

| Algorithm | MNIST | Random |
| --- | --- | --- |
| IA | 97 | 18 |
| Planet | $30.1 \cdot 10^6$ | $2.3 \cdot 10^6$ |
| Ai2z and Planet | $22.2 \cdot 10^6$ | $1.5 \cdot 10^6$ |
| Symbolic Interval Propagation (ReluVal) | 135000 | 201 |
| Symbolic Linear Relaxation (Neurify) | $64.0 \cdot 10^6$ | 24000 |
| Ai2 Box | 94 | 22 |
| Ai2z | 8700 | 55 |
| ConvDual | 1130 | 253 |

**Table 10.1:** Runtime in microseconds on MNIST and random network with 10 hidden layers, each with 20 nodes. The MNIST network has 784 inputs and 10 outputs while the random network has 1 input and 1 output. The subplot shows the earliest layers where some of Planet's bounds are tighter than ConvDual/Ai2z's bounds.

We observe that the bounds typically are in the following order from tightest to loosest: Planet / Ai2z + Planet, Ai2z / ConvDual, Symbolic Linear Relaxation (Neurify), Symbolic Interval Propagation (ReluVal), Interval Arithmetic / Ai2 Box. Most algorithms tend to provide looser bounds as they get deeper into the network, although at different rates. It is less clear whether ConvDual / Ai2z and Planet follow this trend. The change in their bounds is the least marked, and a larger network and wider variety of examples is needed to tell how they will behave deeper into a network. Using ai2z bounds for the triangular relaxed LP doesn't seem to change the bounds, although it does appear to decrease the runtime. This speedup may be the result of more nodes being known to be inactive or active rather than undetermined. Although the triangular LP relaxation provides the tightest bounds, it is also quite computationally expensive. We also notice a large runtime for Symbolic Linear Relaxation on the MNIST network, which may be specific to the method of propagating polytopes in our implementation. We note that bound propagation times are larger on the MNIST network, likely as a result of the larger input size (784 versus 1) and larger output size (10 versus 1) in the MNIST network.

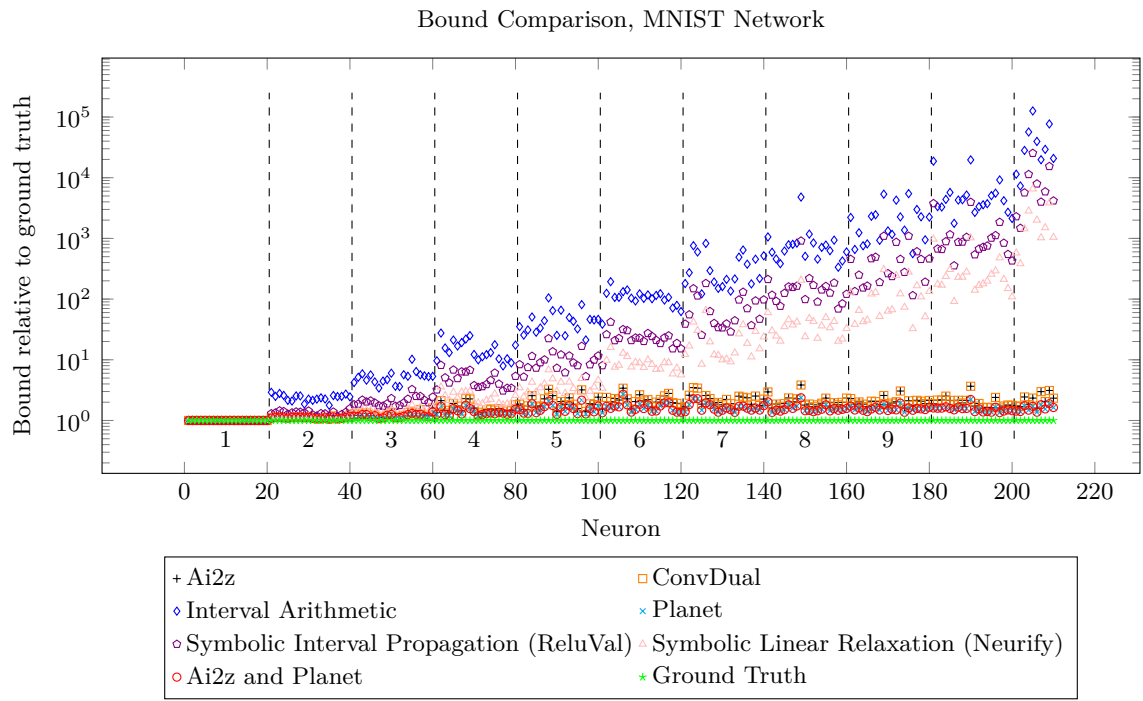

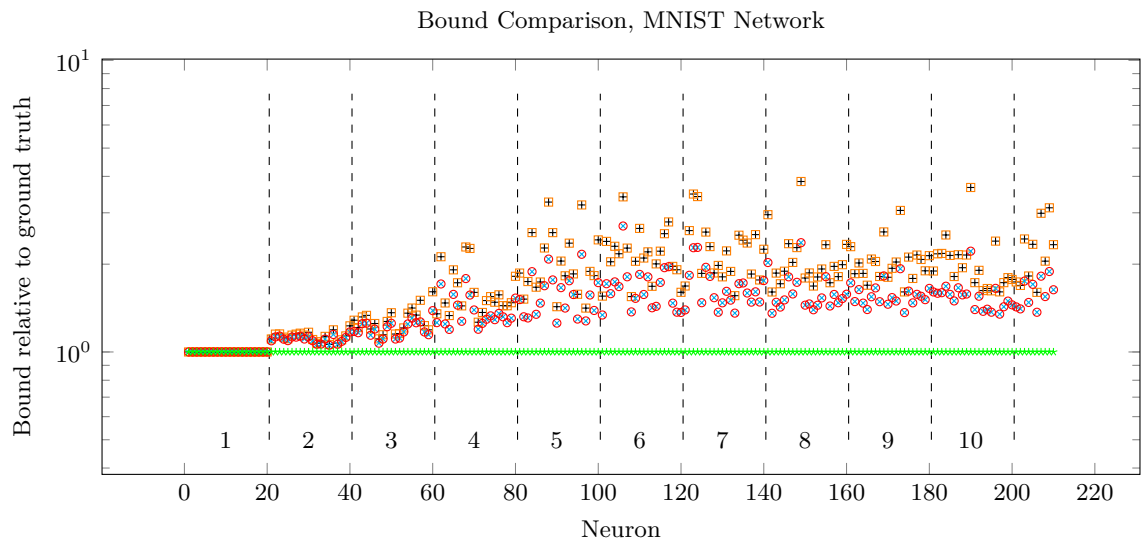

**Figure 10.1:** Bound Comparison on MNIST network with 10 hidden layers, each containing 20 nodes. The hidden layers are numbered and separated by dashed lines. The top figure shows all solvers, and the bottom shows the best performing solvers for easier comparison.

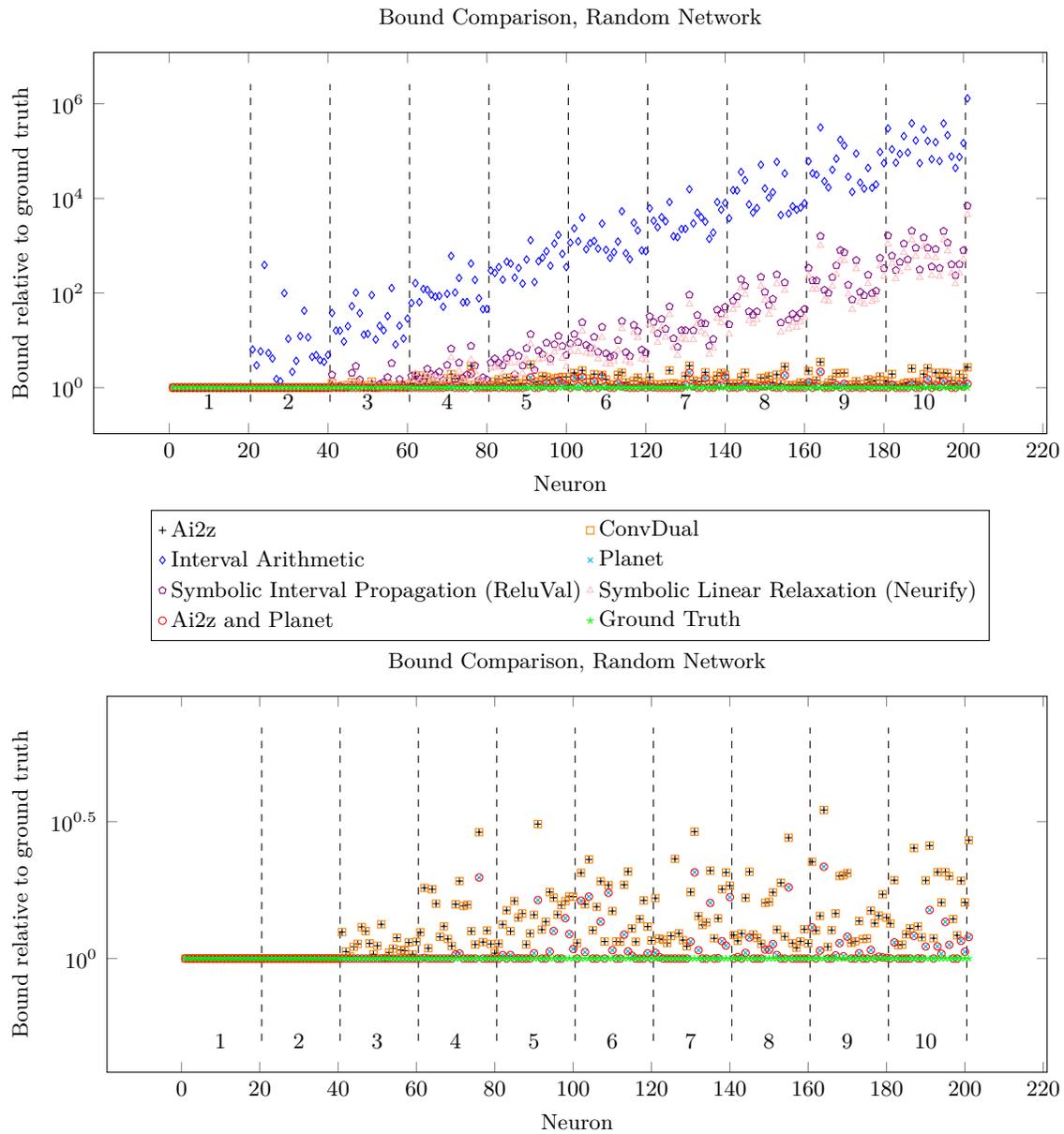

**Figure 10.2:** Bound Comparison on Random network with 10 hidden layers, each containing 20 nodes. The hidden layers are numbered and separated by dashed lines. The top figure shows all solvers, and the bottom shows the best performing solvers for easier comparison.

ConvDual and Ai2z resulted in the same bounds. This merits further investigation into whether the bound finding algorithm for ConvDual can be shown to be equivalent to propagating zonotopes through the network and using the hyperrectangle overapproximation of those zonotopes to find the bounds. Similarly, interval arithmetic and Ai2z box lead to the same bounds.

As these experiments were performed on just two networks, further work is necessary to see whether any of these results will hold in general and to draw a finer comparison between the different approaches. Different network architectures and input regions may lead to different behavior than we have observed here. The crucial role that bounds can play in limiting the runtime of complete verifiers makes this an interesting topic to pursue in the future.

## 10.2 Performance Experiments

We used different neural networks to benchmark the performance of the algorithms in different scenarios. We focused on three contexts: networks of varying sizes trained to classify hand-written digits from the MNIST dataset [35], the Aircraft Collision Avoidance System (ACAS) network [32].[2] and we created a tiny toy network (small nnet) for which we analytically derived its transfer function.

The experiments with algorithms in Group 5 required networks with a single output node. We used pruned versions of the exact same networks. We encoded the analogous properties in the dimension of the preserved output node. For the ACAS network this is the output corresponding to the cost of advising clear-of-conflict. For the MNIST networks we omitted algorithms that can require single output networks in order to evaluate the benchmarks established in [39]. The tiny toy network has a single output node.

For each experiment, we recorded the time that it took for each algorithm to terminate. We also recorded the result of each algorithm to identify cases in which incomplete algorithms failed to correctly identify properties that hold.

Group 1 supports hyperrectangle input sets. Groups 2, 3, 4, and 6 support H-polytope input sets. Group 5 supports hyperrectangle input sets and networks with only one output

---

[2]This network is based on a neural network trained on a very early prototype of ACAS Xu, targeted for unmanned aircraft. Details can be found in the article by Julian *et al.* [31].

node.

**Small nnet**  We manually specified a toy network and analytically derived the one dimensional function that it represents. We evaluated simple properties corresponding to upper and lower bounds of the image of this function for a small interval in the input set. The network has two hidden layers of two units each.

| Algorithm  | Time (s) | Output  |
|------------|----------|---------|
| ExactReach | 55.812   | :holds  |
| Ai2        | 250.53   | :holds  |
| maxSens    | 1.9789   | :holds  |
| NSVerify   | 0.0008   | :holds  |
| MIPVerify  | 0.0009   | :holds  |
| ILP        | 0.0006   | :holds  |
| convDual   | 0.0001   | :holds  |
| Duality    | 0.0009   | :holds  |
| FastLin    | 0.0001   | :holds  |
| FastLip    | 0.0002   | :holds  |
| DLV        | 0.0000   | :holds  |
| Sherlock   | 0.0025   | :holds  |
| BaB        | 0.0014   | :holds  |
| Planet     | 0.0006   | :holds  |
| Reluplex   | 0.0014   | :holds  |
| Reluval    | 0.0000   | :holds  |

**Table 10.2:** Experimental results for small nnet. All results are in seconds.

Table 10.2 contains the results for the experiments performed with Small nnet.

**MNIST**  For the MNIST benchmarks we used three fully-connected networks with 2, 4 and 6 layers, each with 256 hidden nodes and ReLU activation functions.[3]

- MNIST2.

  Input size: 748, 2 hidden layer of size: 256, output size: 10.

- MNIST4.

  Input size: 748, 4 hidden layer of size: 256, output size: 10.

- MNIST6.

  Input size: 748, 6 hidden layers of size: 25, output size: 10.

---

[3]https://github.com/verivital/vnn-comp/tree/master/2020/PWL/benchmark/mnist/oval

We verified the properties suggested in [39]. For each network and input image a local robustness property consisting of an $\epsilon$-perturbation of .02 and .05 is specified. The networks were trained with normalized inputs for each pixel in the range $[0, 1]$ which corresponds to dividing the values of the original images by 255.

We inspected a property associated to the correct classification of the image. The property requires that the logits (outputs of the final layer) correspond to correctly classifying the image. This means that if an image of digit $d$ is fed to the network then the $d$th output has to be maximal. The corresponding output set $\mathcal{Y}$ can be encoded as a polytope. Algorithms in Group 1 support this representation. For algorithms in Groups 2,3,4,5 and 6 we defined a subset of the property that can be encoded as a half-space. In this case, for an image of digit $d$ we verified that the $d$th output is greater than the output for digit $d + 1$, in the case where $d = 9$ we compared the 9th output to the first output which corresponds to $d = 0$.

For each solver we have a total of 150 properties which result from combining 3 networks, 2 values of $\epsilon$ and 25 images. Given that Group 1 and Groups 2, 3, 4, 5, 6 are verifying different properties we report their results separately.

We attempted to verify all 150 properties using all algorithms and recorded the time it took them to terminate. Each algorithm was allowed to run for at most 2,500 seconds per property. We noticed significant differences in the runtime distributions across different algorithms. Given that the there are many factors that determine the complexity of verifying a property (network architecture, size, input set geometry, output set geometry) and that different algorithms rely on fundamentally different approaches to verify the properties, we opted to aggregate the results for the 150 properties and compare the algorithms using statistics of the 150 experiments. Algorithms in Group 5 were not included in this experiments as they are only able to verify properties of networks with a single output. Algorithms in Group 6 did not terminate for any of the 150 properties.

In figure 10.3, we illustrate how many properties were processed by algorithms in Group 1 as a function of time. ExactReach did not terminate for any of the 150 properties. Figure 10.4 summarizes the distribution of the runtimes for Group 1.

In figure 10.5, we illustrate how many properties were processed by algorithms in Group 2, 3 and 4 as a function of time. Figure 10.6 summarizes the distribution of the runtimes for Group 2, 3 and 4.

In the above figures we can observe a significant variance in runtime across different algorithms. Below we analyze this results considering the output of the algorithms. Recall that not all algorithms are complete. Complete algorithms have to perform more work and therefore take longer to run. On the other hand, algorithms that heavily rely on over approximation or other techniques that favor efficiency over completeness can terminate in a very short time but fail to identify many properties that hold. Efficiency is not necessarily the most important characteristic of an algorithm. In some cases it can be unacceptable to consider that a property does not hold if an algorithm returns `:violated` without providing a counterexample. While complete algorithms never return `:violated` without certainty that the property does not hold, this is not the case with most algorithms. However, investigating the rate at which algorithms return `:violated` can inform users about the tradeoff between efficiency and completeness. To do this we defined the true positive rate as the fraction of properties that actually hold that the algorithm was able to identify.

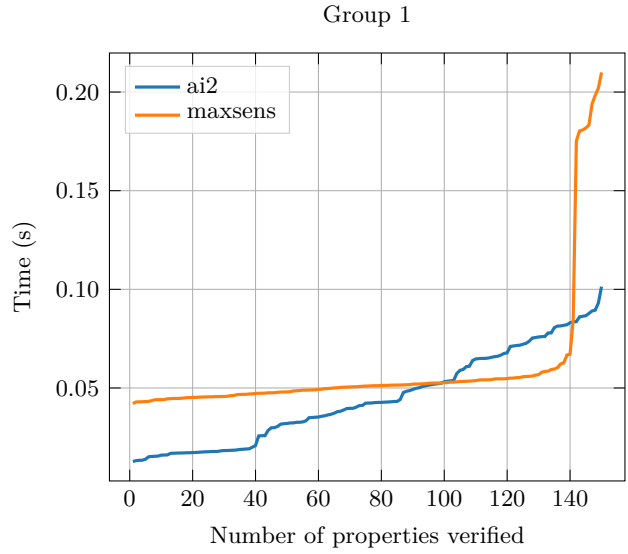

**Figure 10.3:** MNIST runtime results for Group 1.

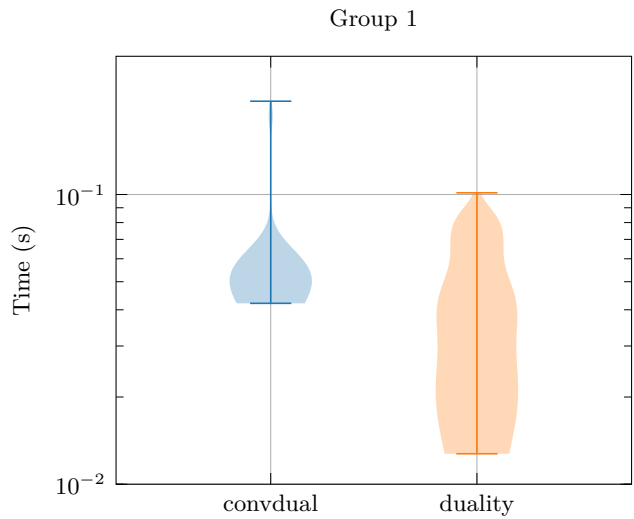

**Figure 10.4:** MNIST runtime distribution comparison for Group 1.

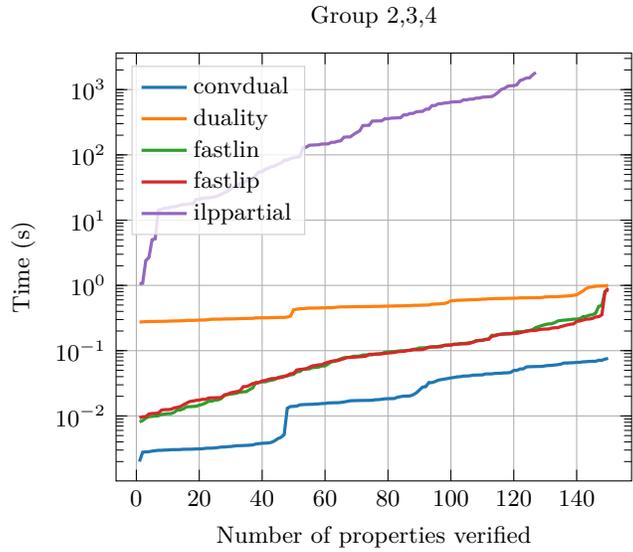

**Figure 10.5:** MNIST runtime results for Groups 2, 3, 4.

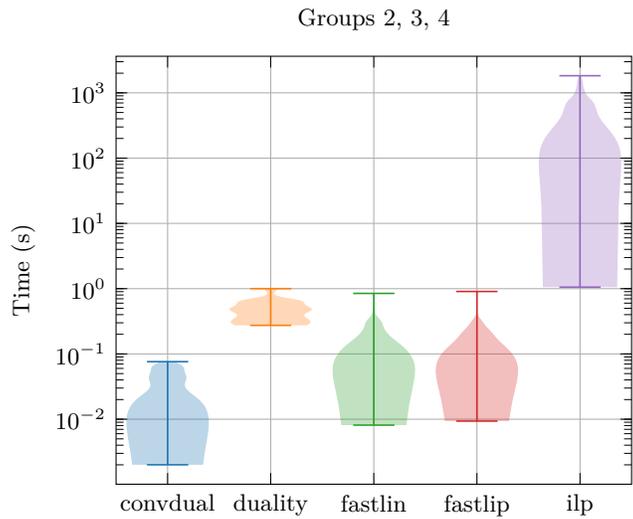

**Figure 10.6:** MNIST runtime distribution comparison for Groups 2, 3, 4.

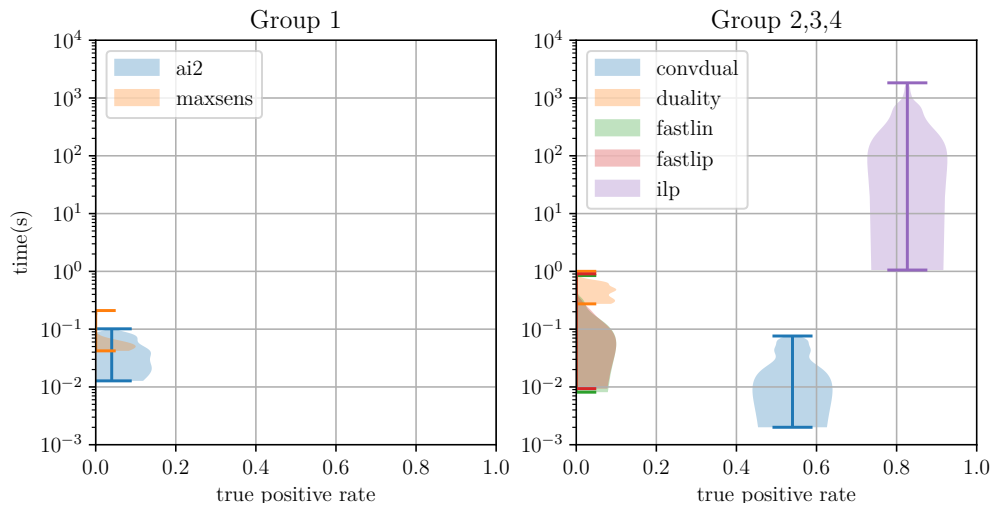

**Figure 10.7:** Runtime distributions as a function of true positive rate.

In figure 10.7, we show how the distribution of the runtime changes as a function of an algorithm's true positive rate. We can observe that overall algorithms that have a higher true positive rate tend to take significantly longer to terminate. There is a considerable difference of multiple orders of magnitude in the average runtime on both ends of the horizontal axis. Among the algorithms with a low true positive rate, on the lower end of the horizontal axis, we can observe a non monotonic relationship of runtime as a function of true positive rate. This unclear relationship can perhaps be explained by the relatively small sample size of 150 properties or by the different approximation schemes used by the algorithms.

**ACAS**  For the ACAS network, we verified property 10 introduced by Katz *et al.* [32]. Property 10 corresponds to the situation where the intruder aircraft is far away from the ownship and the desired output is that the advisory is clear-of-conflict. To make the property faster to verify, we reduced the volume of the input region by fixing the last three inputs of the network to specific values instead of the ranges originally defined for the property. This property has been verified in prior work [32, 65]. For the algorithms that can only support half-space or polytope complement output sets, we encoded the region where the first output is less than the last output. The network has five input units, six hidden layers of 50 units each, and five output units.

| Algorithm  | Time (s) | Output    |
|------------|----------|-----------|
| ExactReach | timeout  | -         |
| Ai2        | timeout  | -         |
| maxSens    | 0.43926  | :violated |
| NSVerify   | timeout  | -         |
| MIPVerify  | timeout  | -         |
| ILP        | 0.01308  | :unknown  |
| convDual   | 0.00043  | :holds    |
| Duality    | 0.05333  | :holds    |
| FastLin    | timeout  | -         |
| FastLip    | timeout  | -         |
| Planet     | timeout  | :holds    |
| Reluplex   | 2911.1   | :holds    |
| Reluval    | 0.0225   | :unknown  |

**Table 10.3:** Experimental results for ACAS. All results are in seconds, missing entries correspond to experiments that timed-out. Time out threshold set to 24 hours.

Table 10.3 contains the results for the experiments performed with ACAS.

**Analysis**  The experimental results shown above demonstrate the capability of the pedagogical implementation to verify realistic networks. Many of the algorithms are able to verify properties for networks as large as the ACAS Xu networks, which has been used for prior benchmarks [32, 65]. Overall we observed that algorithms that are complete take a longer to run. Consequently, complete algorithms are more amenable, at least at this point and this implementation, to verifying properties of smaller networks. Algorithms that are not complete usually rely on over-approximations or other schemes that significantly reduce their computational cost. Incomplete algorithms are faster and can more easily be used to verify properties on larger networks.

Group 1 algorithms terminate very quickly as shown in figure 10.3. However, they also tend to have a very low true positive rate as shown in figure 10.7. All the algorithms in Group 6 timed out for the MNIST experiments. Additionally, we can can observe in table 10.3 that NSVerify, MIPVerify, Planet and Planet timed out too. This result is not surprising as completeness comes at a high computational cost. It is worth noting that 1) many of the algorithms that are not complete were able to terminate in significantly shorter amounts of time compared to their complete counterparts, but 2) many of them exhibited their in-completeness by returning :violated for properties that, in fact, hold.

Other algorithms, particularly in Group 3, were unable to reach a conclusion and returned `:unknown` as a result. Algorithms in Group 5 were only able to run on the tiny toy network. We also observed in figure 10.7 that algorithms that return fewer incorrect `:violated` results tend to take longer to run. This is consistent with the fact increasing accuracy requires additional work. Finally, algorithms that rely on optimization are susceptible to numerical stability issues and we observed this in the case of MIPVerify and Planet.

## 10.3 Conclusion

This article surveyed algorithms for verification of deep neural networks. A unified mathematical framework was introduced to verify satisfiability of a neural network given certain input and output constraints. Three basic verification methods were identified: reachability, optimization, and search. We classified existing methods into five induced categories according to their core methodologies, and pointed out the connections among them. In particular, we reviewed the following methods: 1) reachability methods: ExactReach, Ai2, and MaxSens; 2) primal optimization methods: NSVerify, MIPVerify, and ILP; 3) dual optimization methods: Duality, ConvDual, and Certify; 4) search and reachability methods: ReluVal, Neurify, FastLin, FastLip, and DLV; and 5) search and optimization methods: Sherlock, BaB, Planet, and Reluplex. In the numerical experiments, we compare methods that either use similar methodologies or can solve the same problems. In general, there is a trade-off between completeness of a verification algorithm and its scalability. Complete algorithms run slower on larger networks, while incomplete algorithms are more conservative. Pedagogical implementations of all these methods were provided in Julia. The connections and differences among different methods were pointed out. This article can serve as a tutorial for students and professionals interested in this emerging field as well as a benchmark to facilitate the design of new verification algorithms.

## Acknowledgments

This work is partially supported by the Center for Automotive Research at Stanford (CARS). The authors would like to thank many of the authors of the referenced papers for their help in clarifying their algorithms and reviewing early drafts of this survey: Weiming Xiang, Taylor Johnson, Hoang-Dung Tran, Martin Vechev, Gagandeep Singh, Alessio Lomuscio, Michael Akintunde, Osbert Bastani, Zico Kolter, Shiqi Wang, Huan Zhang, Xiaowei Huang, Rudy Bunel, Reudiger Ehlers, and Guy Katz. The authors would also like to thank Christian Schilling, Marcelo Forets, and Sebastian Guadalupe, the authors of LazySets.jl, for their implementation support; Tianhao Wei for his contribution in the implementation; and Amelia Hardy and Zongzhang Zhang for their comments.

# References

[1]  Akintunde, M. E., A. Kevorchian, A. Lomuscio, and E. Pirovano. 2019. "Verification of RNN-Based Neural Agent-Environment Systems". In: *AAAI Conference on Artificial Intelligence (AAAI)*.

[2]  Akintunde, M. E., A. Lomuscio, L. Maganti, and E. Pirovano. 2018. "Reachability Analysis for Neural Agent-Environment Systems". In: *International Conference on Principles of Knowledge Representation and Reasoning*.

[3]  Anderson, B. G., Z. Ma, J. Li, and S. Sojoudi. 2020a. "Tightened Convex Relaxations for Neural Network Robustness Certification". *ArXiv*. (2004.00570).

[4]  Anderson, R., J. Huchette, W. Ma, C. Tjandraatmadja, and J. P. Vielma. 2020b. "Strong mixed-integer programming formulations for trained neural networks". *Mathematical Programming*: 1–37.

[5]  Bak, S. 2020. "Execution-Guided Overapproximation (EGO) for Improving Scalability of Neural Network Verification". In: *International Workshop on Verification of Neural Networks*.

[6]  Bak, S., H.-D. Tran, K. Hobbs, and T. T. Johnson. 2020. "Improved Geometric Path Enumeration for Verifying ReLU Neural Networks". In: *International Conference on Computer-Aided Verification (CAV)*.

[7]  Barrett, C. and C. Tinelli. 2018. "Satisfiability modulo theories". In: *Handbook of Model Checking*. Springer. 305–343.

[8]  Bastani, O., Y. Ioannou, L. Lampropoulos, D. Vytiniotis, A. Nori, and A. Criminisi. 2016. "Measuring neural net robustness with constraints". In: *Advances in Neural Information Processing Systems (NIPS)*.

[9]  Bezanson, J., A. Edelman, S. Karpinski, and V. B. Shah. 2017. "Julia: A Fresh Approach to Numerical Computing". *SIAM Review*. 59(1): 65–98.

[10] Bogomolov, S., M. Forets, G. Frehse, K. Potomkin, and C. Schilling. 2019. "JuliaReach: a toolbox for set-based reachability". In: *ACM International Conference on Hybrid Systems: Computation and Control*.

[11] Boyd, S., L. El Ghaoui, E. Feron, and V. Balakrishnan. 1994. *Linear Matrix Inequalities in System and Control Theory*. SIAM.

[12] Bunel, R. R., I. Turkaslan, P. Torr, P. Kohli, and P. K. Mudigonda. 2018. "A unified view of piecewise linear neural network verification". In: *Advances in Neural Information Processing Systems*.

[13] Bunel, R., A. De Palma, A. Desmaison, K. Dvijotham, P. Kohli, P. H. Torr, and M. P. Kumar. 2020a. "Lagrangian Decomposition for Neural Network Verification". *Conference on Uncertainty in Artificial Intelligence (UAI)*.

[14] Bunel, R., J. Lu, I. Turkaslan, P. Kohli, P. Torr, and M. P. Kumar. 2020b. "Branch and bound for piecewise linear neural network verification". *Journal of Machine Learning Research.* 21(2020).

[15] Cheng, C.-H., G. Nührenberg, C.-H. Huang, and H. Ruess. 2018. "Verification of Binarized Neural Networks via Inter-neuron Factoring". In: *Verified Software. Theories, Tools, and Experiments.*

[16] Cheng, C.-H., G. Nührenberg, and H. Ruess. 2017. "Verification of binarized neural networks". *ArXiv.* (1710.03107).

[17] Dunning, I., J. Huchette, and M. Lubin. 2017. "JuMP: A Modeling Language for Mathematical Optimization". *SIAM Review.* 59(2): 295–320.

[18] Dutta, S., S. Jha, S. Sanakaranarayanan, and A. Tiwari. 2017. "Output range analysis for deep neural networks". *ArXiv.* (1709.09130).

[19] Dutta, S., S. Jha, S. Sankaranarayanan, and A. Tiwari. 2018. "Learning and Verification of Feedback Control Systems using Feedforward Neural Networks." In: *IFAC Conference on Analysis and Design of Hybrid Systems (ADHS).*

[20] Dvijotham, K., R. Stanforth, S. Gowal, T. Mann, and P. Kohli. 2018. "A dual approach to scalable verification of deep networks". In: *Conference on Uncertainty in Artificial Intelligence (UAI).*

[21] E. Botoeva, P. Kouvaros, J. Kronqvist, A. Lomuscio, and R. Misener. 2020. "Efficient Verification of Neural Networks via Dependency Analysis". In: *AAAI Conference on Artificial Intelligence (AAAI).*

[22] Ehlers, R. 2017. "Formal verification of piece-wise linear feed-forward neural networks". In: *International Symposium on Automated Technology for Verification and Analysis.*

[23] Elboher, Y. Y., J. Gottschlich, and G. Katz. 2020. "An abstraction-based framework for neural network verification". In: *International Conference on Computer Aided Verification.*

[24] Fazlyab, M., M. Morari, and G. J. Pappas. 2019. "Safety verification and robustness analysis of neural networks via quadratic constraints and semidefinite programming". *ArXiv.* (1903.01287).

[25] Gehr, T., M. Mirman, D. Drashsler-Cohen, P. Tsankov, S. Chaudhuri, and M. Vechev. 2018. "Ai2: Safety and robustness certification of neural networks with abstract interpretation". In: *IEEE Symposium on Security and Privacy (SP).*

[26] Goodfellow, I., Y. Bengio, A. Courville, and Y. Bengio. 2016. *Deep learning.* MIT Press.

[27] Hayhurst, K. J., D. S. Veerhusen, J. J. Chilenski, and L. K. Rierson. 2001. "A practical tutorial on modified condition/decision coverage".

[28] He, K., X. Zhang, S. Ren, and J. Sun. 2016. "Deep residual learning for image recognition". In: *IEEE Computer Society Conference on Computer Vision and Pattern Recognition (CVPR).*

[29] Henriksen, P. and A. Lomuscio. 2020. "Efficient Neural Network Verification via Adaptive Refinement and Adversarial Search". In: *European Conference on Artificial Intelligence (ECAI)*.

[30] Huang, X., M. Kwiatkowska, S. Wang, and M. Wu. 2017. "Safety verification of deep neural networks". In: *International Conference on Computer Aided Verification*.

[31] Julian, K., M. J. Kochenderfer, and M. P. Owen. 2019. "Deep neural network compression for aircraft collision avoidance systems". *AIAA Journal of Guidance, Control, and Dynamics*. 42(3): 598–608.

[32] Katz, G., C. Barrett, D. L. Dill, K. Julian, and M. J. Kochenderfer. 2017. "Reluplex: An efficient SMT solver for verifying deep neural networks". In: *International Conference on Computer Aided Verification*.

[33] Katz, G., D. A. Huang, D. Ibeling, K. Julian, C. Lazarus, R. Lim, P. Shah, S. Thakoor, H. Wu, A. Zeljić, *et al.* 2019. "The marabou framework for verification and analysis of deep neural networks". In: *International Conference on Computer Aided Verification*.

[34] Khedr, H., J. Ferlez, and Y. Shoukry. 2020. "Effective Formal Verification of Neural Networks using the Geometry of Linear Regions". *ArXiv*. (2006.10864).

[35] LeCun, Y. and C. Cortes. 2010. "MNIST handwritten digit database".

[36] LeCun, Y., D. Touresky, G. Hinton, and T. Sejnowski. 1988. "A theoretical framework for back-propagation". In: *Proceedings of the 1988 Connectionist Models Summer School*. Vol. 1.

[37] Leofante, F., N. Narodytska, L. Pulina, and A. Tacchella. 2018. "Automated Verification of Neural Networks: Advances, Challenges and Perspectives". *ArXiv*. (1805.09938).

[38] Li, J., J. Liu, P. Yang, L. Chen, X. Huang, and L. Zhang. 2019. "Analyzing Deep Neural Networks with Symbolic Propagation: Towards Higher Precision and Faster Verification". In: *Static Analysis*. Cham.

[39] Liu, C. and T. Johnson. 2020. "Neural Network Verifications Workshop, VNN-COMP". *International Conference on Computer-Aided Verification*.

[40] Lomuscio, A. and L. Maganti. 2017. "An approach to reachability analysis for feed-forward relu neural networks". *ArXiv*. (1706.07351).

[41] Lu, J. and M. P. Kumar. 2020. "Neural Network Branching for Neural Network Verification". In: *International Conference on Learning Representations*.

[42] Manning, C., M. Surdeanu, J. Bauer, J. Finkel, S. Bethard, and D. McClosky. 2014. "The Stanford CoreNLP natural language processing toolkit". In: *Annual Meeting of the Association for Computational Linguistics: System Demonstrations*.

[43] Mirman, M., T. Gehr, and M. Vechev. 2018. "Differentiable Abstract Interpretation for Provably Robust Neural Networks". In: *International Conference on Machine Learning (ICML)*.

[44] Mnih, V., K. Kavukcuoglu, D. Silver, A. A. Rusu, J. Veness, M. G. Bellemare, A. Graves, M. Riedmiller, A. K. Fidjeland, G. Ostrovski, *et al.* 2015. "Human-level control through deep reinforcement learning". *Nature.* 518(7540): 529.

[45] Narodytska, N., S. Kasiviswanathan, L. Ryzhyk, M. Sagiv, and T. Walsh. 2018. "Verifying Properties of Binarized Deep Neural Networks". In: *AAAI Conference on Artificial Intelligence (AAAI)*.

[46] Olden, J. D. and D. A. Jackson. 2002. "Illuminating the "black box": a randomization approach for understanding variable contributions in artificial neural networks". *Ecological Modelling.* 154(1-2): 135–150.

[47] Papernot, N., P. McDaniel, S. Jha, M. Fredrikson, Z. B. Celik, and A. Swami. 2016. "The limitations of deep learning in adversarial settings". In: *IEEE European Symposium on Security and Privacy (EuroS&P)*.

[48] Pei, K., Y. Cao, J. Yang, and S. Jana. 2017. "Deepxplore: Automated whitebox testing of deep learning systems". In: *Symposium on Operating Systems Principles.*

[49] Prabhakar, P. and Z. R. Afzal. 2019. "Abstraction based Output Range Analysis for Neural Networks". In: *Advances in Neural Information Processing Systems.*

[50] Raghunathan, A., J. Steinhardt, and P. Liang. 2018. "Certified Defenses against Adversarial Examples". In: *International Conference on Learning Representations.*

[51] Rubies-Royo, V., R. Calandra, D. M. Stipanovic, and C. Tomlin. 2019. "Fast neural network verification via shadow prices". *ArXiv.* (1902.07247).

[52] Salman, H., G. Yang, H. Zhang, C.-J. Hsieh, and P. Zhang. 2019. "A convex relaxation barrier to tight robustness verification of neural networks". In: *Advances in Neural Information Processing Systems.*

[53] Singh, G., R. Ganvir, M. Püschel, and M. Vechev. 2019a. "Beyond the Single Neuron Convex Barrier for Neural Network Certification". In: *Advances in Neural Information Processing Systems (NeurIPS)*.

[54] Singh, G., T. Gehr, M. Mirman, M. Püschel, and M. Vechev. 2018. "Fast and Effective Robustness Certification". In: *Advances in Neural Information Processing Systems (NeurIPS)*.

[55] Singh, G., T. Gehr, M. Puschel, and M. Vechev. 2019b. "An Abstract Domain for Certifying Neural Networks". In: *ACM Symposium on Principles of Programming Languages.*

[56] Singh, G., T. Gehr, M. Puschel, and M. Vechev. 2019c. "Boosting Robustness Certification of Neural Networks". In: *International Conference on Learning Representations.*

[57] Sun, Y., X. Huang, and D. Kroening. 2018. "Testing Deep Neural Networks". *ArXiv.* (1803.04792).

[58] Tian, Y., K. Pei, S. Jana, and B. Ray. 2018. "Deeptest: Automated testing of deep-neural-network-driven autonomous cars". In: *International Conference on Software Engineering.*

[59] Tjeng, V., K. Xiao, and R. Tedrake. 2017. "Evaluating robustness of neural networks with mixed integer programming". *ArXiv*. (1711.07356).

[60] Tran, H.-D., S. Bak, W. Xiang, and T. T. Johnson. 2020a. "Verification of Deep Convolutional Neural Networks Using ImageStars". *ArXiv*. (2004.05511).

[61] Tran, H.-D., D. M. Lopez, P. Musau, X. Yang, L. V. Nguyen, W. Xiang, and T. T. Johnson. 2019a. "Star-based reachability analysis of deep neural networks". In: *International Symposium on Formal Methods*.

[62] Tran, H.-D., P. Musau, D. M. Lopez, X. Yang, L. V. Nguyen, W. Xiang, and T. T. Johnson. 2019b. "Parallelizable reachability analysis algorithms for feed-forward neural networks". In: *IEEE/ACM International Conference on Formal Methods in Software Engineering (FormaliSE)*.

[63] Tran, H.-D., X. Yang, D. M. Lopez, P. Musau, L. V. Nguyen, W. Xiang, S. Bak, and T. T. Johnson. 2020b. "NNV: The Neural Network Verification Tool for Deep Neural Networks and Learning-Enabled Cyber-Physical Systems". In: *International Conference on Computer-Aided Verification (CAV)*.

[64] Wang, S., K. Pei, J. Whitehouse, J. Yang, and S. Jana. 2018a. "Efficient formal safety analysis of neural networks". In: *Advances in Neural Information Processing Systems*.

[65] Wang, S., K. Pei, J. Whitehouse, J. Yang, and S. Jana. 2018b. "Formal Security Analysis of Neural Networks using Symbolic Intervals". In: *USENIX Security Symposium*.

[66] Weng, L., H. Zhang, H. Chen, Z. Song, C.-J. Hsieh, L. Daniel, D. Boning, and I. Dhillon. 2018a. "Towards Fast Computation of Certified Robustness for ReLU Networks". In: *International Conference on Machine Learning (ICML)*. Vol. 80. *Proceedings of Machine Learning Research*.

[67] Weng, T.-W., H. Zhang, P.-Y. Chen, J. Yi, D. Su, Y. Gao, C.-J. Hsieh, and L. Daniel. 2018b. "Evaluating the Robustness of Neural Networks: An Extreme Value Theory Approach". In: *International Conference on Learning Representations*.

[68] Wong, E. and Z. Kolter. 2018. "Provable Defenses against Adversarial Examples via the Convex Outer Adversarial Polytope". In: *International Conference on Machine Learning (ICML)*.

[69] Xiang, W., H. Tran, and T. T. Johnson. 2018a. "Output Reachable Set Estimation and Verification for Multilayer Neural Networks". *IEEE Transactions on Neural Networks and Learning Systems*. 29(11): 5777–5783.

[70] Xiang, W., H. Tran, J. A. Rosenfeld, and T. T. Johnson. 2018b. "Reachable Set Estimation and Safety Verification for Piecewise Linear Systems with Neural Network Controllers". In: *American Control Conference (ACC)*.

[71] Xiang, W., P. Musau, A. A. Wild, D. M. Lopez, N. Hamilton, X. Yang, J. Rosenfeld, and T. T. Johnson. 2018c. "Verification for Machine Learning, Autonomy, and Neural Networks Survey". *ArXiv*. (1810.01989).

[72]  Xiang, W., H.-D. Tran, and T. T. Johnson. 2017. "Reachable set computation and safety verification for neural networks with ReLU activations". *ArXiv*. (1712.08163).

[73]  Xiang, W., H.-D. Tran, and T. T. Johnson. 2018d. "Specification-Guided Safety Verification for Feedforward Neural Networks". *ArXiv*. (1812.06161).

[74]  Yang, X., H.-D. Tran, W. Xiang, and T. Johnson. 2020. "Reachability Analysis for Feed-Forward Neural Networks using Face Lattices". *ArXiv*. (2003.01226).

[75]  Zhang, H., T.-W. Weng, P.-Y. Chen, C.-J. Hsieh, and L. Daniel. 2018. "Efficient Neural Network Robustness Certification with General Activation Functions". In: *Advances in Neural Information Processing Systems (NeurIPS)*.

[76]  Zhang, H., P. Zhang, and C.-J. Hsieh. 2019. "RecurJac: An Efficient Recursive Algorithm for Bounding Jacobian Matrix of Neural Networks and Its Applications". In: *AAAI Conference on Artificial Intelligence (AAAI)*.

[77]  Zhang, Y. and Z. Zhang. 2013. "Dual Neural Network". In: *Repetitive Motion Planning and Control of Redundant Robot Manipulators*. Springer. 33–56.